%% file: main.tex
\documentclass[runningheads]{llncs}

\usepackage{eccv}

\usepackage{eccvabbrv}

\usepackage{graphicx}
\usepackage{booktabs}
\usepackage{pgfplots}
\usepackage{multirow}
\usepackage{adjustbox}
\usepackage{caption}
\usepackage{algorithm}
\usepackage{algpseudocode}
\usepackage{amsmath}
\usepackage{subcaption}
\usepackage{booktabs}
\usepackage{wrapfig}
\usepackage{placeins}
\usepackage{bm}
\usepackage{verbatim} 
\usepackage[frozencache,cachedir=minted-cache]{minted}
\pgfplotsset{width=12cm,compat=1.16}
\usepackage[accsupp]{axessibility}  

\usepackage{hyperref}

\usepackage{orcidlink}

\begin{document}


\title{Inf-DiT: Upsampling Any-Resolution Image with Memory-Efficient Diffusion Transformer} 

\titlerunning{Inf-DiT}

\author{Zhuoyi Yang\inst{1,\thanks{Done as intern at Zhipu AI, \ \email{yangzy22@mails.tsinghua.edu.cn} }}  \and
Heyang Jiang\inst{1} \and
Wenyi Hong\inst{1} \and
Jiayan Teng\inst{1} \\
Wendi Zheng\inst{1} \and
Yuxiao Dong\inst{1} \and
Ming Ding\inst{2} \and
Jie Tang\inst{1, \thanks{Corresponding author \email{jietang@tsinghua.edu.cn}}}
}

\authorrunning{Yang et al.}

\institute{\mbox{Tsinghua University \and Zhipu.AI}}

\maketitle

\begin{figure}[ht]
\centering
\includegraphics[width=\textwidth]{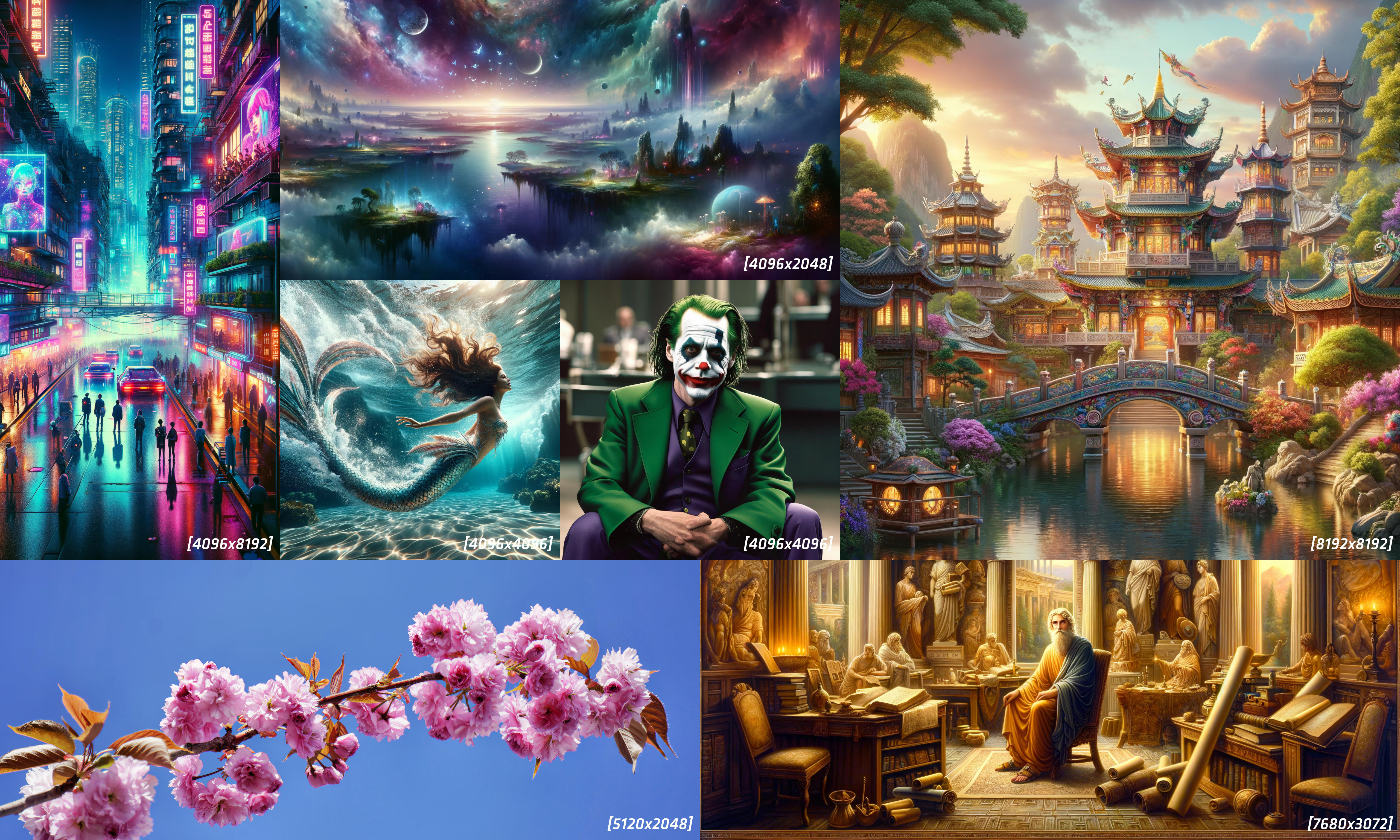}
\caption{
Selected ultra-high-resolution upsampling examples of Inf-DiT, based on SDXL, DALL-E 3, and real images.}
\label{fig:exampleImage}
\end{figure}
\vspace{-10mm}
\begin{abstract}
Diffusion models have shown remarkable performance in image generation in recent years. However, due to a quadratic increase in memory during generating ultra-high-resolution images (e.g. $4096\times4096$), the resolution of generated images is often limited to $1024\times1024$. In this work. we propose a unidirectional block attention mechanism that can adaptively adjust the memory overhead during the inference process and handle global dependencies. Building on this module, we adopt the DiT structure for upsampling and develop an infinite super-resolution model capable of upsampling images of various shapes and resolutions. Comprehensive experiments show that our model achieves SOTA performance in generating ultra-high-resolution images in both machine and human evaluation. Compared to commonly used UNet structures, our model can save more than $5 \times $ memory when generating $4096\times4096$ images. The project URL is \url{https://github.com/THUDM/Inf-DiT}.

\end{abstract}

\input{sections/1_introduction}

\input{sections/3_methodology}
\input{sections/4_experiments}
\input{sections/2_related_works}
\input{sections/5_conclusion}


%
%
\bibliographystyle{splncs04}
\bibliography{main}

\clearpage
\input{sections/6_appendix}
\end{document}

%% file: sections/1_introduction.tex
\section{Introduction}
\label{sec:intro}

\begin{wrapfigure}{tr}{0pt} 
\centering
\begin{minipage}[t]{0.4\columnwidth}
\vspace{-12mm}
\includegraphics[width=\linewidth]{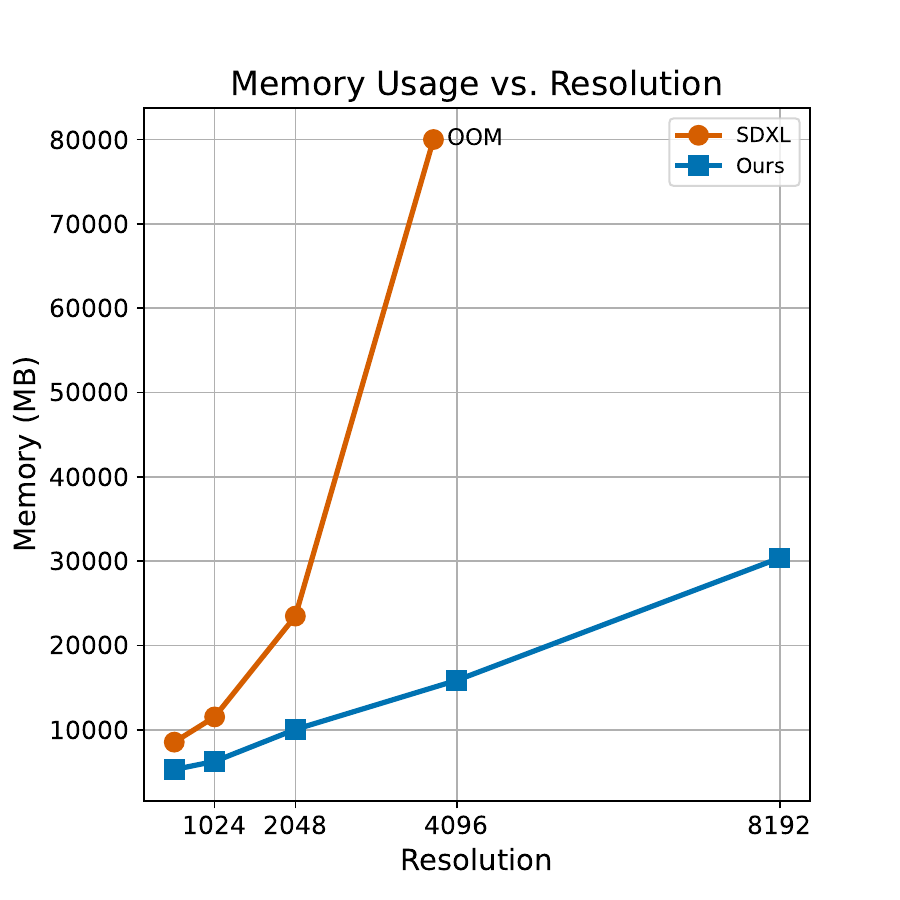}
\vspace{-7mm}
\caption{Comparison of memory usage during inference at different resolutions between our model and the SDXL architecture.}
\label{fig:memoryfig}
\vspace{-8mm}
\end{minipage}
\end{wrapfigure}

Recent years have witnessed rapid advancements in diffusion models, which significantly propelling the field of image generation and editing forward. Despite the advancements, a critical limitation persists: the resolution of images produced by existing image diffusion models is generally confined to $1024\times1024$ pixels or lower, posing a significant challenge for generating ultra-high-resolution images, which are indispensable in various real-world applications including intricate design projects, advertising, and the creation of posters and wallpapers, etc.



A commonly used approach for generating high resolution is cascaded generation~\cite{ho2022cascaded}, which first produces a low-resolution image, then applies multiple upsampling models to increase the image's resolution step by step. This approach breaks down the generation of high-resolution images into multiple tasks. Based on the results generated in the previous stage, the models in the later stages only need to perform local generation. Building upon the cascaded structure, both DALL-E2~\cite{rombach2022high} and Imagen~\cite{saharia2022photorealistic} can effectively generate images with 1024 resolution. 

The biggest challenge for upsampling to much higher resolution images is the significant GPU memory demands. For example, if utilizing the widely-adopted U-Net architecture such as SDXL~\cite{podell2023sdxl} for image inference (see \Cref{fig:memoryfig}), we observe a dramatic escalation in memory consumption with increasing resolution. Specifically, generating a 4096$\times$4096 resolution image, which comprises over 16 million pixels requires more than 80GB of memory, exceeding the capacities of standard RTX 4090 or A100 graphics cards. Furthermore, the process of training models for high-resolution image generation exacerbates these demands, as it necessitates additional memory for storing gradients, optimizer states, etc. LDM~\cite{rombach2022high} reduces the memory consumption by utilizing Variational Autoencoder(VAE) to compress images and generating images in a smaller latent space. However, it is also emphasized that an excessively high compression ratio can substantially deteriorate the quality of generation, imposing severe limitations on the reduction of memory consumption.

We observe that one common and critical problem among the aforementioned models is the necessity to input the entire image to the model, which requires keeping $O(N^2)$ hidden states in memory, where $N$ is the width(height) of the image. Based on this observation, we propose a Unidirectional Block Attention (UniBA) algorithm that can dramatically reduce the space complexity of generation from $O(N^2)$ to $O(N)$, increasing the highest available resolution for a large margin (\cref{fig:memoryfig}). Specifically, in every diffusion step, we split the image into blocks and perform a sequential batch generation order among them, where each batch simultaneously produces a subset of the blocks, and any amount of blocks can be generated in parallel as long as the memory restriction allows. It is worth noting that, though the model only inputs part of image into the model at the same time, UniBA successfully preserves the ability to interact with hidden states of faraway blocks and maintain high-level semantic consistency. This is different from other block-based generation methods such as \cite{bar2023multidiffusion, jimenez2023mixture, wang2023exploiting} that only interact with blocks in pixel space or compressed latent space. 

Based on this algorithm, we optimize diffusion transformer (DiT) and train a model named Inf-DiT, which is capable of upsampling images of varying resolution and shape. Furthermore, we design several techniques including providing global image embedding to enhance the global semantic consistency and offer zero-shot text control ability, and provision of all neighboring low-resolution(LR) blocks through cross-attention mechanisms to further enhance the local consistency.
Evaluation results show that Inf-DiT achieved significantly superior results compared to other high-resolution generative models in both machine and human evaluation.

To summarize, our main contributions are as follows:
\begin{enumerate}
    \item We propose Unidirectional Block Attention (UniBA) algorithm, which reduces the minimum memory consumption from $O(N^2)$ to $O(N)$ during inference, where $N$ represents the edge length. This mechanism is also capable of adapting to various memory restrictions by adjusting the number of blocks generated in parallel, trading off between memory and time overhead.
    \item  Based on these methods, we train an image upsampling diffusion model, Inf-DiT, a 700M model capable of upsampling images of varying resolutions and shapes. Inf-DiT achieves state-of-the-art performance in both machine (HPDv2 and DIV2K datasets) and human evaluation. 
    \item We design multiple techniques to further enhance local and global consistency, and offer a zero-shot ability for flexible text control.
\end{enumerate}

%% file: sections/3_methodology.tex
\section{Methodology}



\subsection{Unidirectional Block Attention (UniBA)}

\begin{figure}[htp]
    \centering
    
    \includegraphics[width=0.93\linewidth]{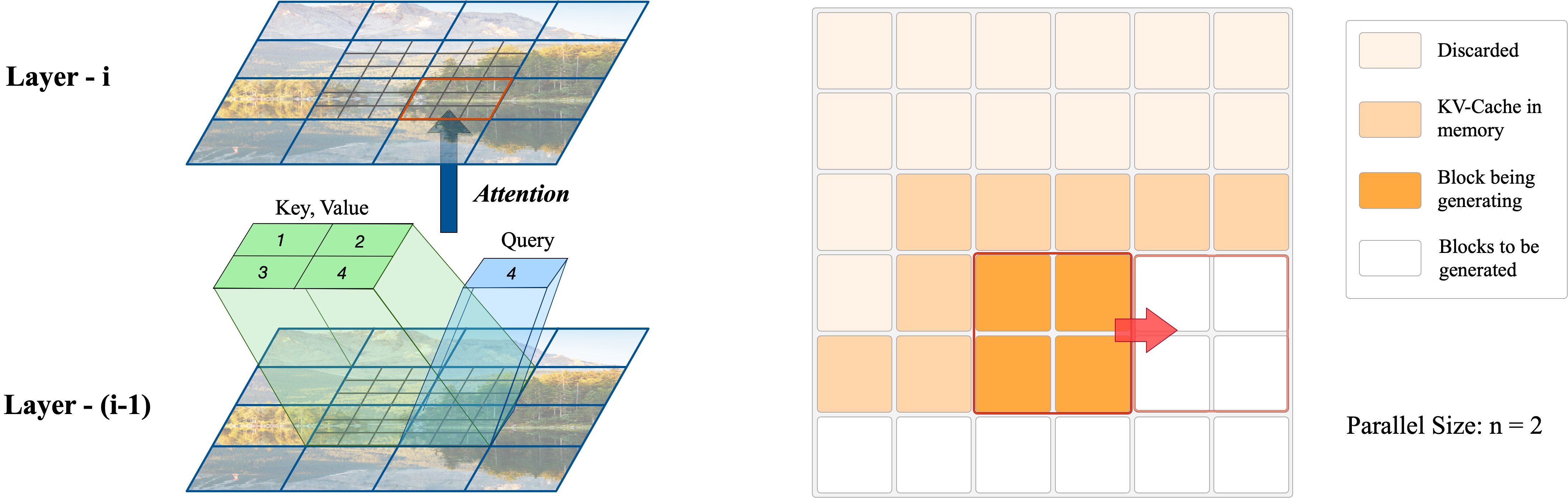}
    \caption{\textbf{Left}: Unidirectional block attention. In our implementation, each block directly depends on the three blocks in each layer: the blocks in the upper left corner, on the left, and on the top. \textbf{Right}: The inference process of Inf-DiT. Inf-DiT generates $n\times n$ block each time, based on the memory size. During this process, only the KV-cache of the blocks that subsequent blocks depend on are stored in memory. }
    \label{fig:cache}
     \vspace{-5mm}
    
\end{figure}


We observe that the critical obstacle for generating ultra-high resolution images is memory limitation. As the resolution of the image increases, the corresponding hidden state's size within the network expands quadratically. For example, a single hidden state with the shape of $2048 \times 2048 \times 1280$ in only one layer requires 20GB of memory, making it formidable to generate very large images. How to avoid storing the entire image's hidden state in memory becomes the key issue. 



Our main idea is to divide an image 
$\bm{x} \in \mathbb{R}^{H \times W \times C}$
into blocks 
$\bm{x}_b \in \mathbb{R}^{h \times w \times B^2 \times C}$
, where $B$ is block size and $h=\frac{H}{B}, w=\frac{W}{B}$. When the image is fed into the network, the channel size and resolution of a block may change, but the layout and the relative positional relationships between blocks will remain unchanged. If there is a way to apply sequential batch generation of blocks where each batch simultaneously produces a subset of the blocks, only a small number of block hidden states have to be kept in memory simultaneously, making it possible to generate ultra-high-resolution images. 


Here we define that $\text{block}_\text{A}$ is dependent on $\text{block}_\text{B}$ if the generation of $\text{block}_\text{A}$ involves the hidden state of $\text{block}_\text{B}$ in computation. It can be observed that the dependencies between blocks are bidirectional in most previous structures (UNet, DiT, etc.), in which case all blocks in the image must be generated simultaneously. Take UNet as an example: two adjacent elements in neighboring blocks use each other's hidden state in the convolution operation, therefore all pairs of neighboring blocks must be generated simultaneously. Given the aim to save the memory of blocks' hidden states, we hope to devise an algorithm that allows the blocks in the same image to be divided into several batches for generation, with each batch only needing to generate a portion of the blocks at the same time, and the batches are generated in sequence. Generally, an image generation algorithm can perform such a sequential batch generation among blocks if it meets the following conditions: 
\begin{enumerate}
    \item The generative dependency between blocks are unidirectional, and can form a directed acyclic graph (DAG). 
    \item Each block has only a few direct (1st-order) dependencies on other blocks, since the hidden states of the block and its direct dependencies should simultaneously be kept in the memory.
\end{enumerate}
Furthermore, to ensure consistency across the whole image, it also required that the blocks have a large enough receptive field to manage long-range dependencies.


According to the conditions and analysis above, we choose an efficient implementation illustrated in \cref{fig:cache}, Unidirectional Block Attention(UniBA). For each layer, every block directly depends on three 1st-order neighboring blocks: the block on the top, on the left, and in the upper left corner. 
For example, if we adopt Diffusion Transformer (DiT) architecture which is the base architecture of Inf-DiT, the dependency between blocks is attention operation, where the query vectors of each block interact with the key, value vectors of 4 blocks: 3 blocks located to its upper left and itself, illustrated in \cref{fig:cache}. 

Formally, the UniBA process in the transformer can be formulated as
\begin{align}
\bm{Q} &= \bm{z}^n_{(i, j)} \bm{W}^Q  , \\ 
\bm{K} &= [\bm{z}^n_{(i, j)}+\bm{P}_1 ; \bm{z}^n_{(i-1, j)} + \bm{P}_2; \bm{z}^n_{(i, j-1)} + \bm{P}_3; \bm{z}^n_{(i-1, j-1)} + \bm{P}_4]\bm{W}^K  , \\
\bm{V} &= [\bm{z}^n_{(i, j)} ; \bm{z}^n_{(i-1, j)}; \bm{z}^n_{(i, j-1)}; \bm{z}^n_{(i-1, j-1)}]\bm{W}^V , \\
\bm{z}^{n+1}_{(i,j)} &= \text{FFN}(\text{Attention}(\bm{Q},\bm{K},\bm{V})) ,
\end{align} 
where $\bm{z}^n_{(i, j)}$ is the hidden states of the block at row $i$, column $j$ in layer $n$,and $\bm{P}_i$ is the block-level relative position encoding.
We also implement an efficient approach to apply UniBA for full image in pytorch style, which is attached in the \cref{sec:appendix:imp}.

Note that, though each block only attends to a few number of neighboring blocks in each layer, as features propagate layer by layer,  blocks can indirectly interact with faraway blocks, thereby capturing both long- and short-range relationships. Our design shares a similar spirit with the natural language model Transformer-XL~\cite{dai2019transformer}, which can be viewed as a special form of ours in one dimension case. 

\subsubsection{Inference process with O(N) Memory Consumption}
Although our method can generate each block sequentially, it differs from auto-regressive generative models, in which the next block depends on the final output of the previous block. Any number of blocks can also be generated in parallel in our model, as long as the union of their dependent blocks has been generated. Based on this property, we implement a simple but effective inference process. As illustrated in \cref{fig:cache}, we generate $n \times n$ blocks at once, from the top-left to bottom-right. After generating a set of blocks, we discard hidden states i.e. KV-cache that are no longer used and append newly generated KV-cache to the memory.   

It can be easily proved that the number of block KV-cache retained in memory during the process is always $ \leq w + n$. 
Assume the space needed by the model when generating a single block is $M_1$, the space for one block's KV-cache is $M_2$, and other essential space consumption (such as storing the raw input image) is $C$, then the maximum space usage of the inference process is $n^2M_1 + (w+n)M_2 + C$. When $n$ is much smaller than $w$, the memory consumption is directly proportional to $w$. If $w$ is bigger than $h$, we can easily change the trajectory of generation to column-major.

In practical terms, despite the total FLOPs of image generation remaining constant for various values of n, due to the overhead such as operator initialization time and memory allocation time, the generation time decreases when $n$ increases. Therefore, it's optimal to choose the largest $n$ allowed by the memory limitation.







\subsection{Basic Model Architecture}

\begin{figure}
    \centering
    
    \includegraphics[width=\linewidth]{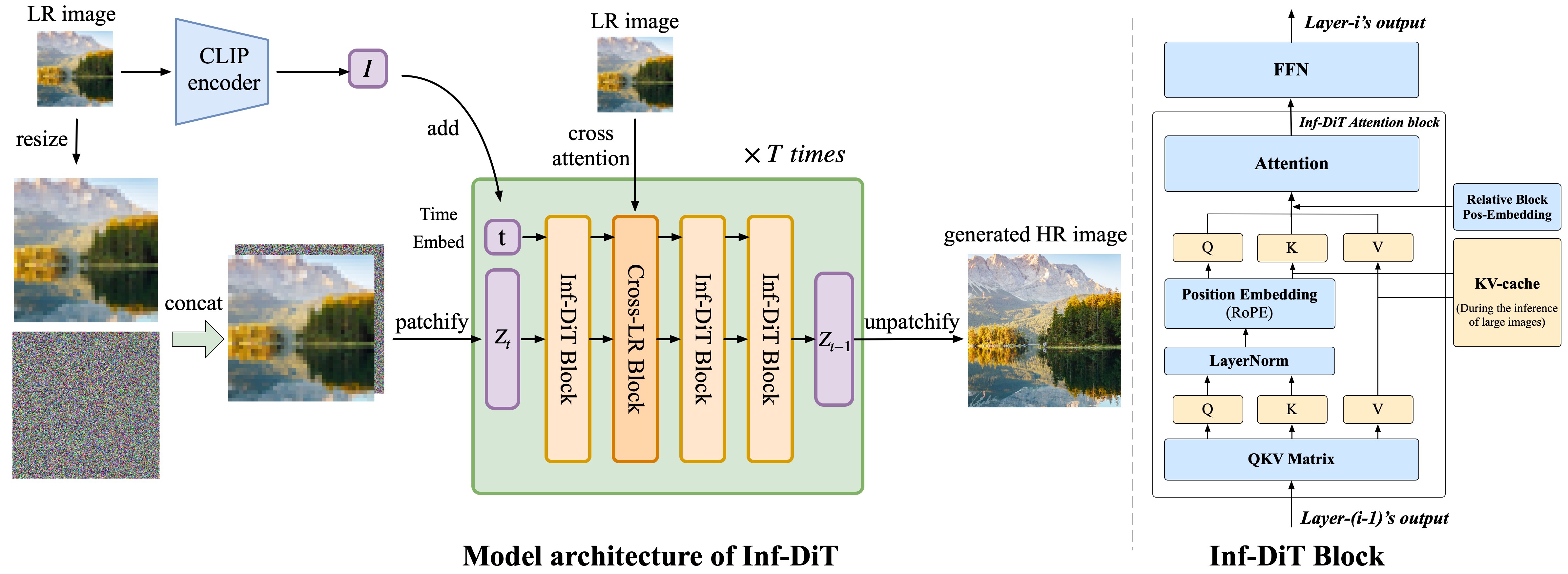}
    \caption{(Left) Overall architecture of Inf-DiT. (Right) The inner structure of Inf-DiT block. We do not depict the Layernorm that originally existed in the DiT for simplicity. }
    \label{fig:dependency}
    \vspace{-4mm}
\end{figure}

\Cref{fig:dependency} provides an overview of our model, Inf-DiT's architecture. The model uses a similar backbone as DiT~\cite{peebles2023scalable}, which applies Vision Transformer (ViT)~\cite{dosovitskiy2020image} to diffusion models and proves its efficacy and scalability. In addition to its superior performance, compared to convolution-based architectures such as UNet\cite{ronneberger2015u}, DiT solely utilizes attention as the mechanism of interaction among patches, which can conveniently implement unidirectional block attention. To adapt to unidirectional block attention and enhance the performance of upsampling, we make several modifications and optimizations detailed as follows.

\subsubsection{Model input}
Inf-DiT first partitions input images into multiple non-overlapping \textit{blocks}, which are further divided into \textit{patches} with a side length equal to the \textit{patch size}. Unlike DiT, considering the compression loss such as color shifting and detail loss, the patchifying of Inf-DiT is conducted in \textbf{RGB pixel space} instead of latent space. In the case of super-resolution by $f$ times, Inf-DiT first upsamples the low-resolution RGB image condition by a factor of $f$, then concatenates it with the diffusion's noised input on the feature dimension before feeding it to the model.

\subsubsection{Position Encoding}
Unlike UNet-based diffusion models\cite{rombach2022high} that can perceive positional relationships through convolution operations, all operations including self-attention, FFN in transformers are permutation invariant functions. Therefore, transformer-based models require auxiliary input of explicit positional information to learn the relationships among patches. 
As recent research in LLMs has shown that relative positional encoding is more effective in capturing word position relevance compared to absolute positional encoding, we refer to the design of Rotary Positional Encoding (RoPE) \cite{su2024roformer}, which performs well in the long context generation, and adapt it into two-dimensional form for image generation. 
Specifically, we divide channels of hidden states in half, one for encoding the x-coordinate and the other for the y-coordinate, and apply RoPE in both halves. 

We create a sufficiently large rope positional encoding table to ensure it meets the requirements during generation. To ensure all parts of the positional encoding table can be seen by the model during training, we employ the \textit{Random Starting Point}: For each training image, we randomly assign a position $(x,y)$ for the top-left corner of the image, instead of the default $(0,0)$. 

In addition, considering the difference in interactions within the same block and between different blocks, we additionally introduce block-level relative position encoding $\bm{P}_{1\sim4}$, which assigns a distinct learnable embedding based on the relative position before attention.


\subsection{Global and Local Consistency}



\subsubsection{Global Consistency with CLIP Image Embedding}
The global semantic information within low-resolution (LR) images, such as artistic style and object material, plays a crucial role during upsampling. However, compared to text-to-image generation models, the upsampling model has an additional task: understanding and analyzing the semantic information of LR images, which significantly increases the model's burden. This is particularly challenging when training without text data, as high-resolution images rarely have high-quality paired texts, making these aspects difficult for the model.



Inspired by DALL·E2~\cite{ramesh2022hierarchical}, we utilize the image encoder from pre-trained CLIP~\cite{radford2021learning} to extract image embedding $I_{LR}$ from low-resolution images, which we refer to as \textit{Semantic Input}. Since CLIP is trained on massive image-text pairs from the Internet, its image encoder can effectively extract global information from low-resolution images. We add the global semantic embedding to the time embedding of the diffusion transformer and input it into each layer, enabling the model to learn directly from high-level semantic information.

Another interesting advantage of global semantic embedding is that, using the aligned image-text latent space in CLIP, \emph{we can use text to guide the direction of generation, even if our model has not been trained on any image-text pairs}. Given a positive prompt $C_{pos}$ and a negative prompt $C_{neg}$, we can update our image embedding:
\setlength{\abovedisplayskip}{5pt}
\setlength{\belowdisplayskip}{5pt}
\begin{align}
\tilde{\bm{I}}_{LR} = \text{norm}(\bm{I}_{LR} + \alpha(\text{TextEnc}(C_{pos}) - \text{TextEnc}(C_{neg}))) ,
\end{align} 
where $\alpha$ can control the intensity of the guidance. During inference, we can simply use $\Tilde{\bm{I}}_{LR}$ in place of $\bm{I}_{LR}$ as the global semantic embedding to conduct the control.
For example, to get a more clear result set, $C_{pos}=\text{``}clear\text{''}$ and $C_{neg} =\text{``}blur\text{''}$ sometimes help.

\vspace{-4mm}
\subsubsection{Local Consistency with Nearby LR Cross Attention}
Although concatenating the LR image with noised input already provides a good inductive bias for models to learn the local correspondence between LR and HR images, there can still be continuity issues. The reason is that, there are several possibilities of upsampling for a given LR block, which require analysis in conjunction with several nearby LR blocks to select one solution. Assume that the upsampling is only performed based on the LR blocks to its left and above, it may select an HR generation solution that conflicts with the LR block to the right and below. Then when upsampling the LR block to the right, if the model considers conforming to its corresponding LR block more important than being continuous with the block to the left, a HR block discontinuous with previous blocks would be generated. 
A naive solution is to input the entire LR image to every block, but it's too costly when the resolution of LR image is also big. 

To handle this problem, we introduce Nearby LR Cross Attention. In the first layer of the transformer, each block conducts cross-attention on the surrounding $3 \times 3$ LR blocks to capture nearby LR information. Our experiments show that this approach significantly reduces the probability of generating discontinuous images. It is worth noting that this operation will not change our inference process since the entire LR image is known before generation. 

We further devise techniques including continuity class-free guidance, LR-based noise initialization, QK Normalization, etc., which are elaborated in detail in the \cref{sec:appendix:imp}.



%% file: sections/4_experiments.tex
\section{Experiments}

In this section, we first introduce the detailed training process of Inf-DiT, then comprehensively evaluate Inf-DiT by both machine and human evaluation. The results show that Inf-DiT surpasses all baselines, excelling in both ultra-high-resolution image generation and upsampling tasks. Finally, we conduct ablation studies to validate the effectiveness of our design.

 \vspace{-5mm}
\begin{table*}[htbp]
    \renewcommand{\arraystretch}{1}
    \caption{Quantitative comparison results with state-of-the-art methods of ultra-high-resolution generation on HPDV2 dataset. The best results are marked in \textbf{bold}, and the second best results are marked by \underline{underline}.}
    \centering
    \begin{adjustbox}{width=\linewidth,center}
    \setlength{\tabcolsep}{11.5pt}
    \begin{tabular}{c|cc|cc|c}
    \toprule[1pt]
    \multirow{2}{*}{\textbf{Method}} & \multicolumn{2}{c|}{$2048 \times 2048$} & \multicolumn{2}{c|}{$4096 \times 4096$} & \multirow{2}{*}{\textbf{Mean} $\downarrow$} \\
    \cline{2-5} & \rule{0pt}{2.6ex}FID $\downarrow$  & FID$_{\text{crop}}$ $\downarrow$  & FID $\downarrow$  & FID$_{\text{crop}}$ $\downarrow$  & \\
    \midrule[1pt]
        Direct Inference & $92.2$ & $92.3$ & OOM & OOM & $92.3$ \\
        MultiDiffusion & $99.7$ & $109.3$ & OOM & OOM & $104.5$ \\
        SDXL+BSRGAN & $\underline{66.3}$ & $81.7$ & $\textbf{66.0}$ & $85.0$ & $74.8$ \\
        ScaleCrafter & $79.9$ & $95.5$ & $115.1$ & $153.3$ & $111.0$ \\
        DemoFusion & $67.5$ & $\underline{72.0}$ & $70.8$ & $\underline{82.6}$ & $\underline{73.2}$ \\
        SDXL+Inf-DiT(Ours) & $\textbf{66.0}$ & $\textbf{71.9}$ & $\underline{67.0}$ & $\textbf{76.2}$ & $\textbf{70.3}$ \\
    \midrule[1pt]
    \end{tabular}
    \end{adjustbox}
    \vspace{0.1mm}
    \label{tbl:comparison}
     \vspace{-1mm}
    \end{table*}
\vspace{-10mm}
\begin{table*}[htbp]
    \renewcommand{\arraystretch}{1}
    \caption{Quantitative comparison results with state-of-the-art methods of super resolution on DIV2K dataset. The best results are marked in \textbf{bold}.}
    \centering
    \begin{adjustbox}{width=\linewidth,center}
    \setlength{\tabcolsep}{8.5pt}
    \begin{tabular}{c|c|cccc}
    \toprule[1pt]
      \textbf{Setting} & \textbf{Method} & FID $\downarrow$ & FID$_{\text{crop}}$ $\downarrow$ & PSNR $\uparrow$ & SSIM $\uparrow$ \\
    \midrule[1pt]
     \multirow{3}{*}{Variable resolution} & BSRGAN & 35.0 & 143.2 & 25.9 & 0.73 \\
     & Real-ESRGAN & 35.5 & 127.6 & 25.2 & 0.72 \\
     & Inf-DiT(Ours) & \textbf{20.2} & \textbf{76.5} & \textbf{26.3} & \textbf{0.74} \\
    \midrule[0.5pt]
     \multirow{2}{*}{Fix resolution (512x512)} & StableSR & 74.8 & 112.6 & 22.0 & 0.60  \\
     & Inf-DiT(Ours) & \textbf{38.6} & \textbf{83.3} & \textbf{24.6} & \textbf{0.67}\\
    \midrule[0.5pt]
     \multirow{2}{*}{Fix resolution (256x256)} & LDM & 152.6 & - & 23.9 & 0.66 \\
     & Inf-DiT(Ours) & \textbf{86.8} & - & \textbf{24.6} & \textbf{0.67} \\
    \bottomrule[1pt]
    \end{tabular}
    \end{adjustbox}
    \vspace{0.1mm}
    \label{tbl:resolution_comparison}
     \vspace{-5mm}
\end{table*}




\subsection{Training Details}

\subsubsection{Datasets}

Our dataset comprises a subset of LAION-5B\cite{schuhmann2022laionb} with a resolution higher than $1024\times1024$ and aesthetic score higher than 5, and 100 thousand high-resolution wallpapers from the Internet.
Following the previous works\cite{saharia2022photorealistic, ramesh2022hierarchical, wang2021real}, we use fixed-size image crops of $512 \times 512$ resolution during training. Since upsampling can be conducted with local information only, it can be directly employed at higher resolution during inference, which is not easy for most generation models. 
 \vspace{-5mm}
\subsubsection{Data Processing}
Since the images generated by diffusion models often contain residual noise and various detail inaccuracies, it becomes crucial to enhance the robustness of the upsampling model to address these issues. We adopt the approach similar to Real-ESRGAN\cite{wang2021real} to perform a variety of degradation on low-resolution input images within the training data. 

When processing training images with a resolution higher than 512, there are two alternative methods: directly performing a random crop, or resizing the shorter side to 512 before performing a random crop. While the direct cropping method preserves high-frequency features in high-resolution images, the resize-then-crop method avoids frequently cropping out areas with a single color background, which is detrimental to the model's convergence. Therefore, in practice, we randomly select from these two processing methods to crop training images. 


 \vspace{-5mm} 
\subsubsection{Training settings}
During training, We set $block \ size=128$ and $patch\  size=4$, which means every training image is divided into $4\times4$ blocks and every block has $32\times 32$ patches. 
We employ the framework of EDM\cite{karras2022elucidating} for training, and set the upsampling factor to $4 \times$. Because the upsampling task is more concerned with the high-frequency details of images, we adjusted the mean and std of training noise distribution to $-1.0$ and $1.4$. To address overflow problem during training, we employed BF16 format due to its broader numerical range. Our CLIP model is a ViT-L/16 pre-trained on Datacomp dataset\cite{ilharco_gabriel_2021_5143773}. Since the CLIP can only process images with $224 \times 224$ resolution, we first resize LR images to $224 \times 224$ and then input them to CLIP.
Other hyperparameters are listed in the \cref{sec:appendix:imp}. 

\begin{figure}
\centering
\begin{subfigure}{.14\textwidth}
  \centering
  \includegraphics[width=\linewidth]{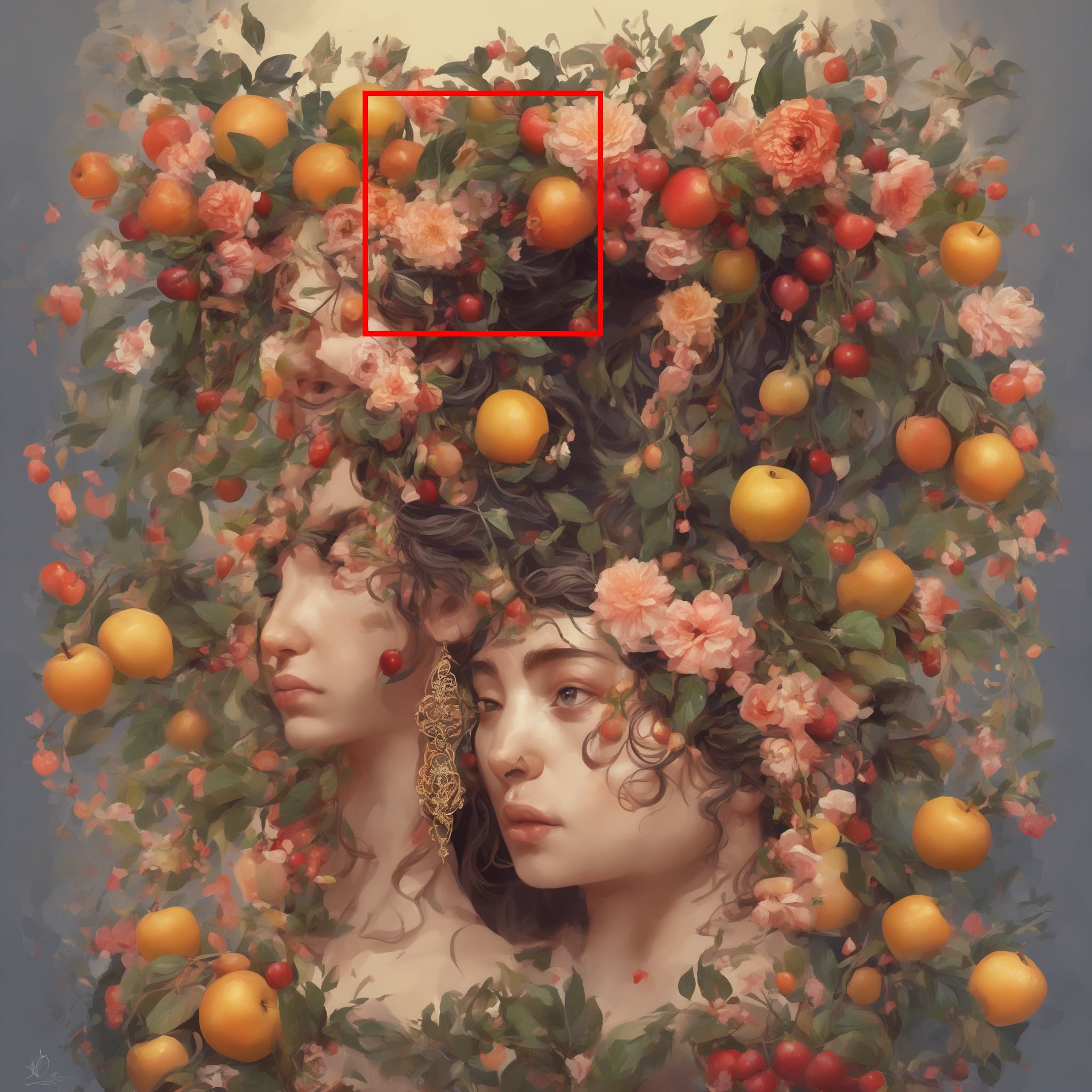}
\end{subfigure}%
\begin{subfigure}{.14\textwidth}
  \centering
  \includegraphics[width=\linewidth]{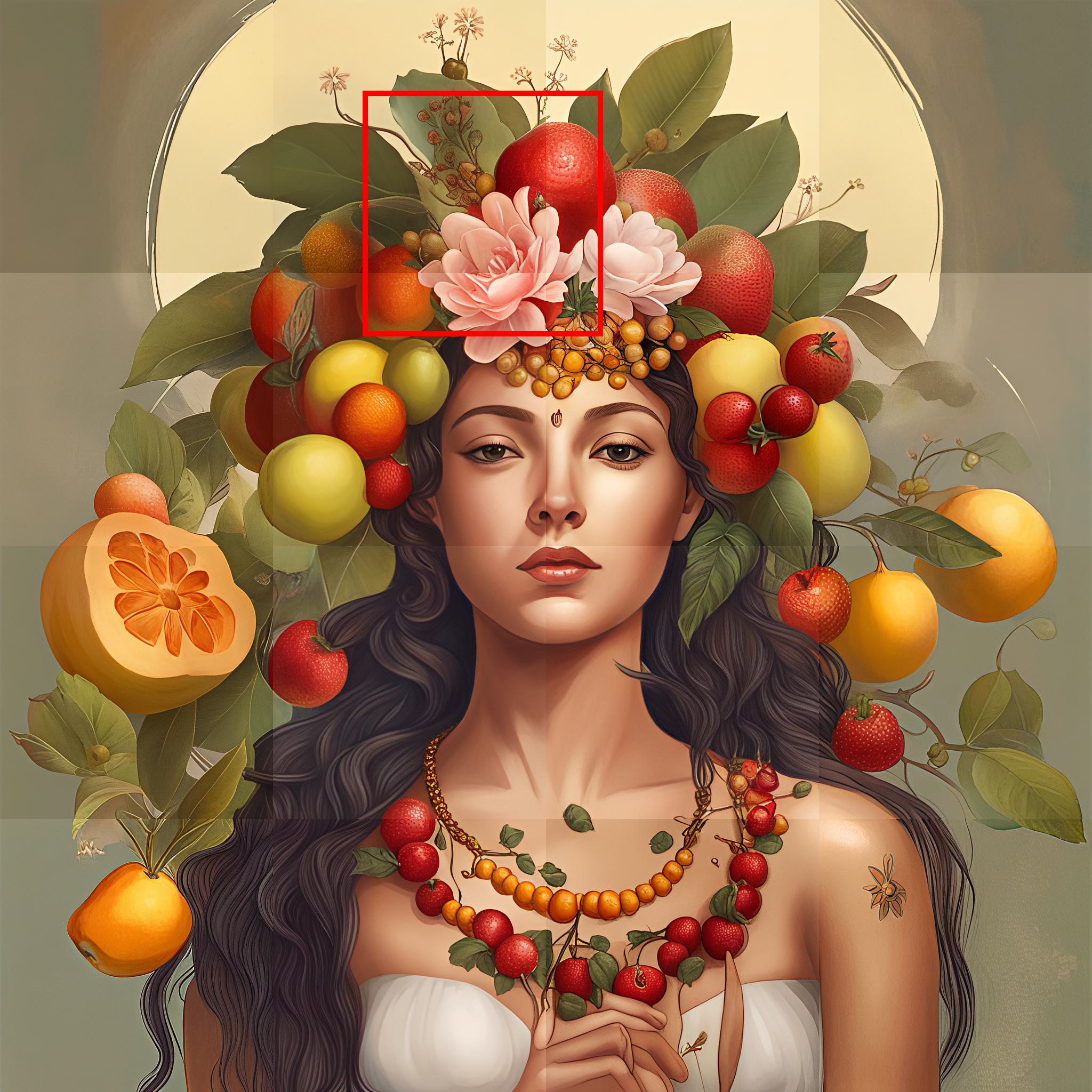}
\end{subfigure}%
\begin{subfigure}{.14\textwidth}
  \centering
  \includegraphics[width=\linewidth]{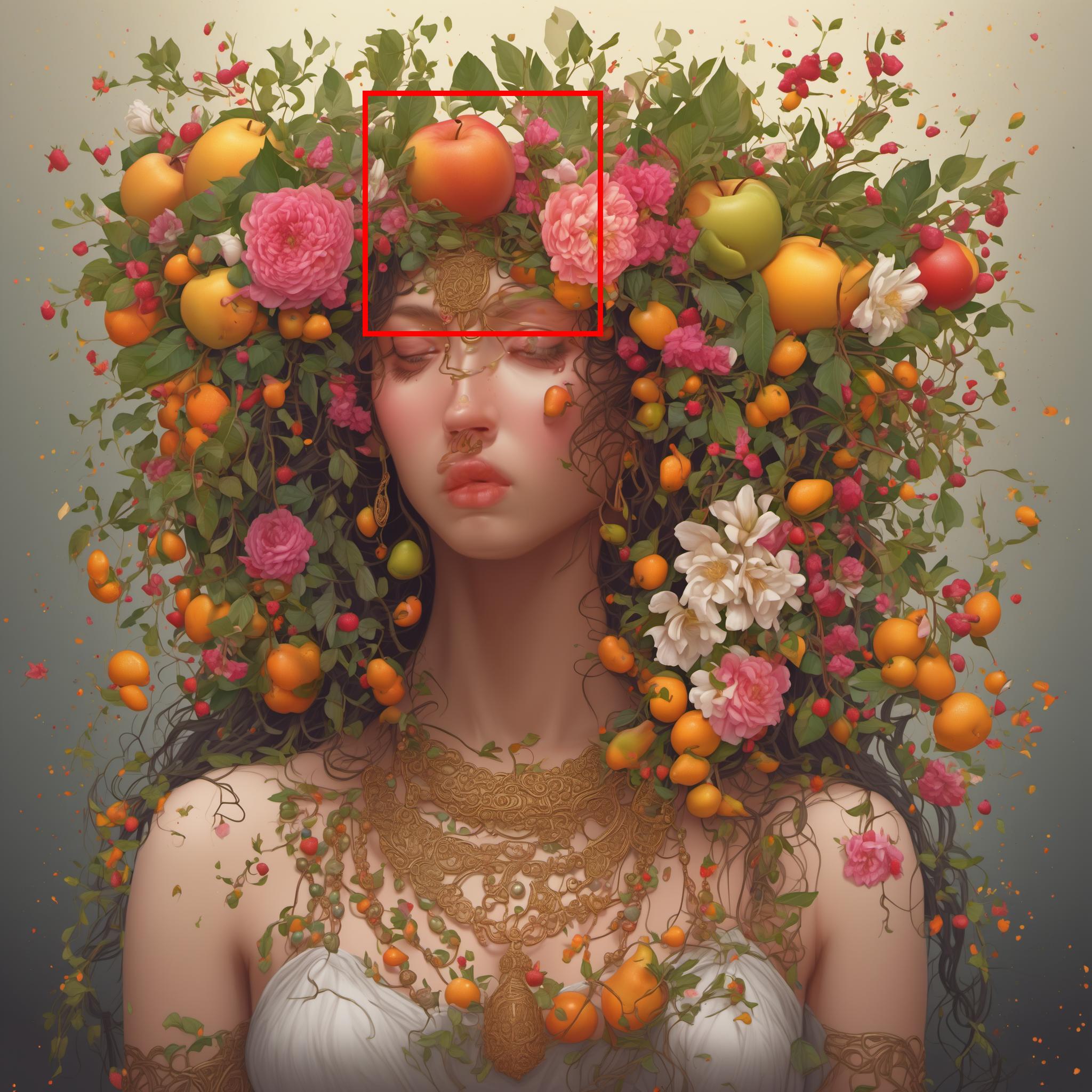}
\end{subfigure}%
\begin{subfigure}{.14\textwidth}
  \centering
  \includegraphics[width=\linewidth]{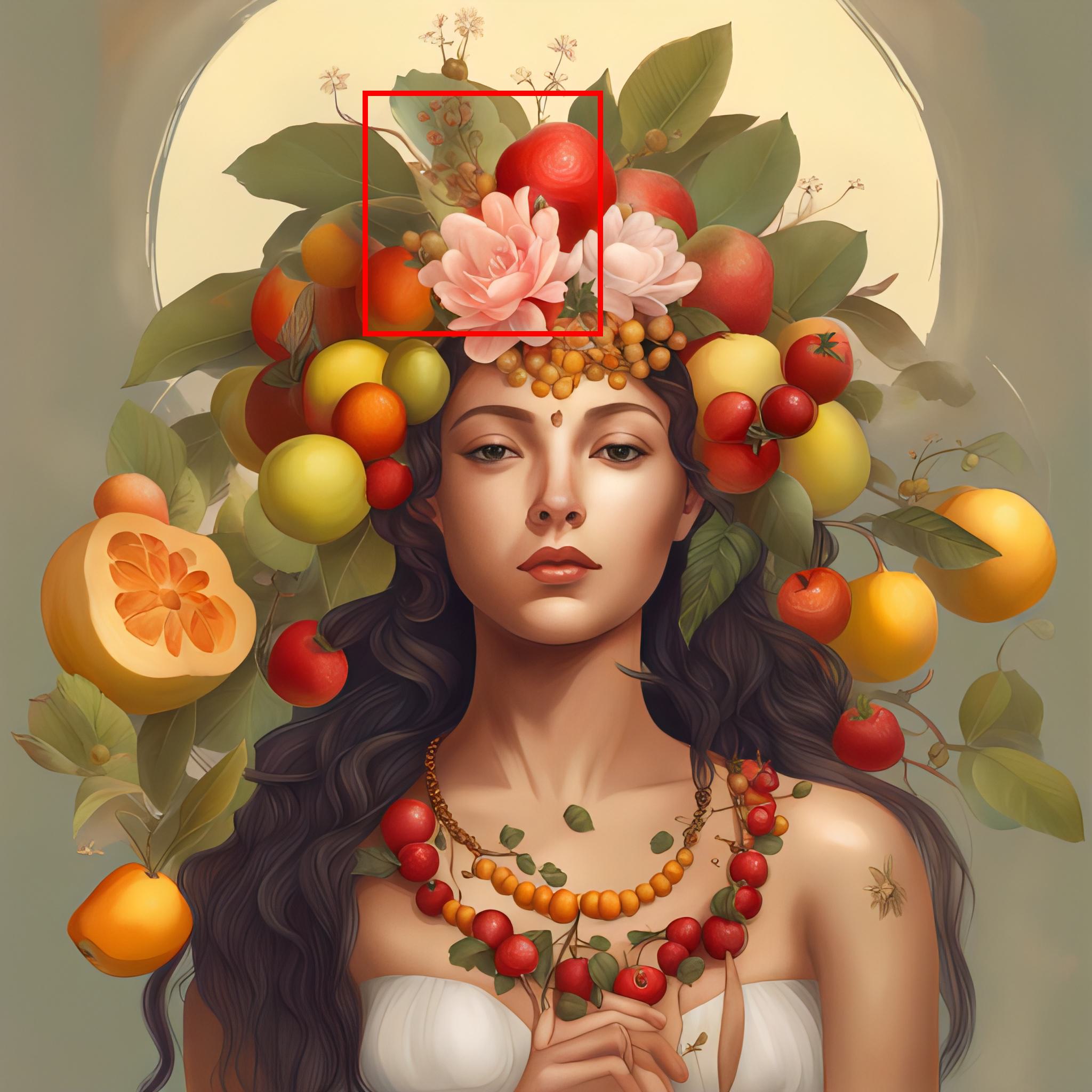}
\end{subfigure}%
\begin{subfigure}{.14\textwidth}
  \centering
  \includegraphics[width=\linewidth]{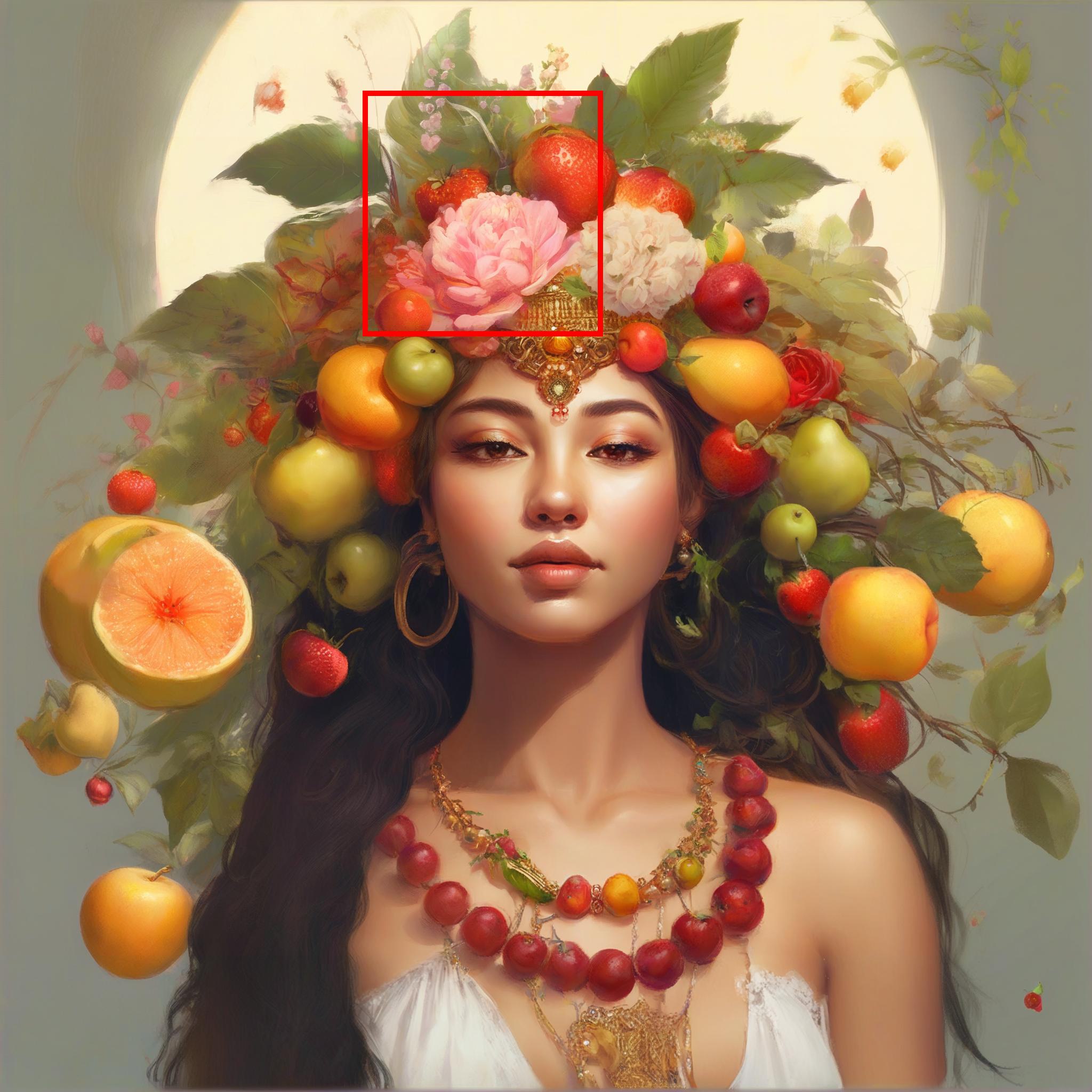}
\end{subfigure}%
\begin{subfigure}{.14\textwidth}
  \centering
  \includegraphics[width=\linewidth]{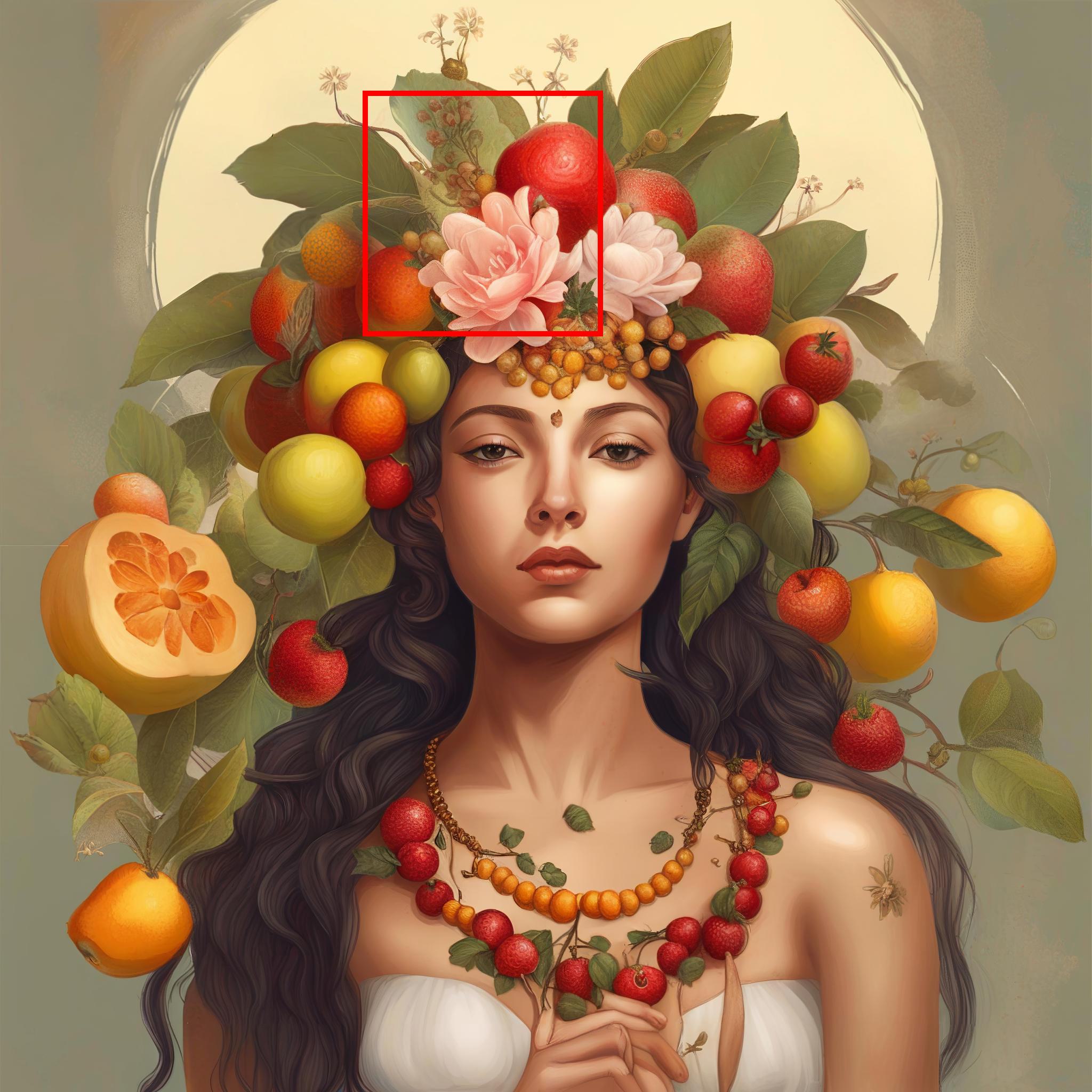}
\end{subfigure}
\begin{subfigure}{.14\textwidth}
  \centering
  \includegraphics[width=\linewidth]{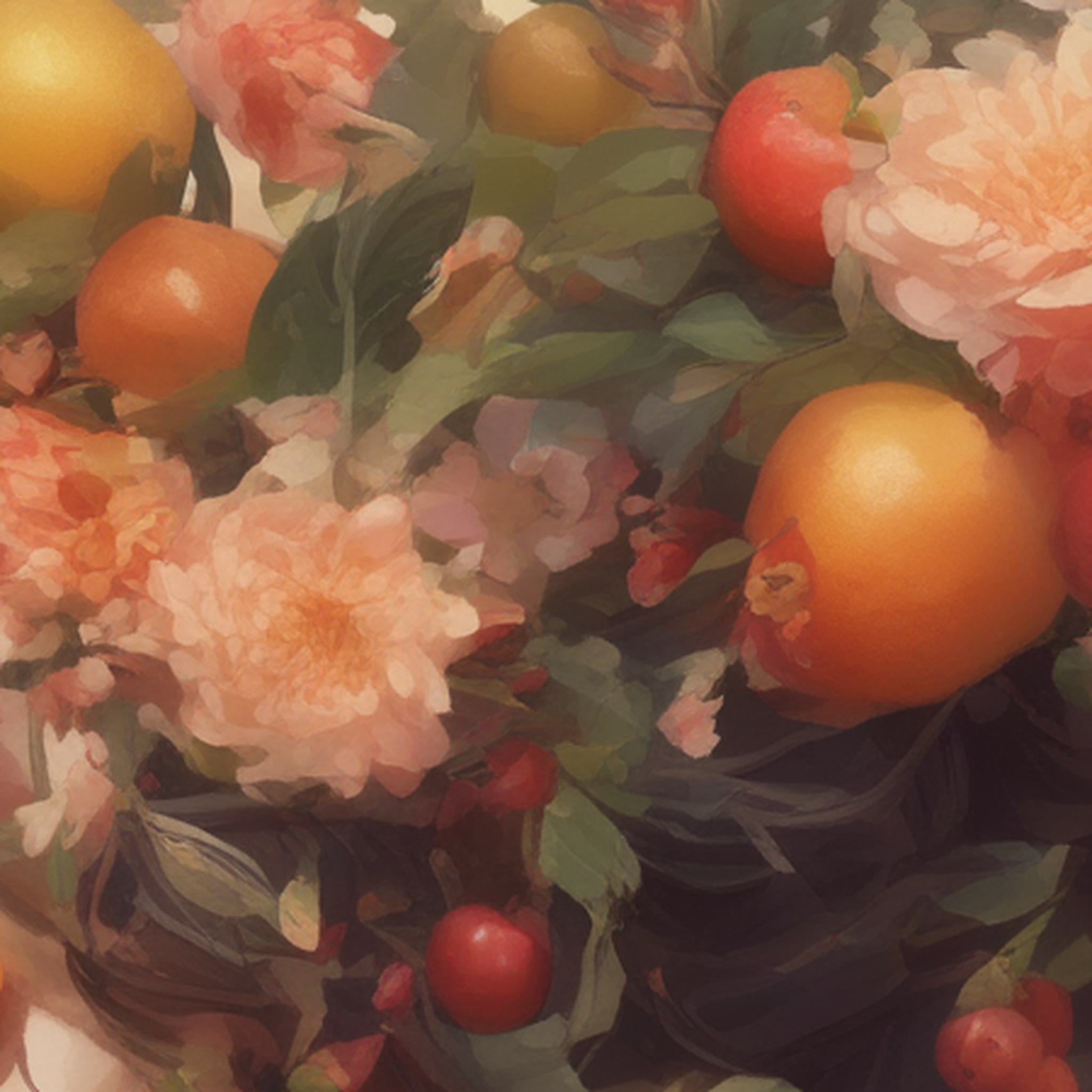}
\end{subfigure}%
\begin{subfigure}{.14\textwidth}
  \centering
  \includegraphics[width=\linewidth]{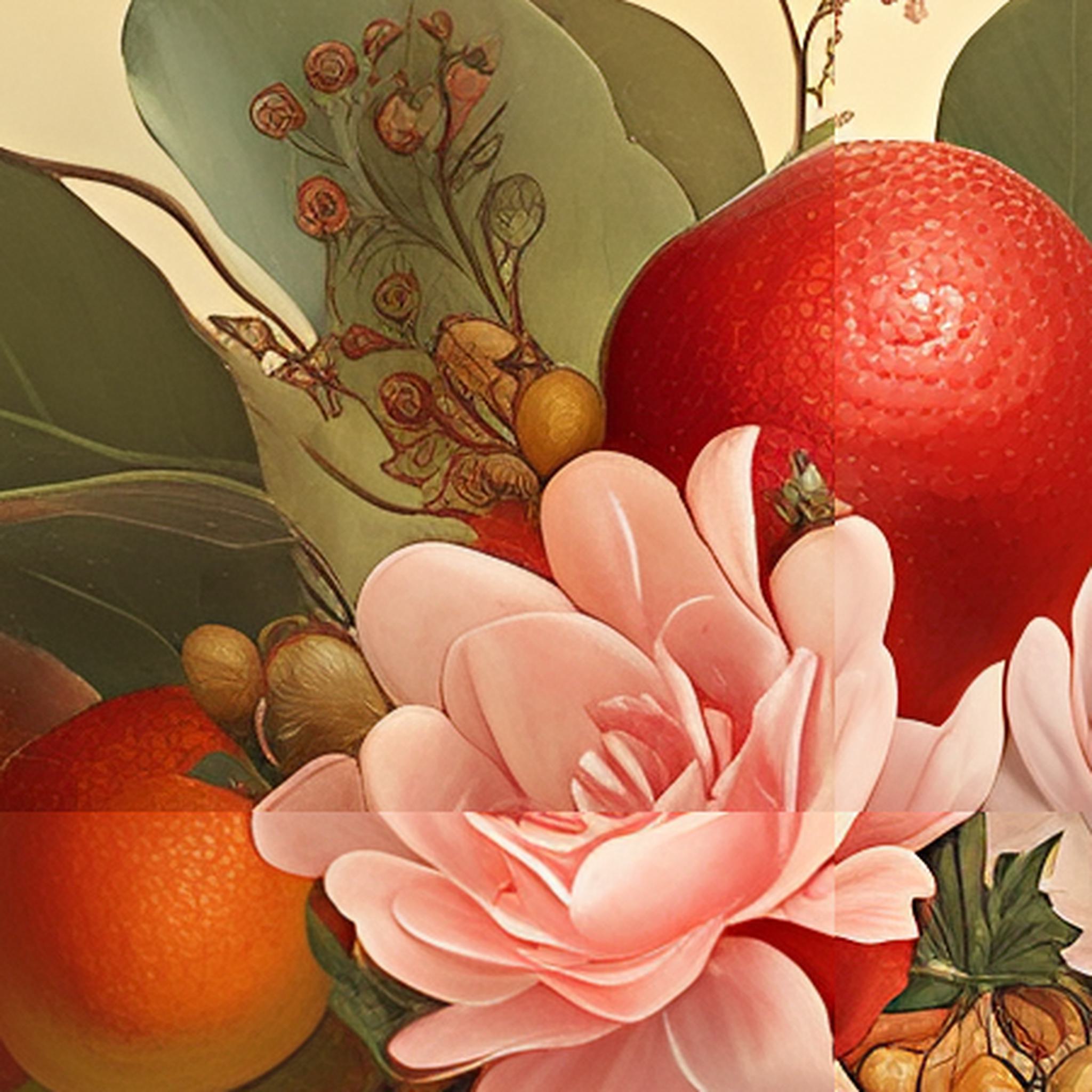}
\end{subfigure}%
\begin{subfigure}{.14\textwidth}
  \centering
  \includegraphics[width=\linewidth]{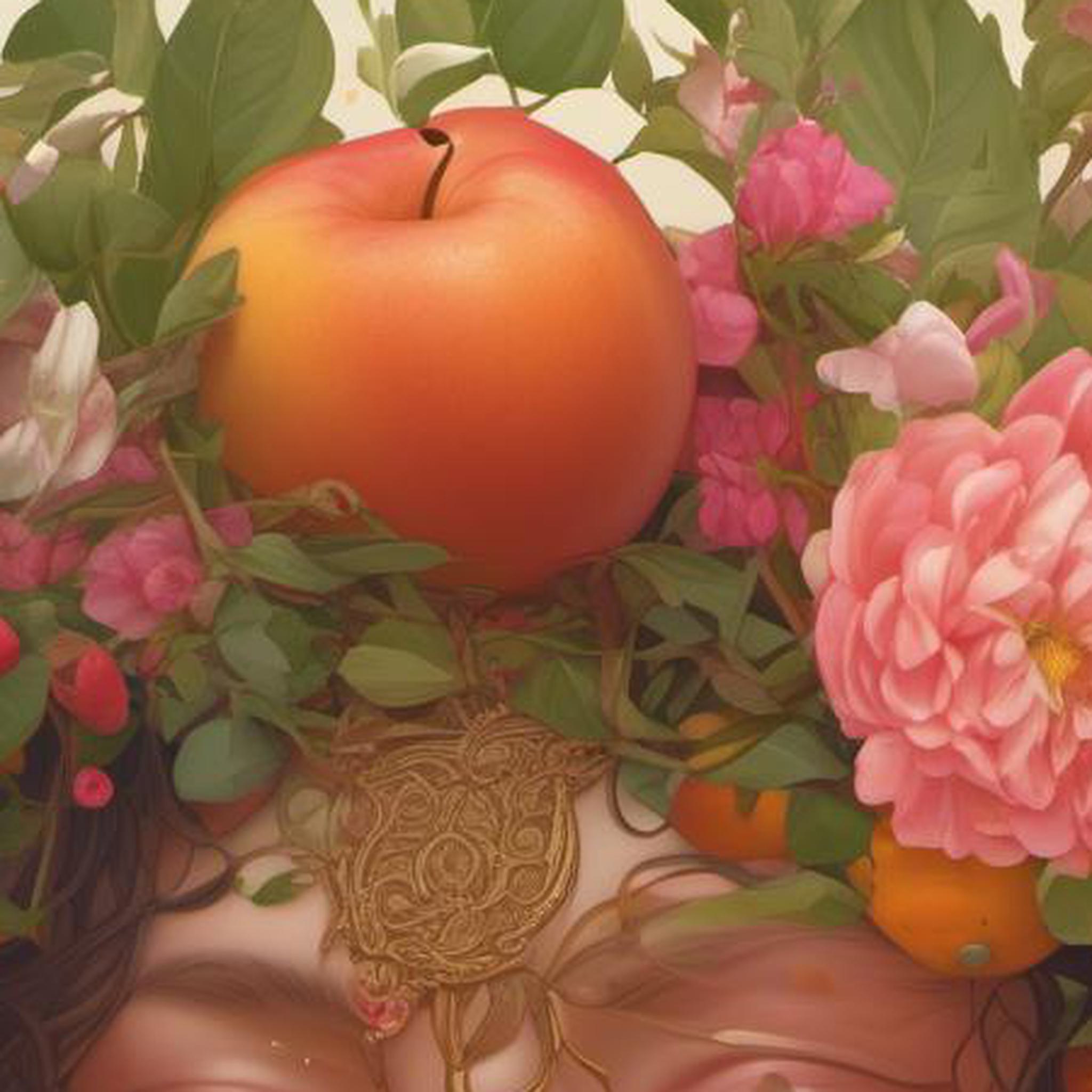}
\end{subfigure}%
\begin{subfigure}{.14\textwidth}
  \centering
  \includegraphics[width=\linewidth]{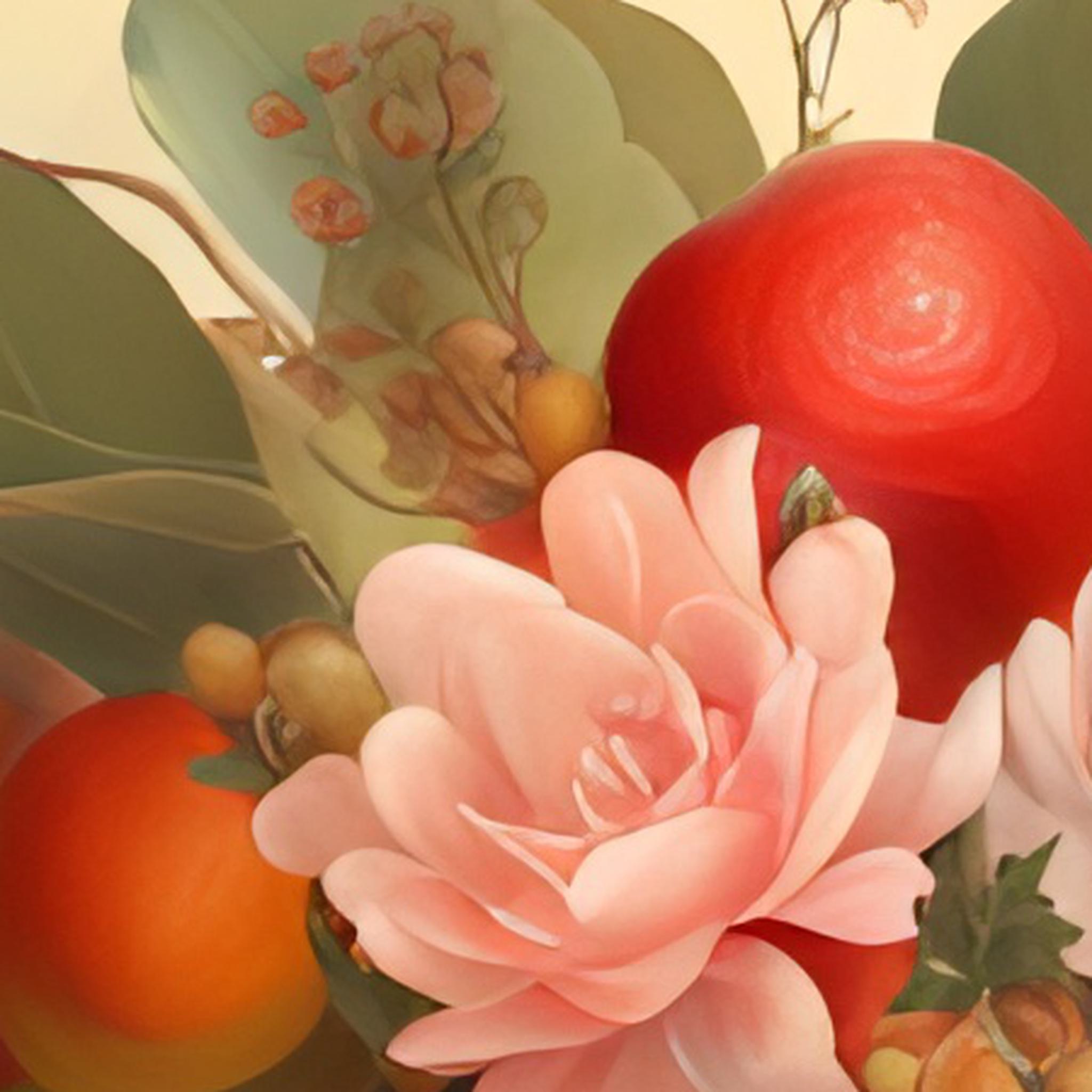}
\end{subfigure}%
\begin{subfigure}{.14\textwidth}
  \centering
  \includegraphics[width=\linewidth]{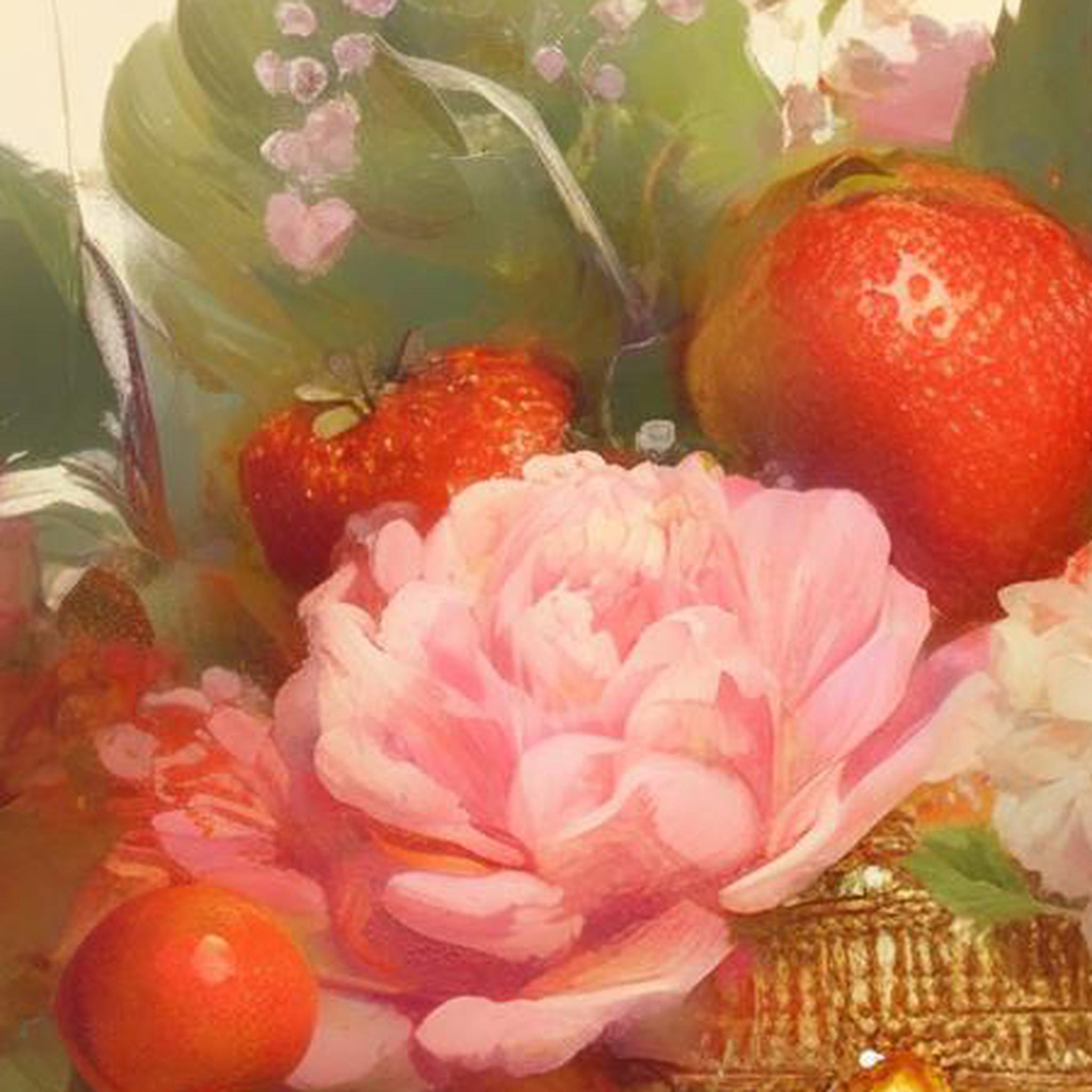}
\end{subfigure}%
\begin{subfigure}{.14\textwidth}
  \centering
  \includegraphics[width=\linewidth]{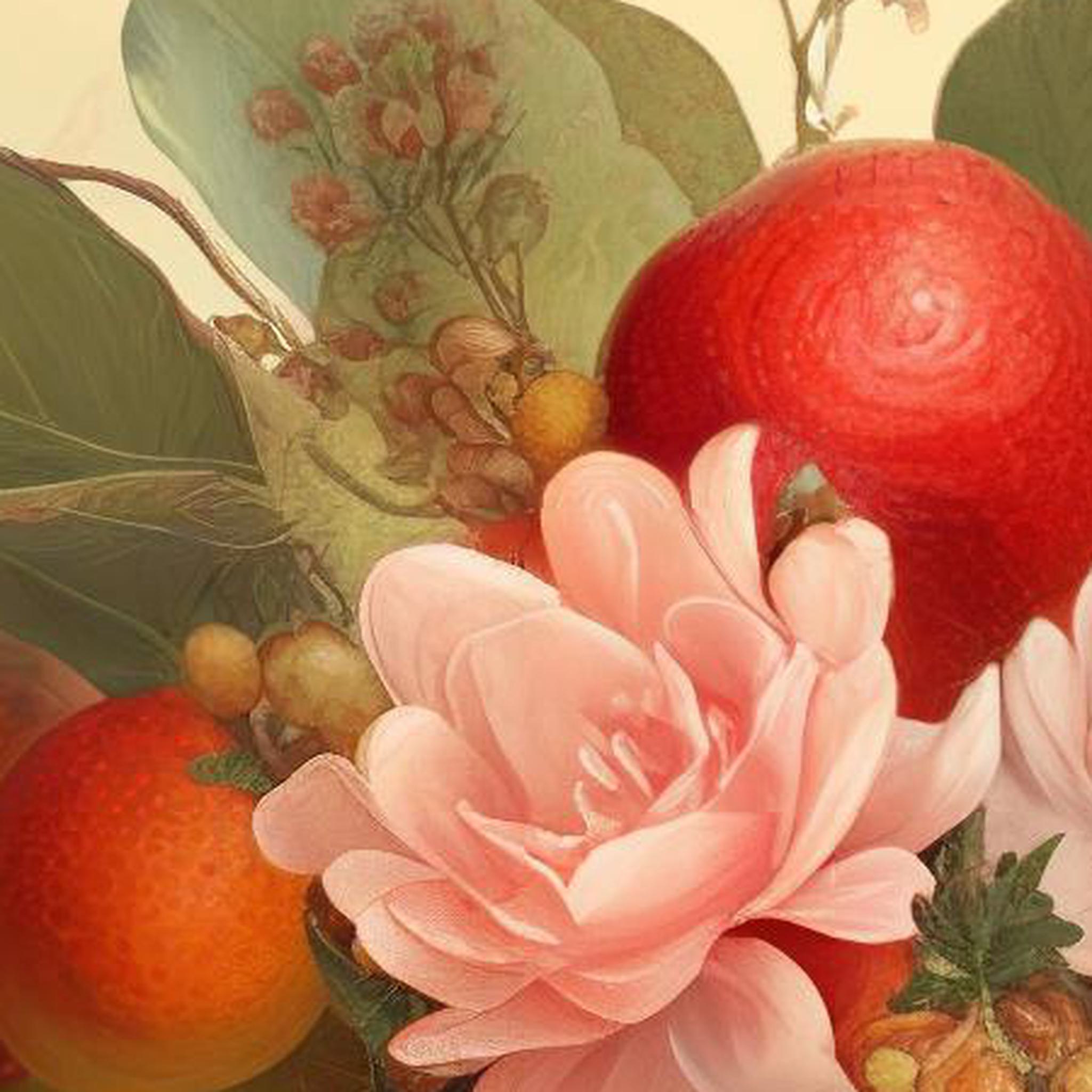}
\end{subfigure}
\begin{subfigure}{.14\textwidth}
  \centering
  \includegraphics[width=\linewidth]{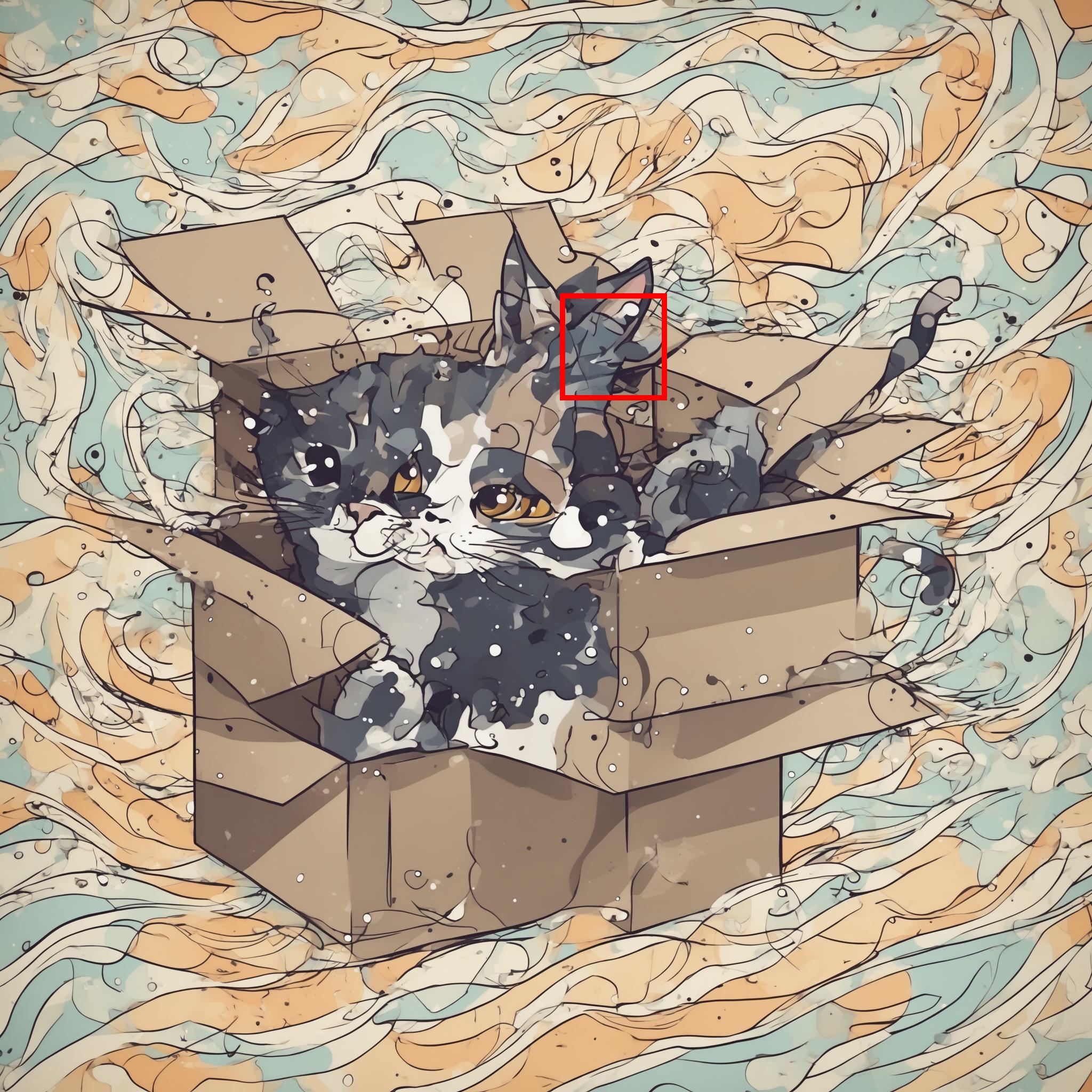}
\end{subfigure}%
\begin{subfigure}{.14\textwidth}
  \centering
  \includegraphics[width=\linewidth]{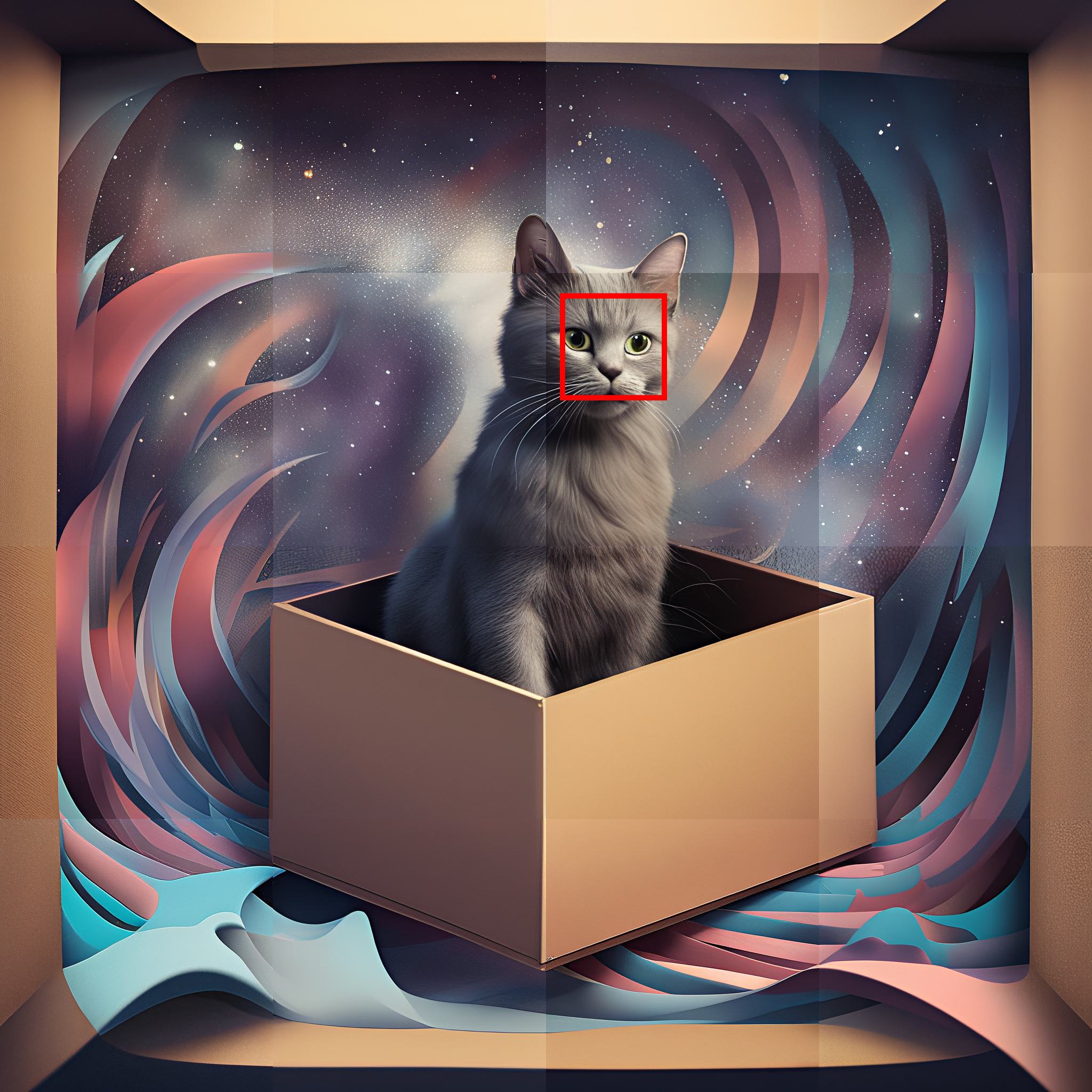}
\end{subfigure}%
\begin{subfigure}{.14\textwidth}
  \centering
  \includegraphics[width=\linewidth]{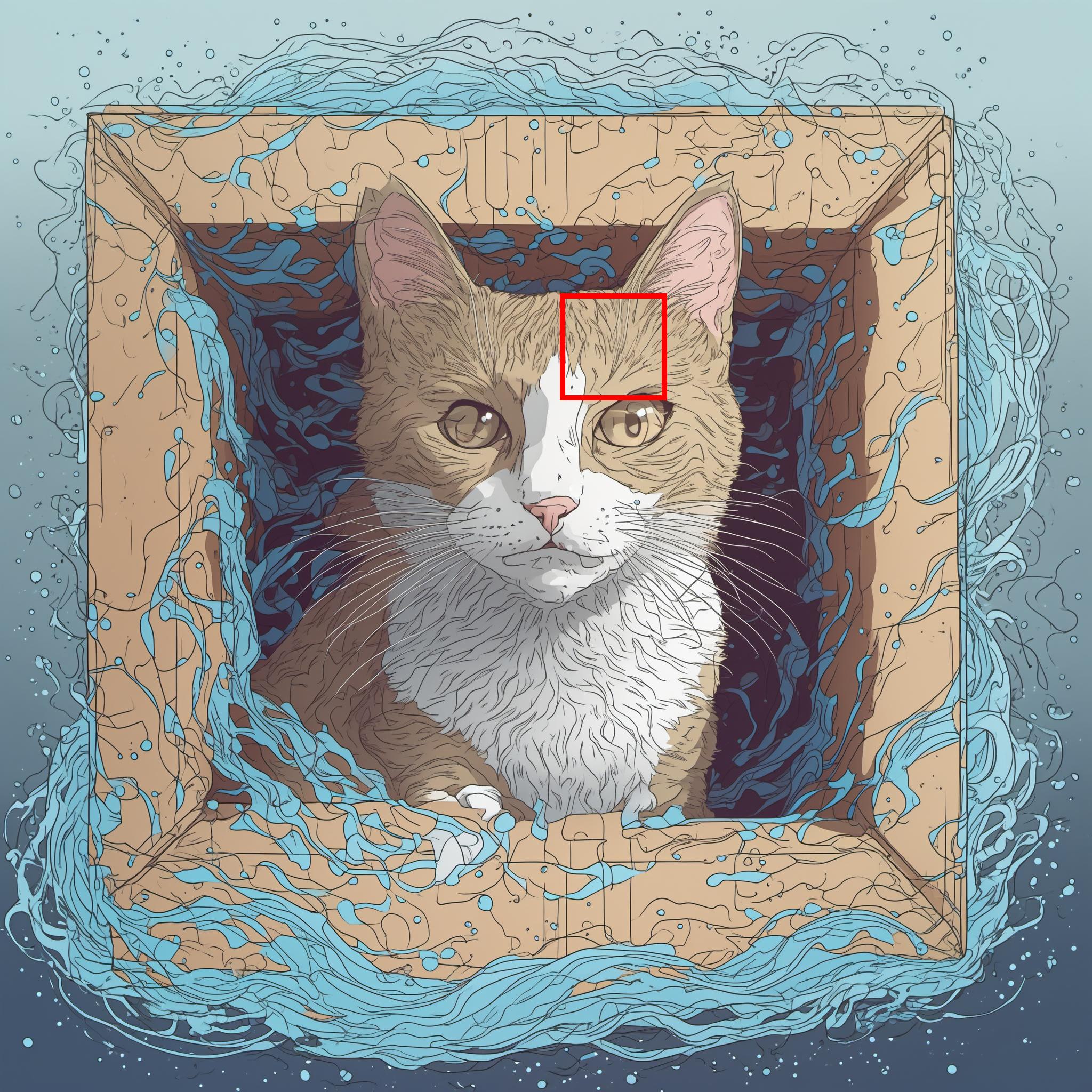}
\end{subfigure}%
\begin{subfigure}{.14\textwidth}
  \centering
  \includegraphics[width=\linewidth]{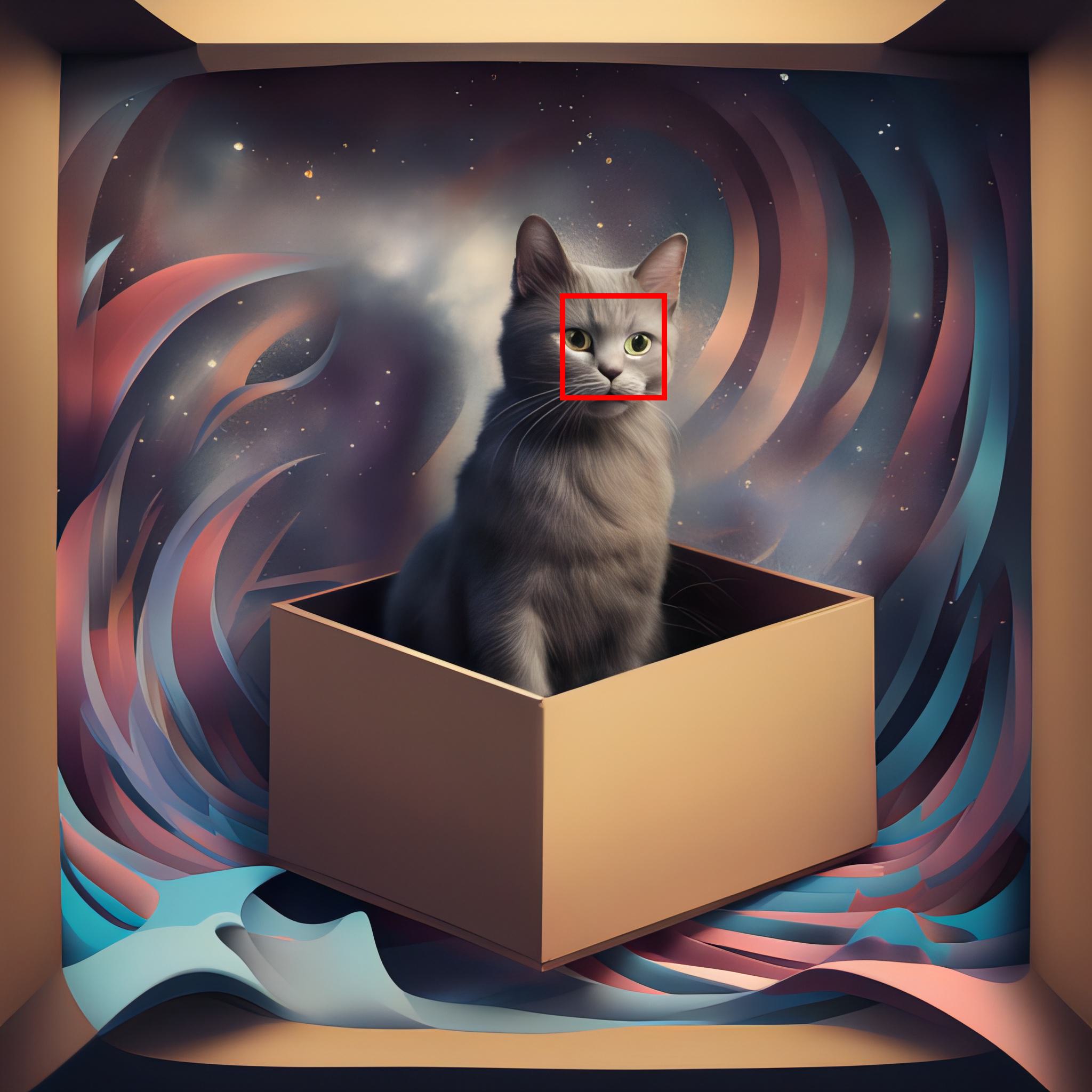}
\end{subfigure}%
\begin{subfigure}{.14\textwidth}
  \centering
  \includegraphics[width=\linewidth]{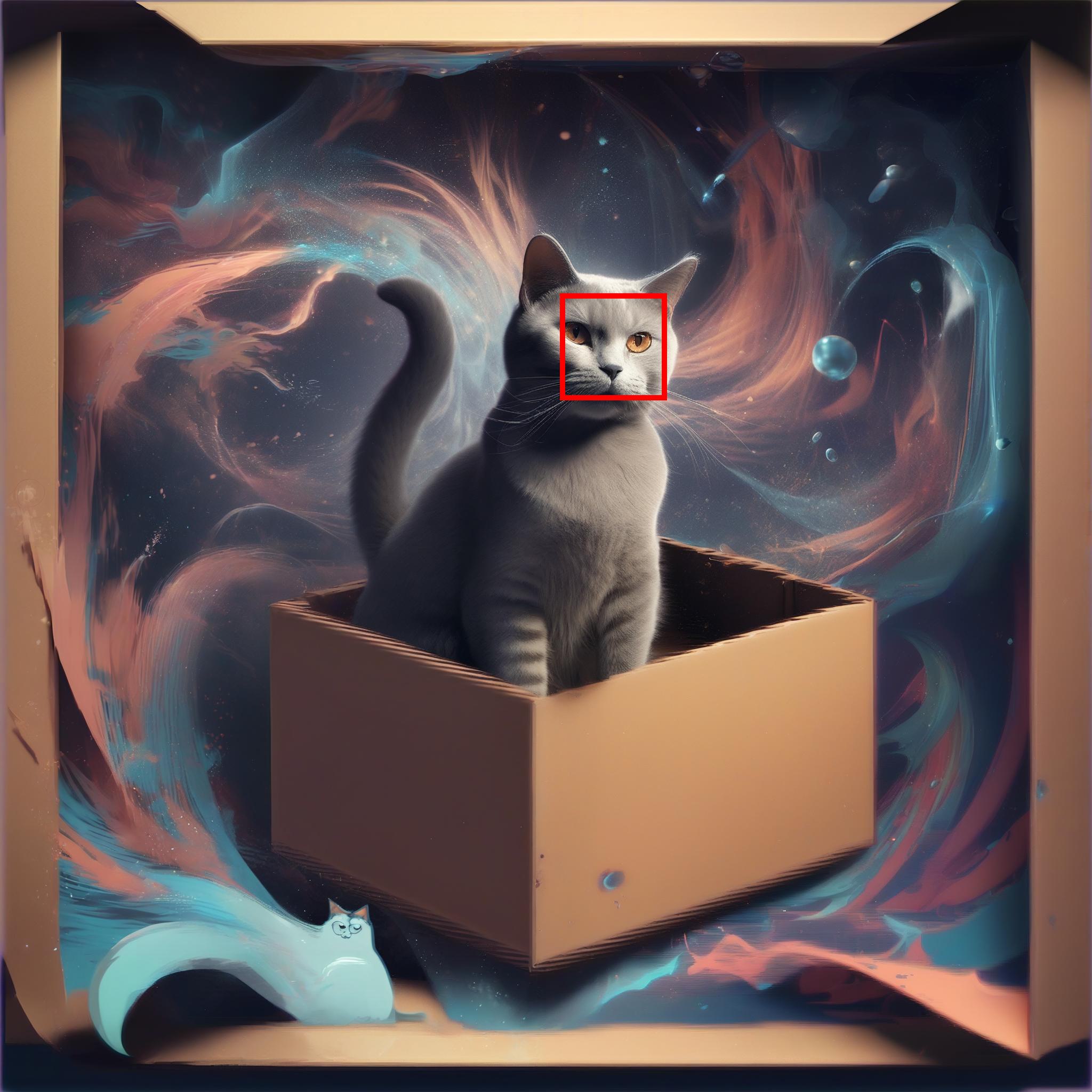}
\end{subfigure}%
\begin{subfigure}{.14\textwidth}
  \centering
  \includegraphics[width=\linewidth]{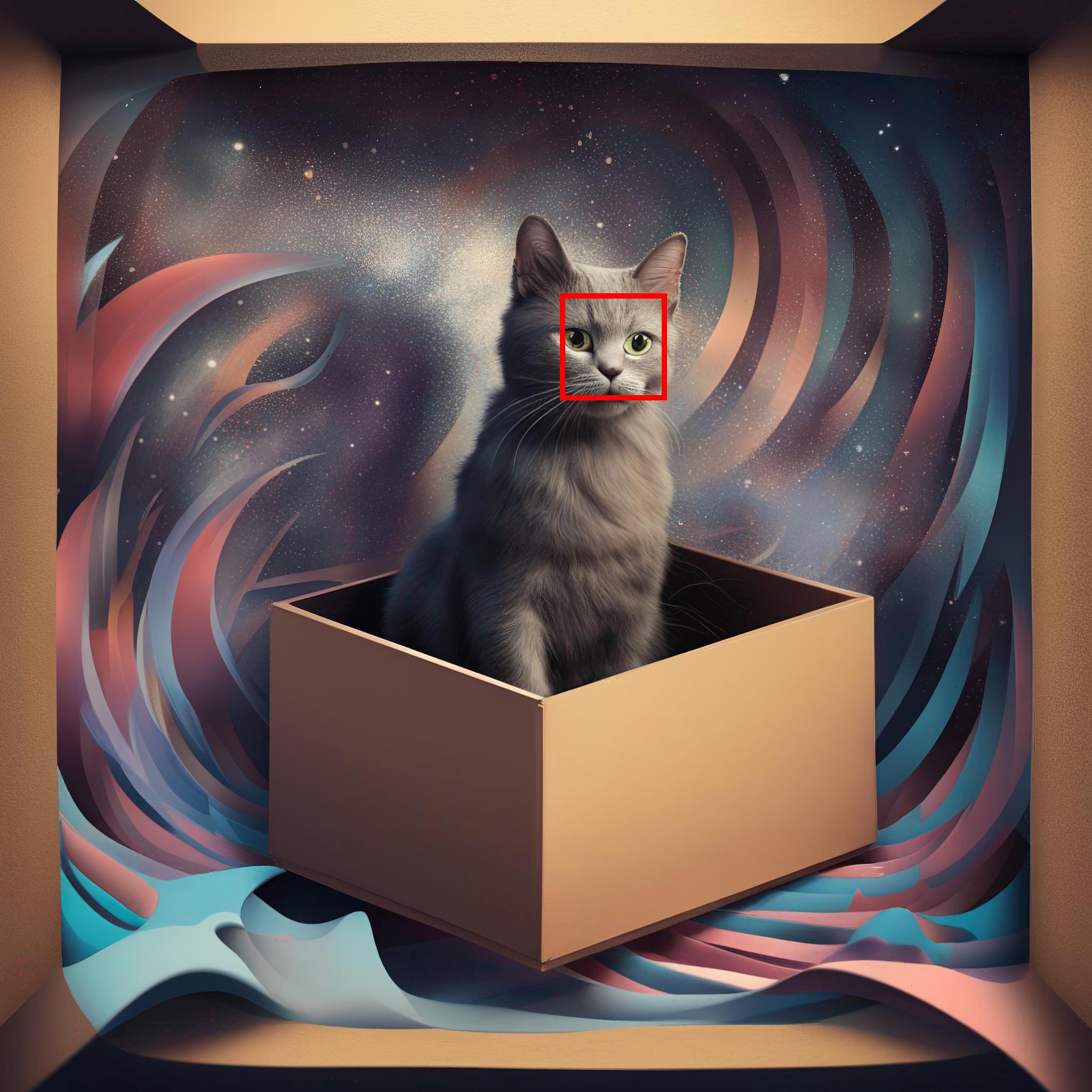}
\end{subfigure}
\begin{subfigure}{.14\textwidth}
  \centering
  \includegraphics[width=\linewidth]{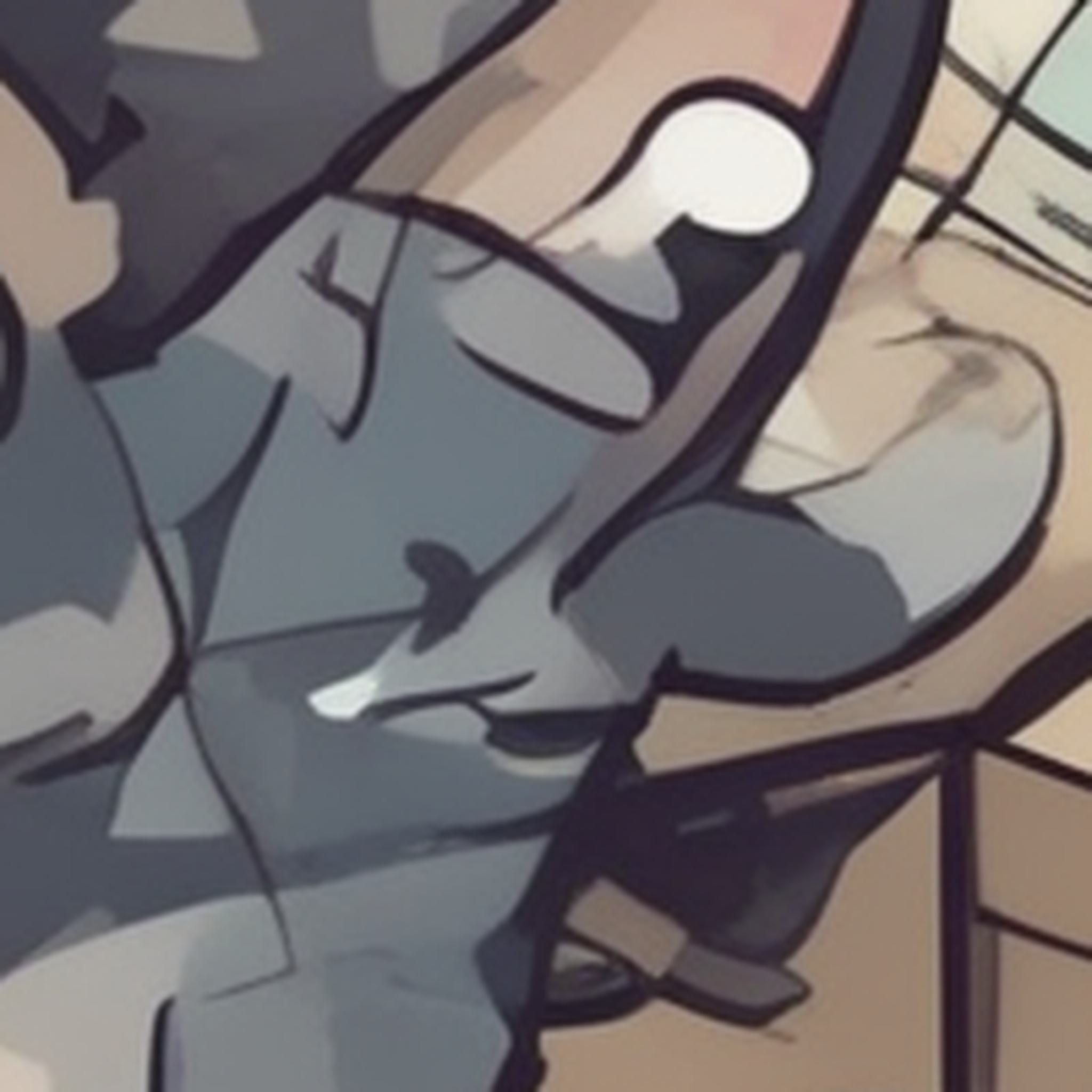}
  {\scriptsize\textbf{Direct}}
\end{subfigure}%
\begin{subfigure}{.14\textwidth}
  \centering
  \includegraphics[width=\linewidth]{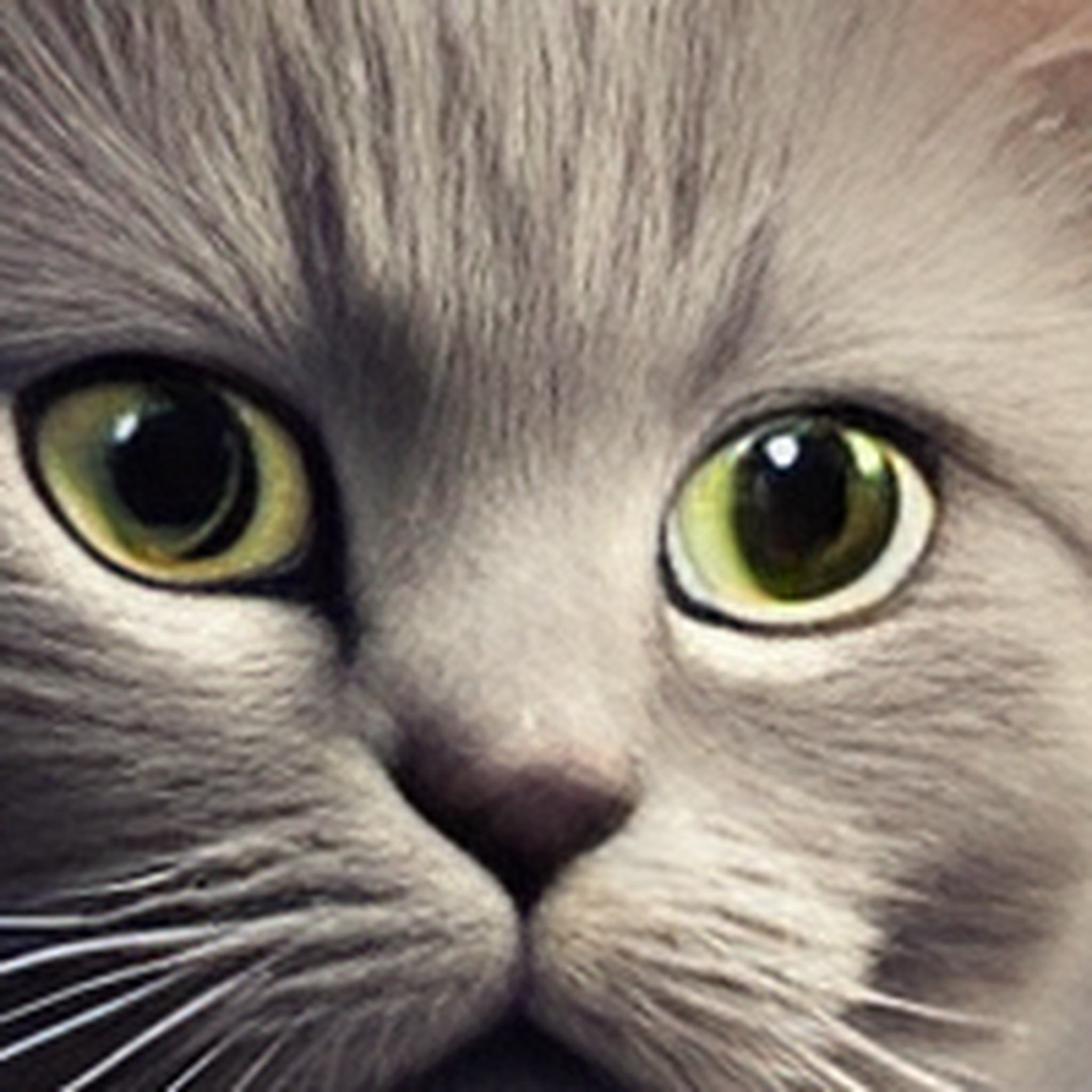}
  {\scriptsize\textbf{Patch}}
\end{subfigure}%
\begin{subfigure}{.14\textwidth}
  \centering
  \includegraphics[width=\linewidth]{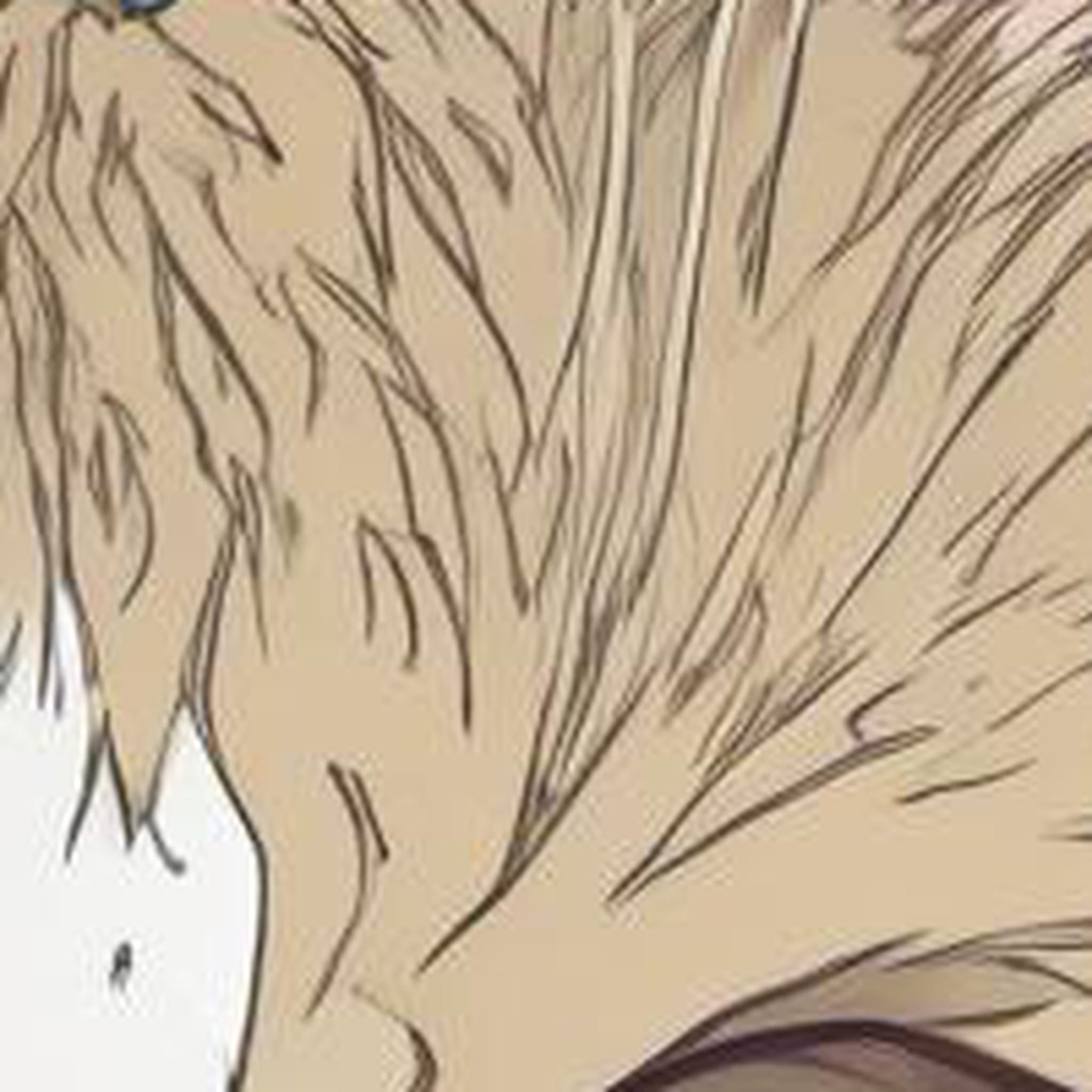}
  {\scriptsize\textbf{Scalecrafter}}
\end{subfigure}%
\begin{subfigure}{.14\textwidth}
  \centering
  \includegraphics[width=\linewidth]{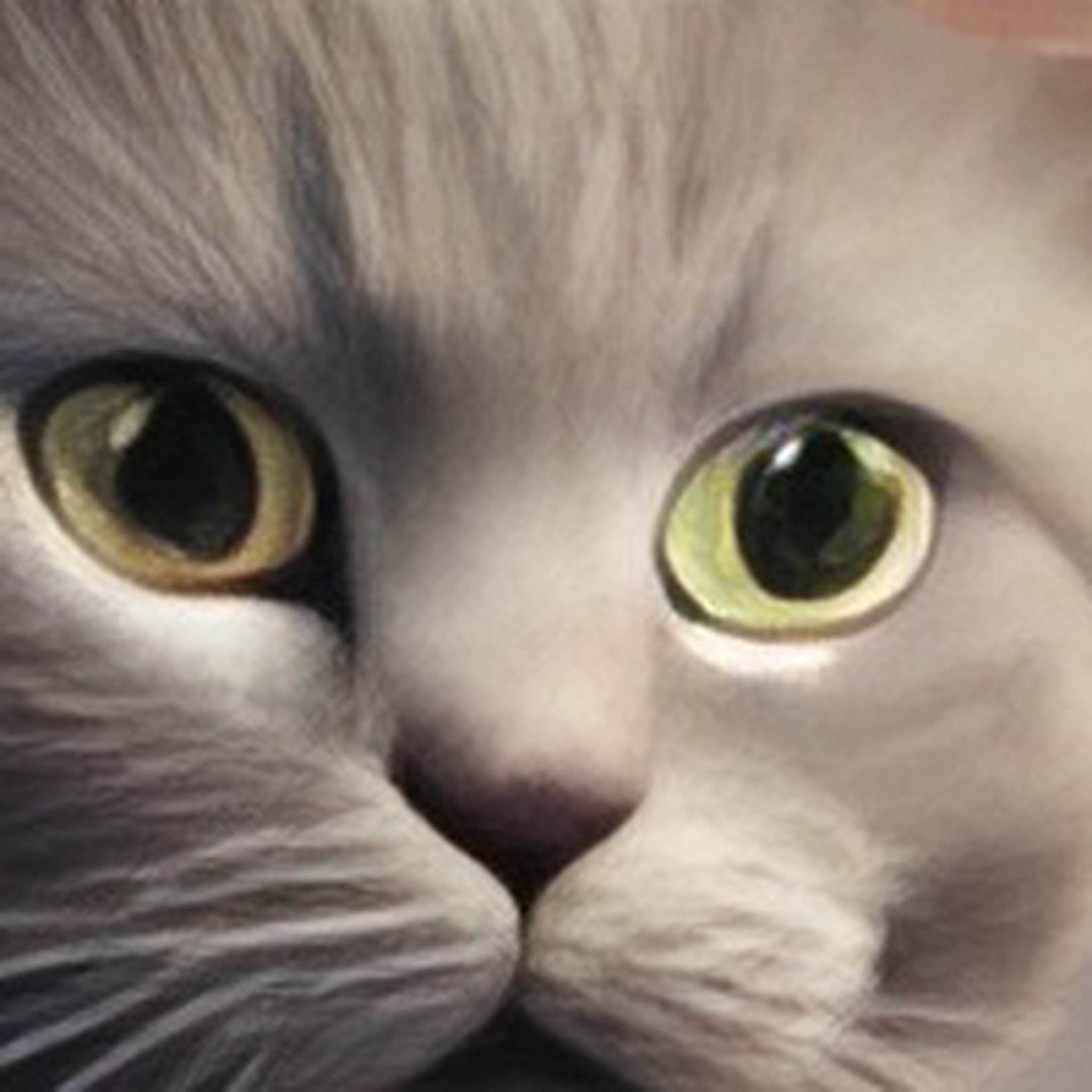}
  {\scriptsize\textbf{BSRGAN}}
\end{subfigure}%
\begin{subfigure}{.14\textwidth}
  \centering
  \includegraphics[width=\linewidth]{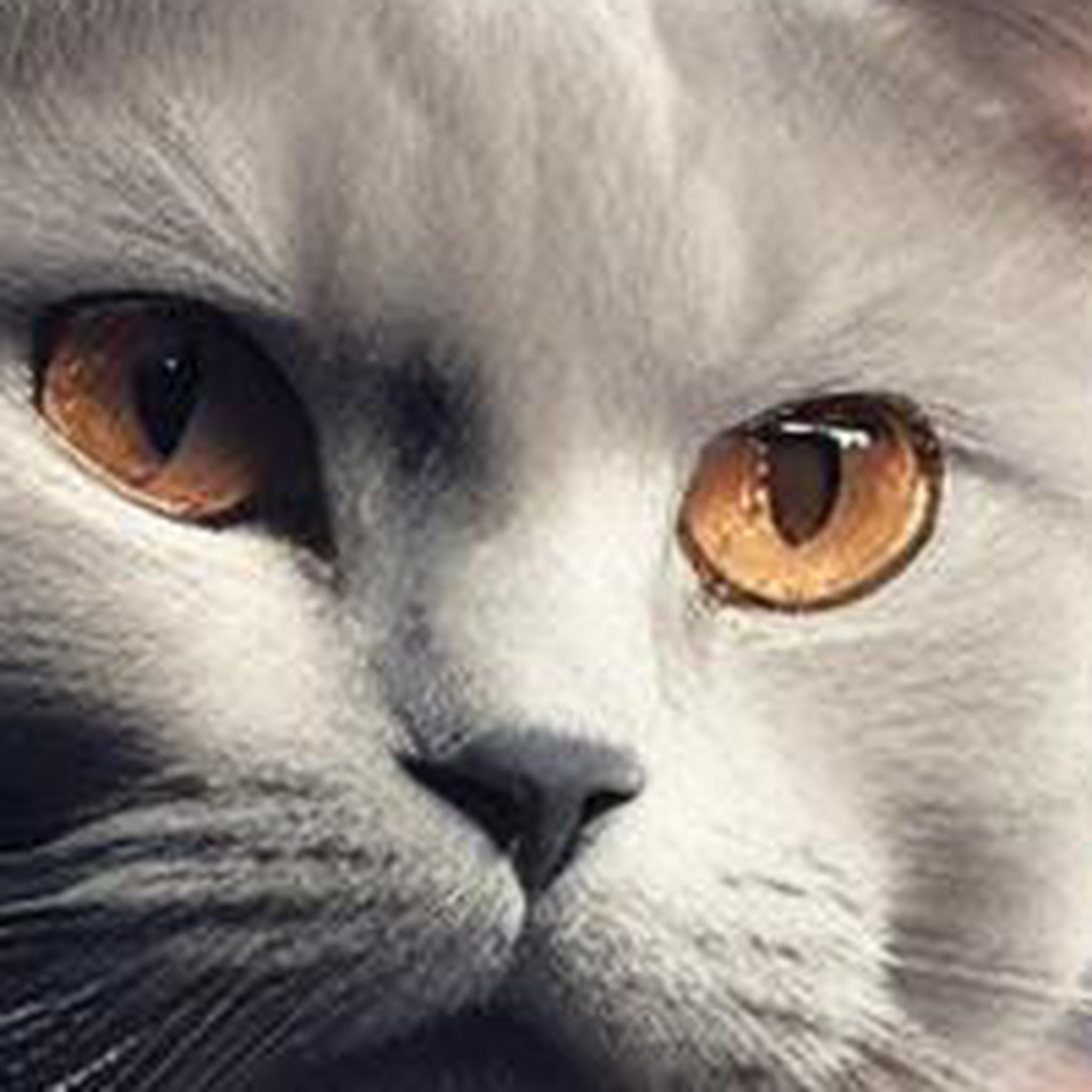}
  {\scriptsize\textbf{Demofusion}}
\end{subfigure}%
\begin{subfigure}{.14\textwidth}
  \centering
  \includegraphics[width=\linewidth]{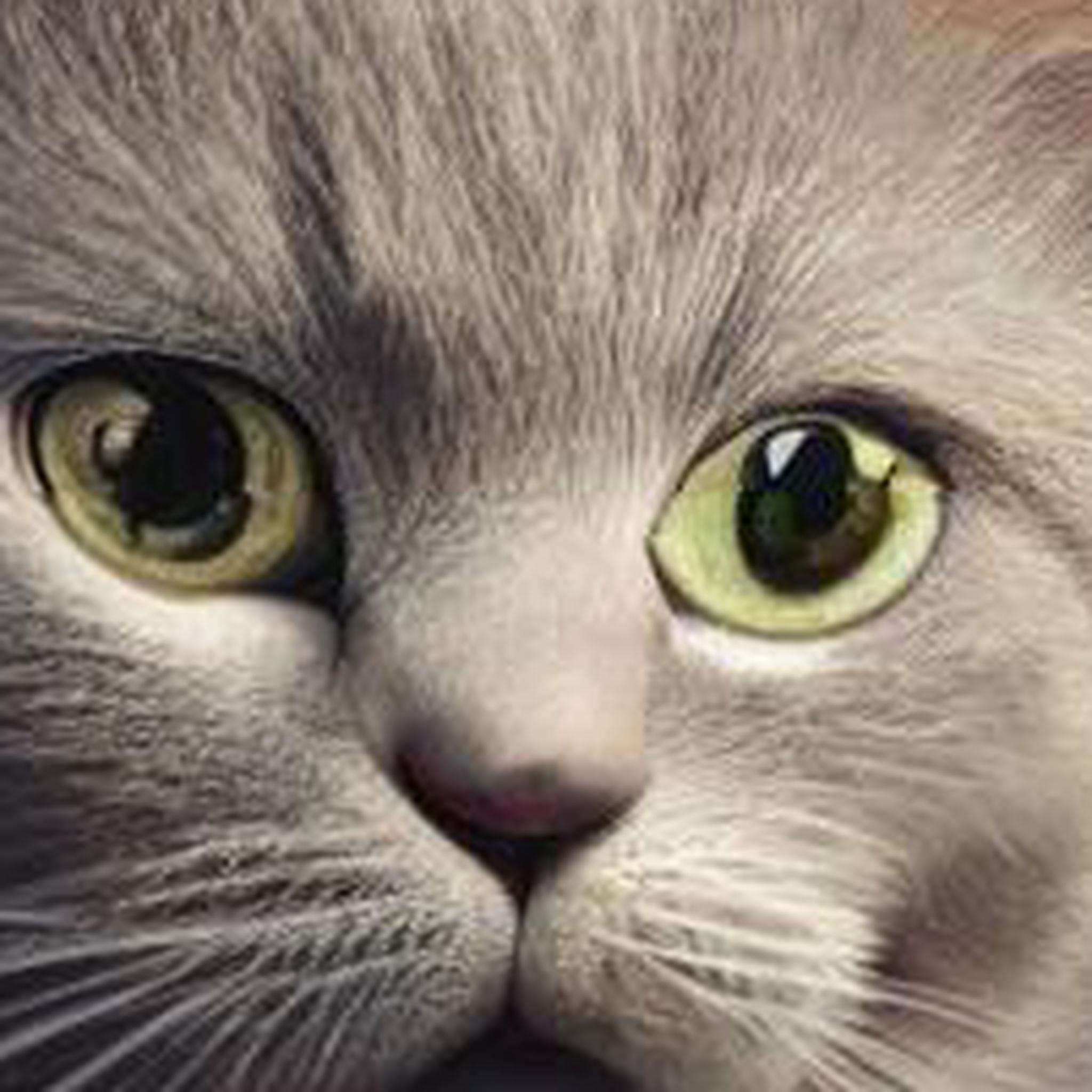}
  {\scriptsize\textbf{Inf-DiT}}
\end{subfigure}%
\captionsetup{skip=2pt}
\caption{Qualitative comparison of different methods in detail at $2048\times2048$ resolution}
\label{fig:images}
\vspace{-4mm}
\end{figure}
\begin{figure}[!h]
\centering
\begin{subfigure}{.16\textwidth}
  \centering
  \includegraphics[width=\linewidth]{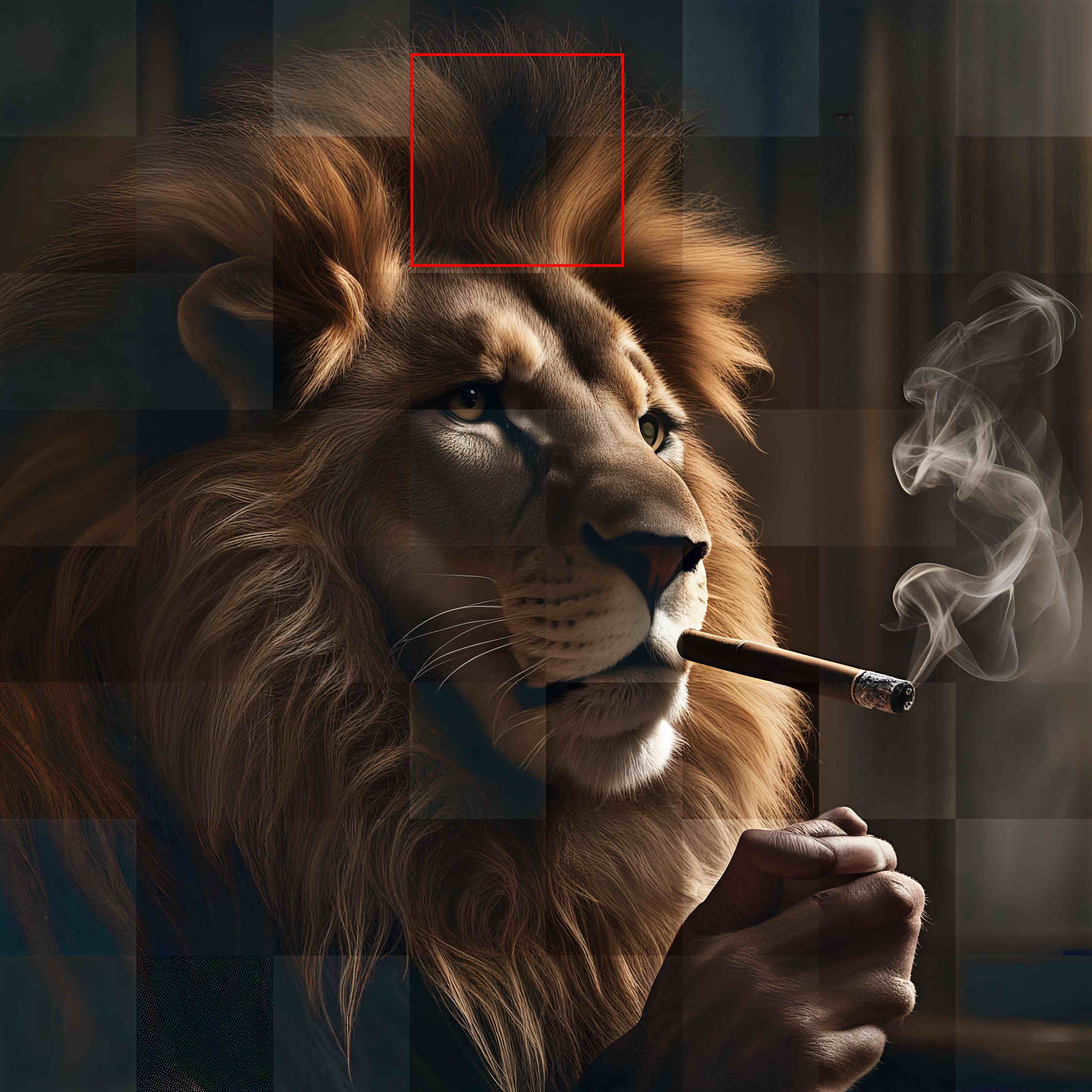}
\end{subfigure}%
\begin{subfigure}{.16\textwidth}
  \centering
  \includegraphics[width=\linewidth]{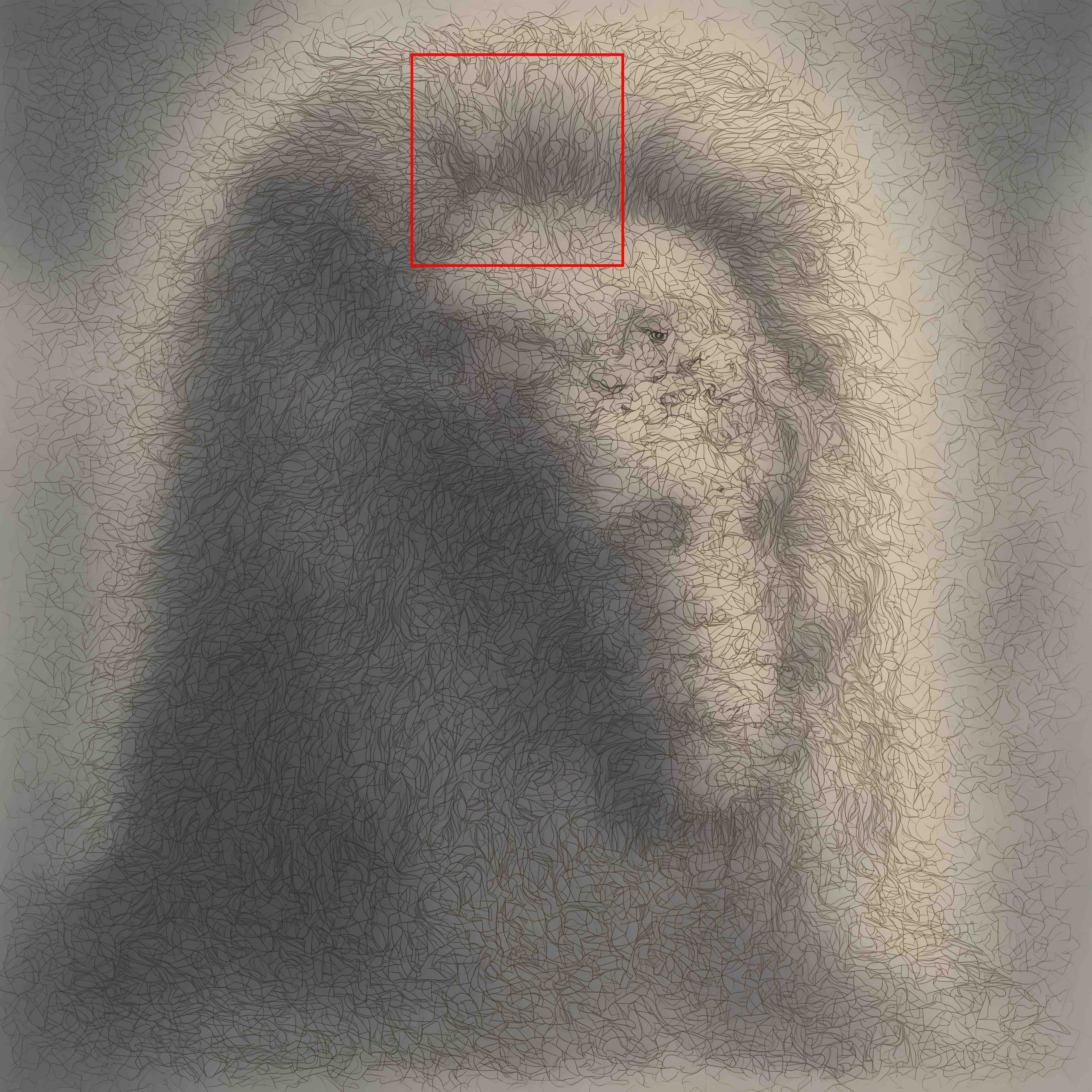}
\end{subfigure}%
\begin{subfigure}{.16\textwidth}
  \centering
  \includegraphics[width=\linewidth]{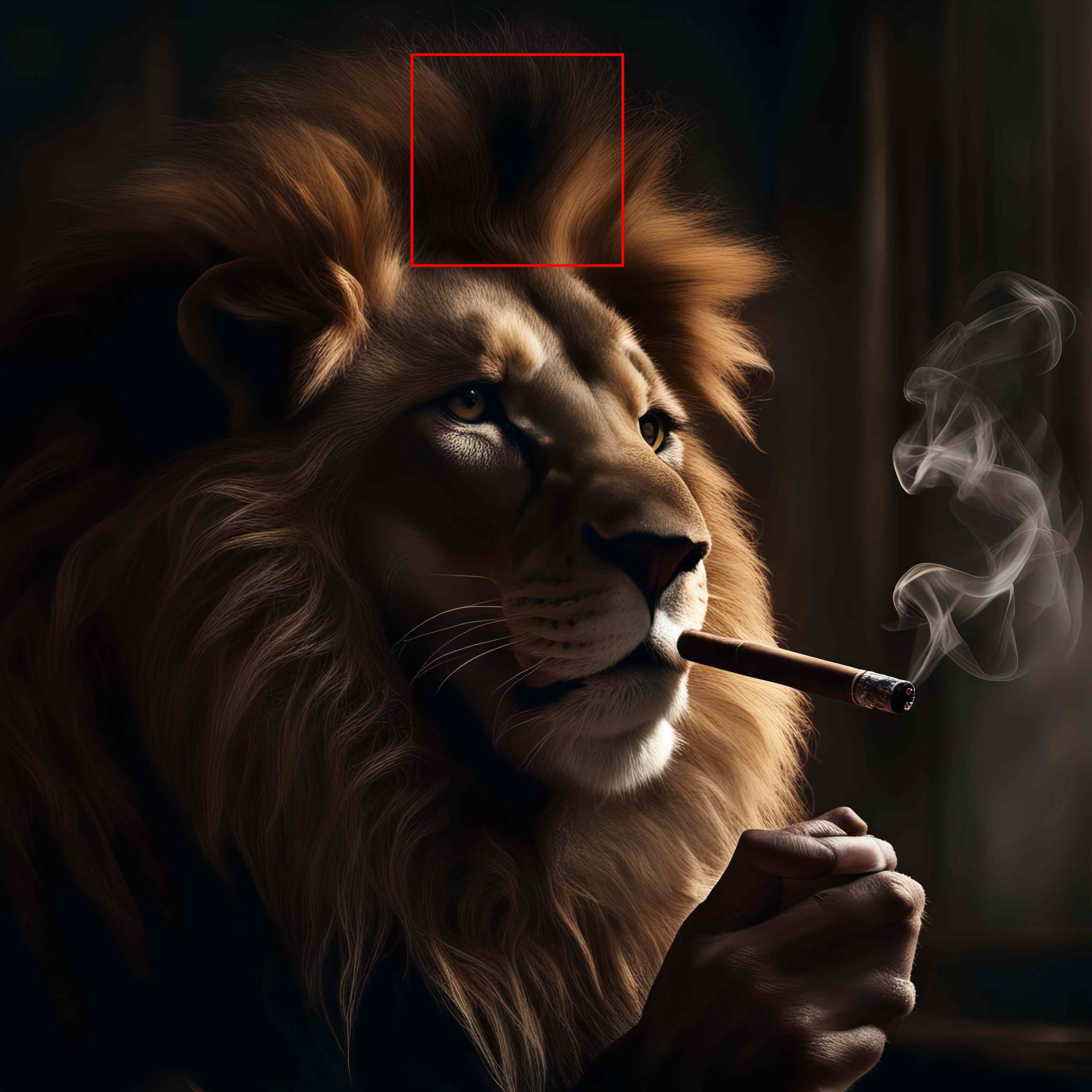}
\end{subfigure}%
\begin{subfigure}{.16\textwidth}
  \centering
  \includegraphics[width=\linewidth]{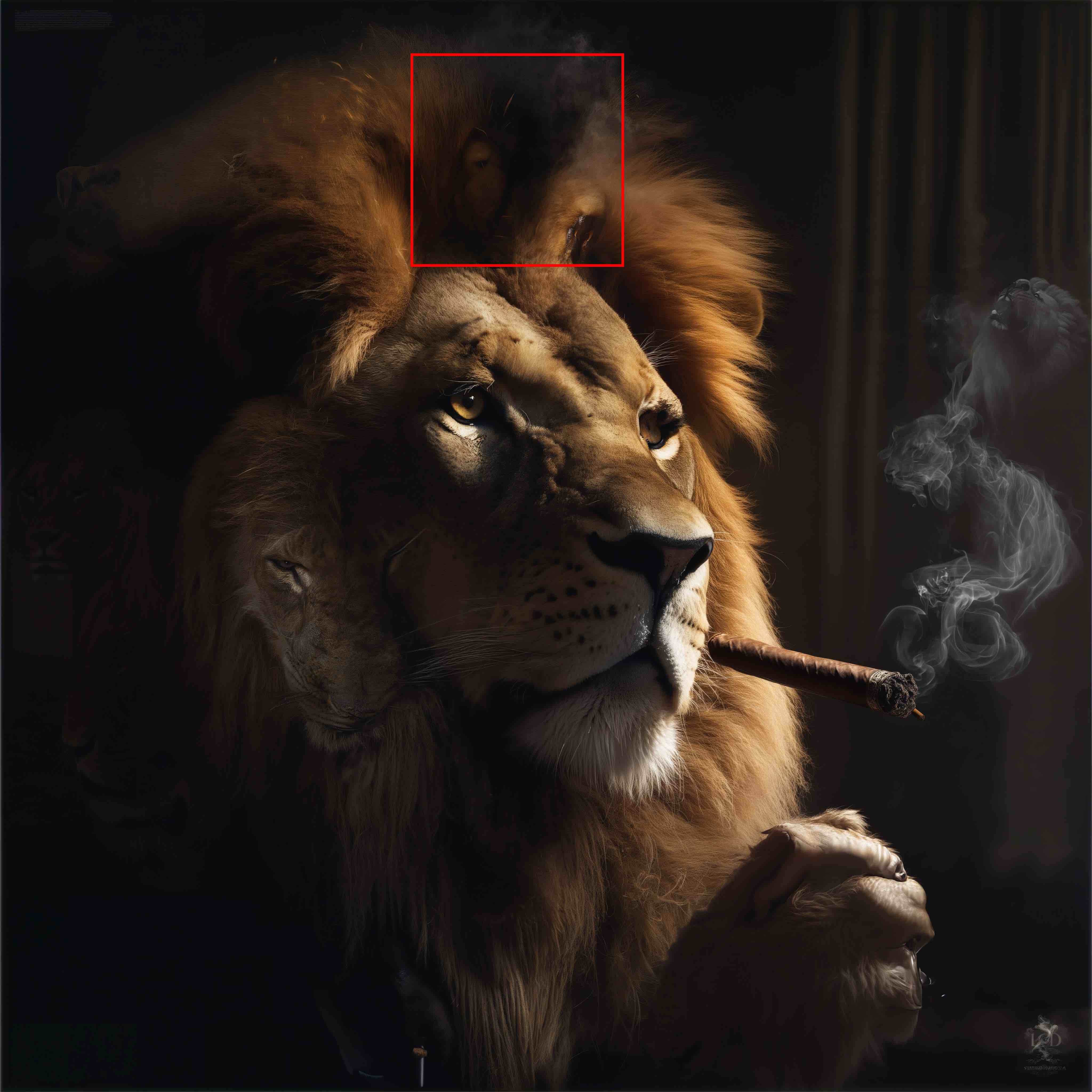}
\end{subfigure}%
\begin{subfigure}{.16\textwidth}
  \centering
  \includegraphics[width=\linewidth]{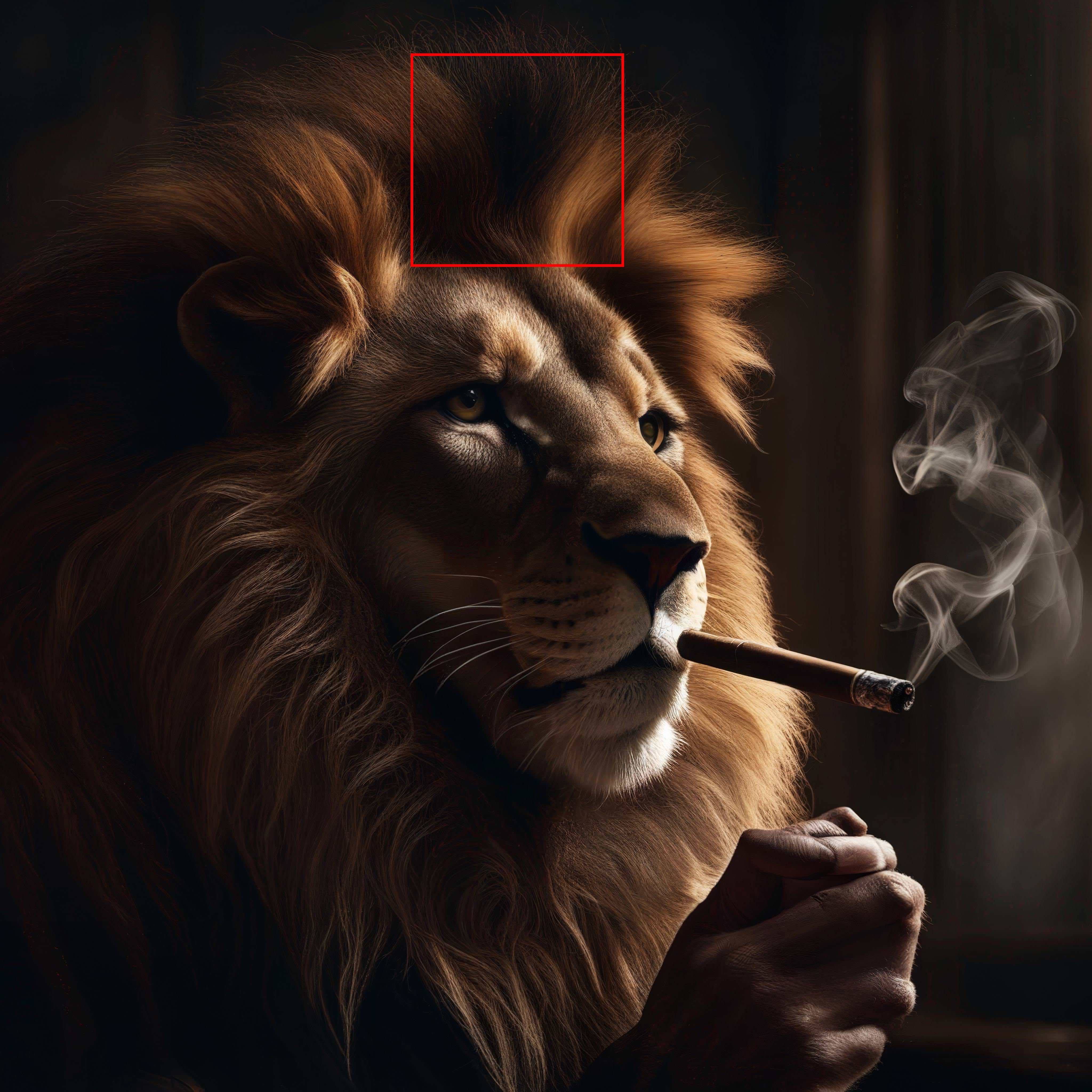}
\end{subfigure}
\begin{subfigure}{.16\textwidth}
  \centering
  \includegraphics[width=\linewidth]{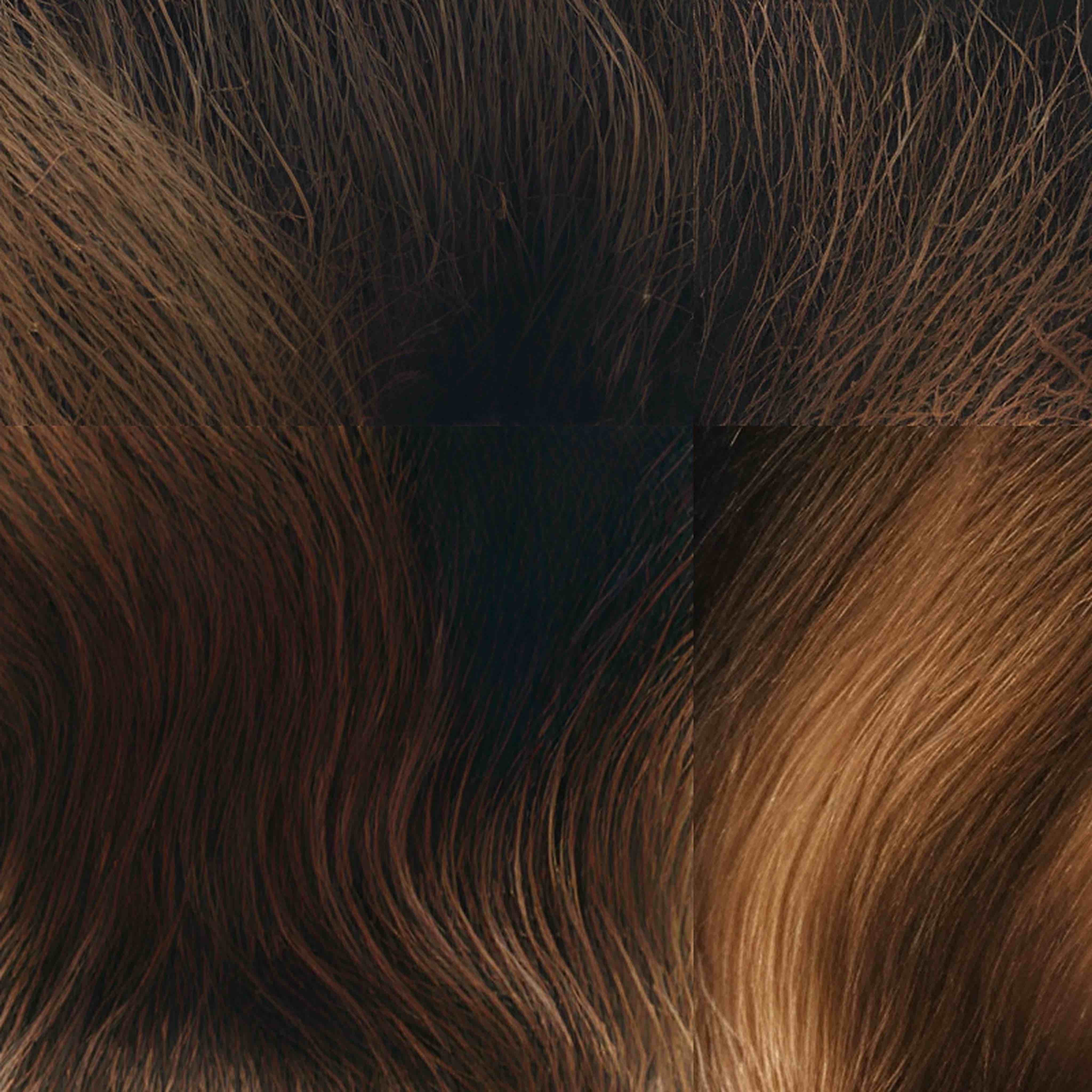}
\end{subfigure}%
\begin{subfigure}{.16\textwidth}
  \centering
  \includegraphics[width=\linewidth]{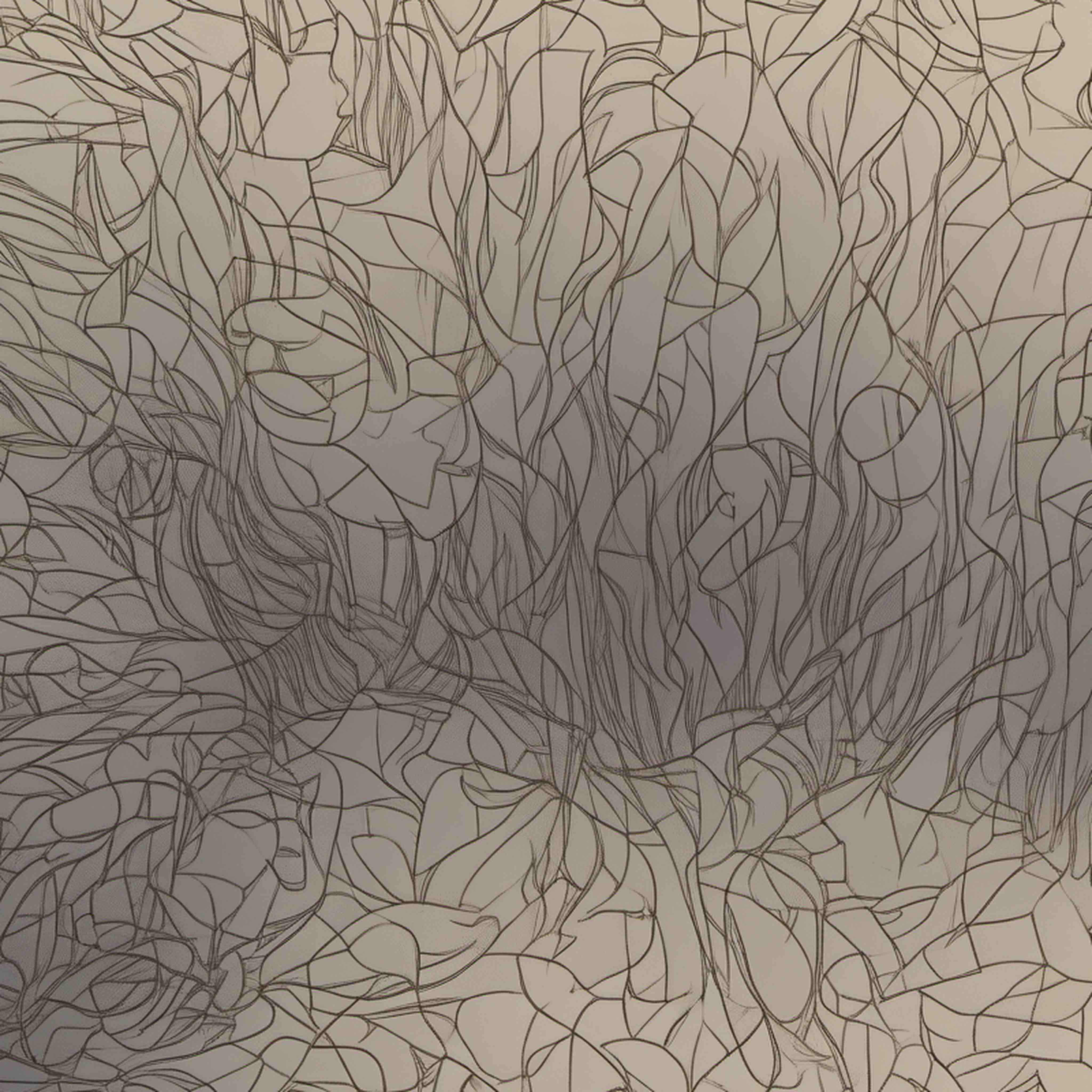}
\end{subfigure}%
\begin{subfigure}{.16\textwidth}
  \centering
  \includegraphics[width=\linewidth]{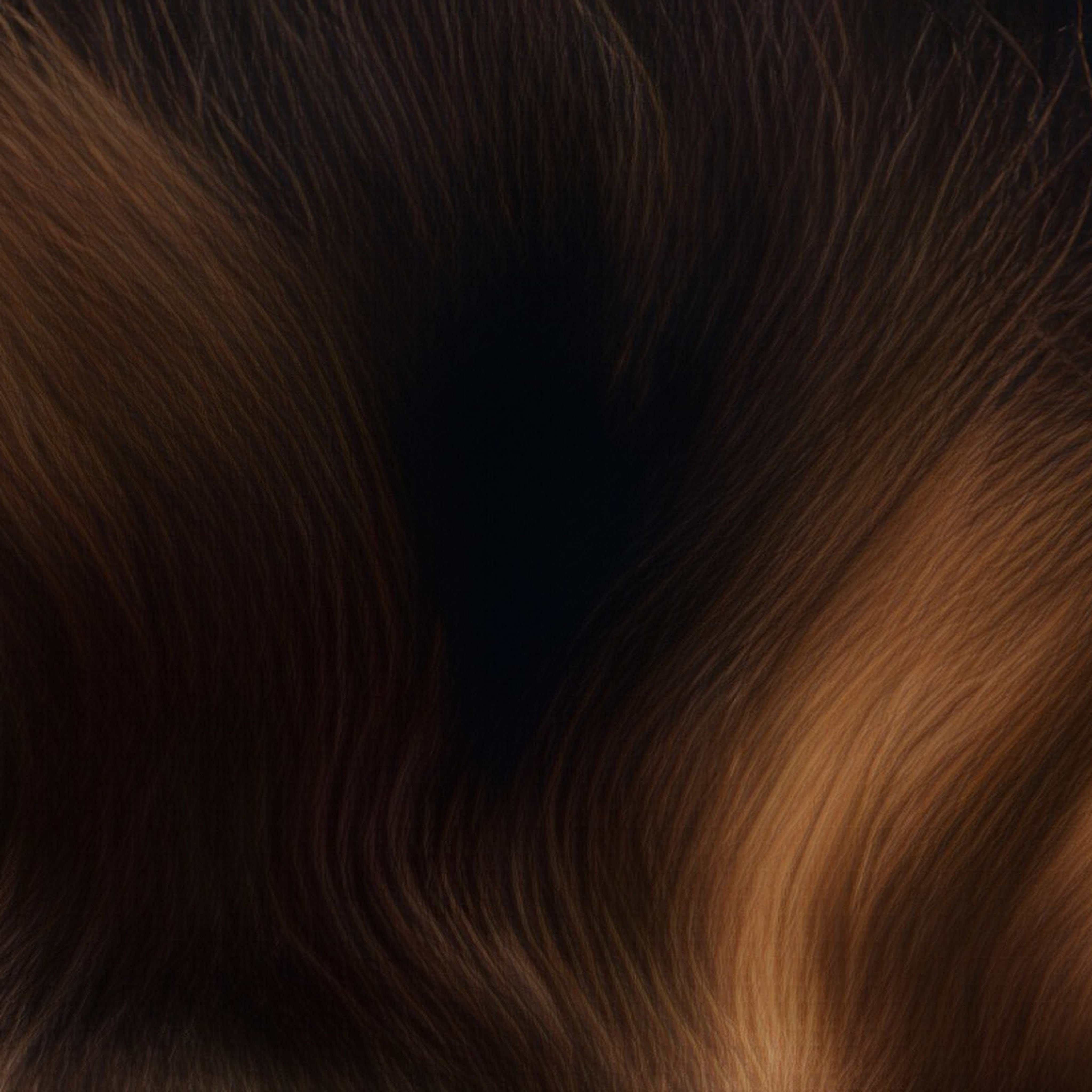}
\end{subfigure}%
\begin{subfigure}{.16\textwidth}
  \centering
  \includegraphics[width=\linewidth]{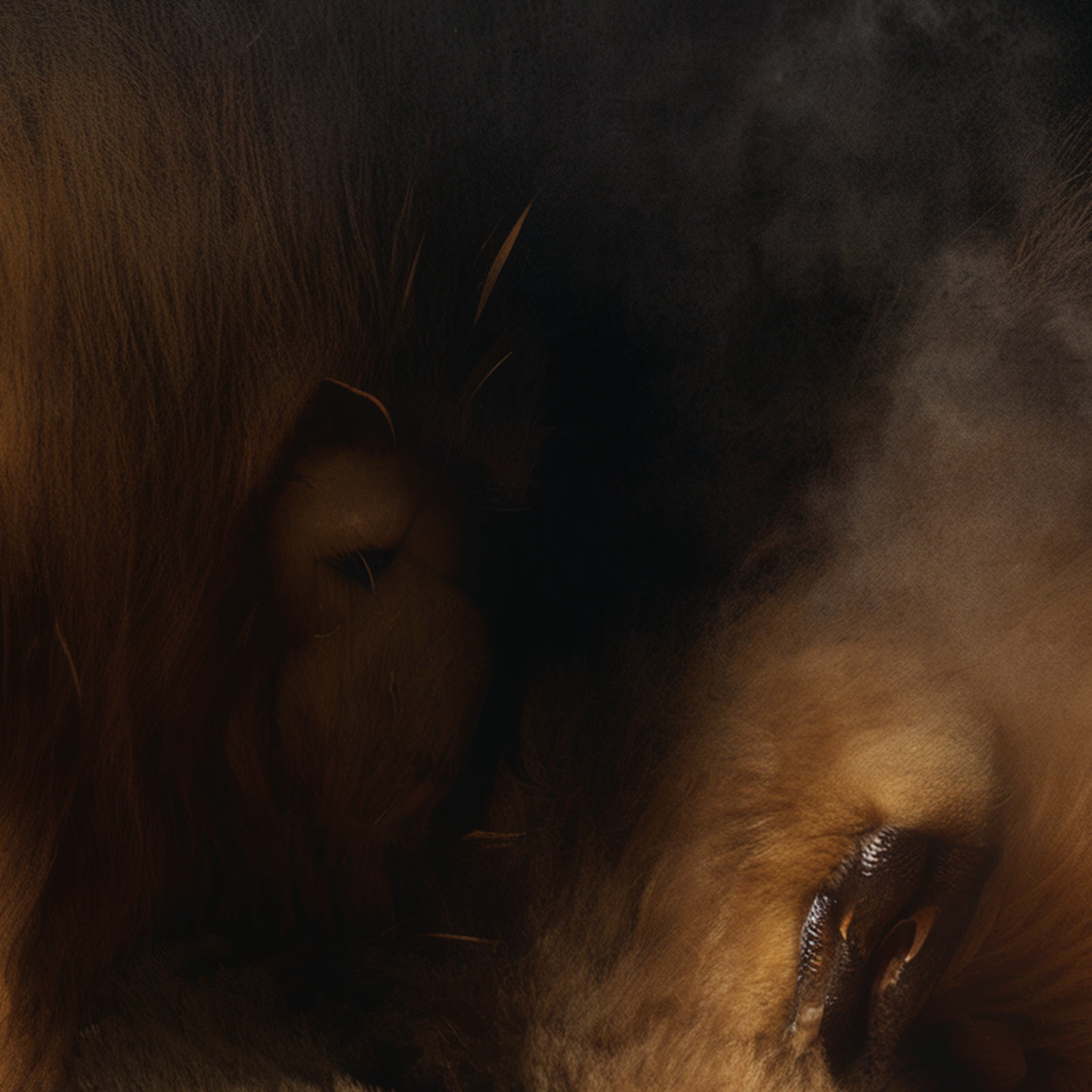}
\end{subfigure}%
\begin{subfigure}{.16\textwidth}
  \centering
  \includegraphics[width=\linewidth]{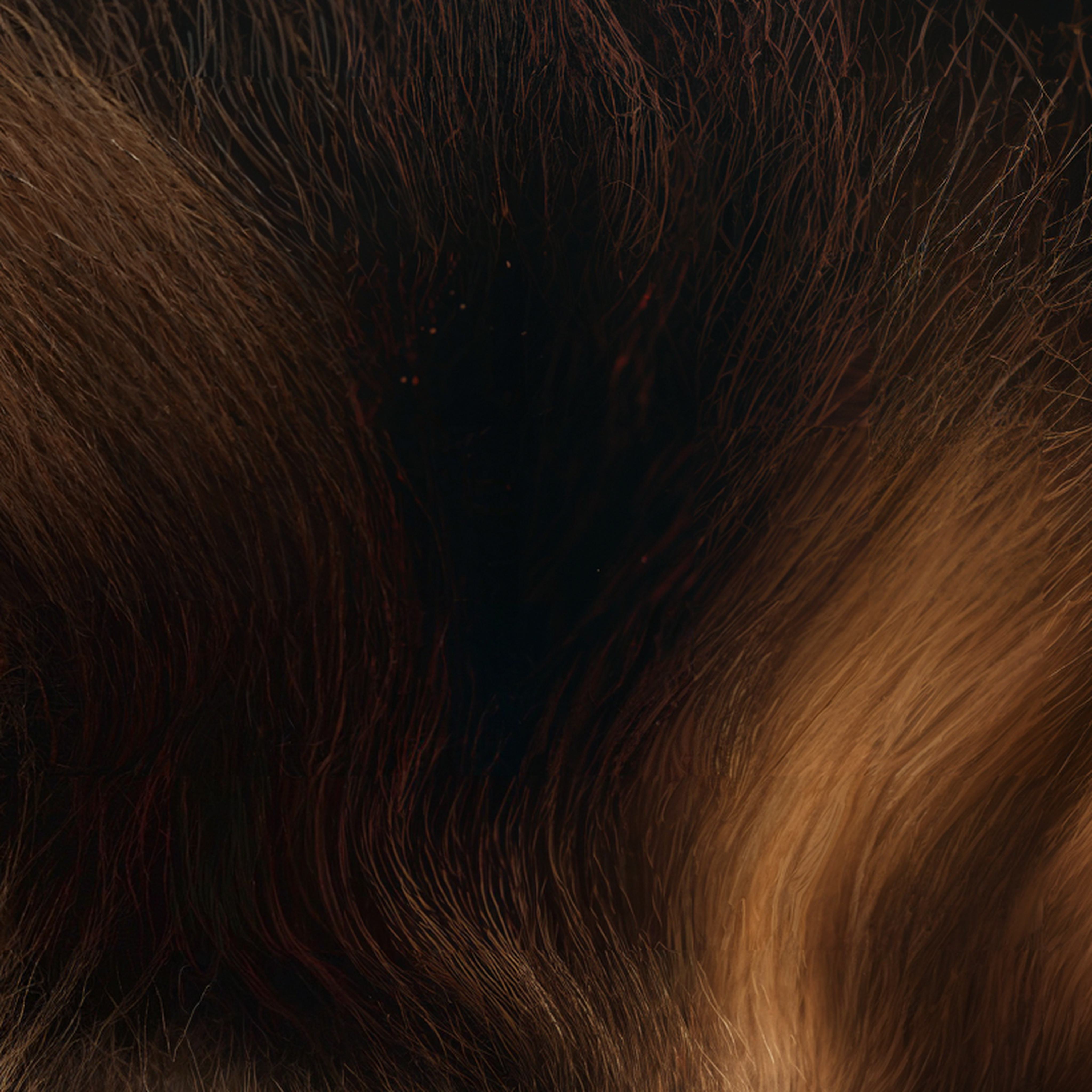}
\end{subfigure}
\begin{subfigure}{.16\textwidth}
  \centering
  \includegraphics[width=\linewidth]{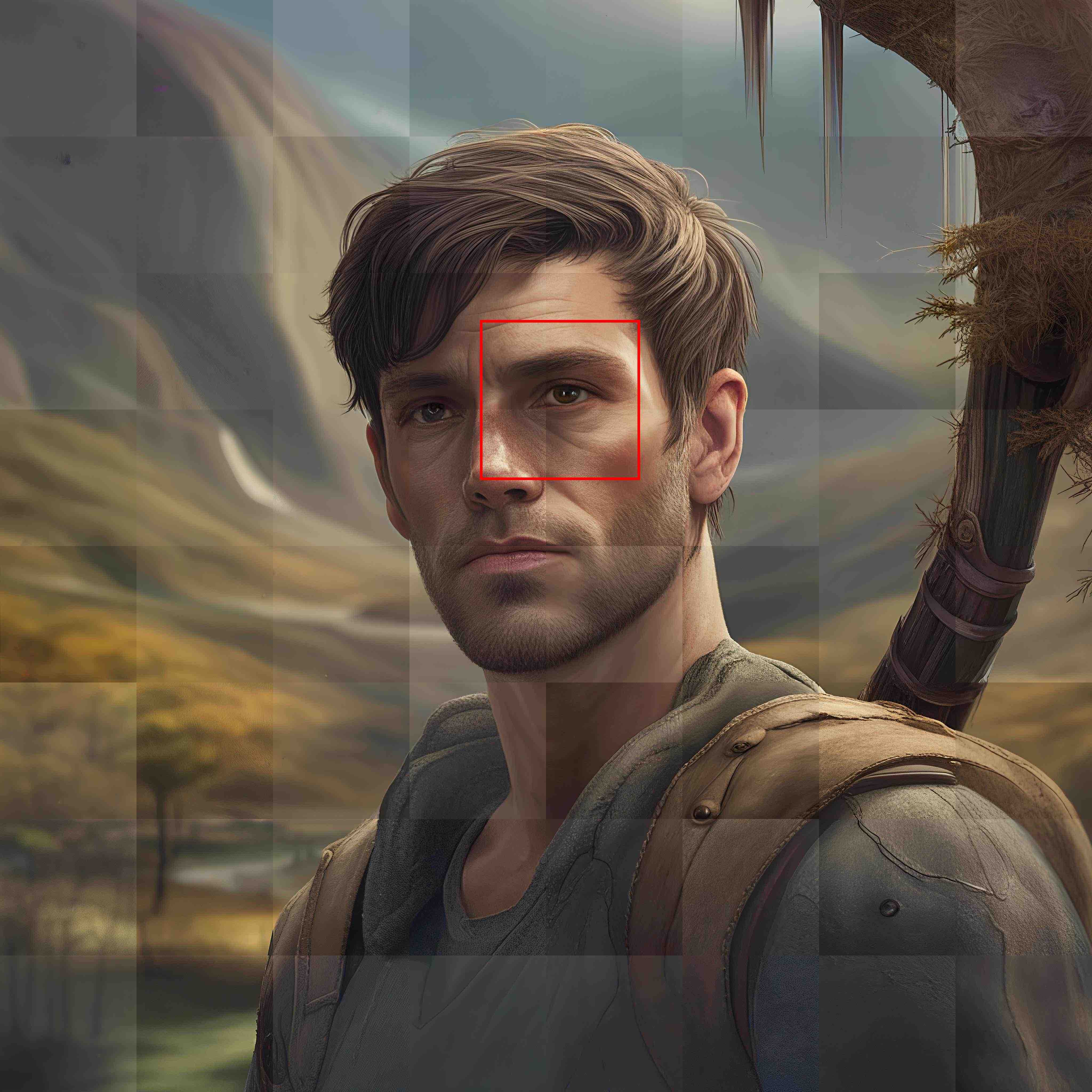}
\end{subfigure}%
\begin{subfigure}{.16\textwidth}
  \centering
  \includegraphics[width=\linewidth]{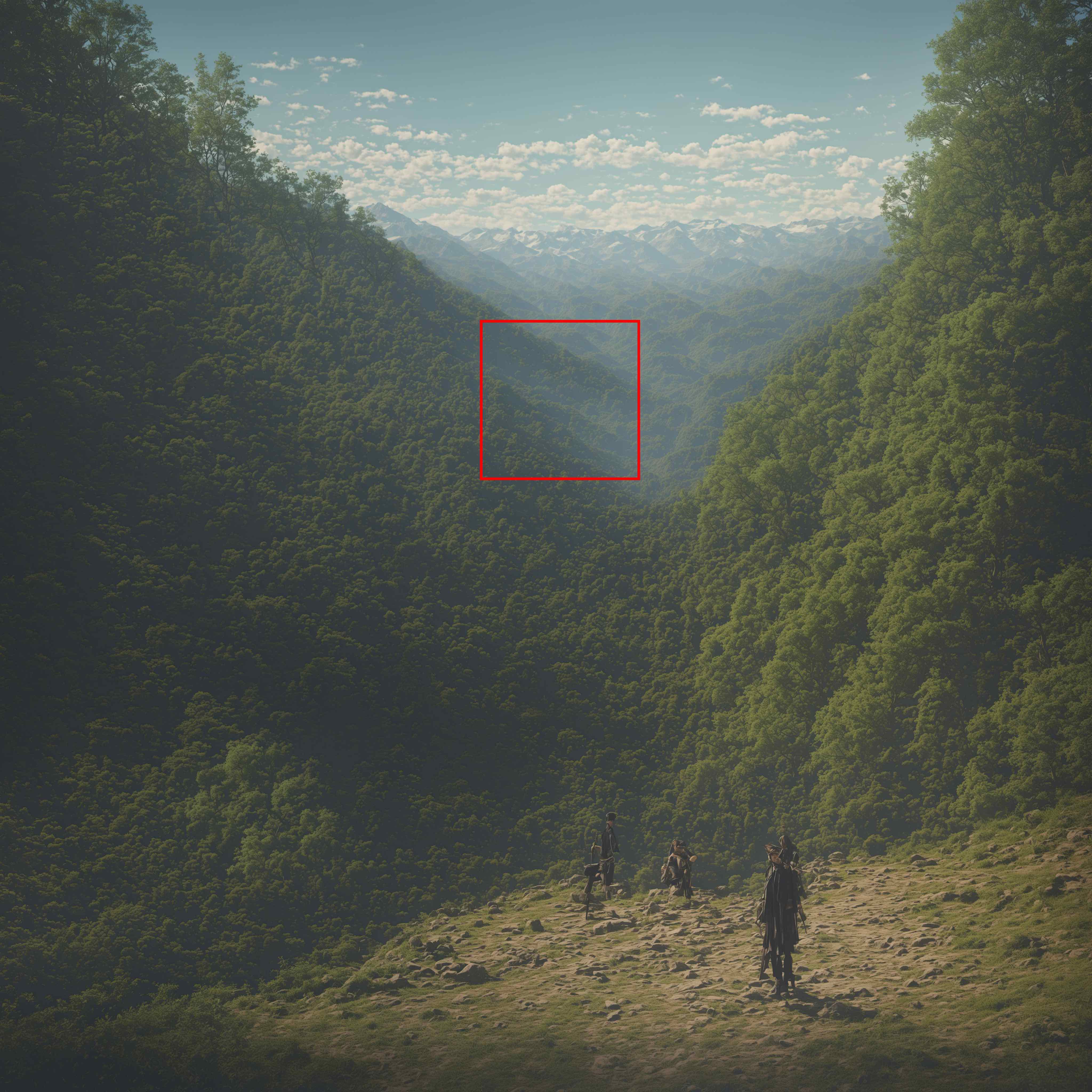}
\end{subfigure}%
\begin{subfigure}{.16\textwidth}
  \centering
  \includegraphics[width=\linewidth]{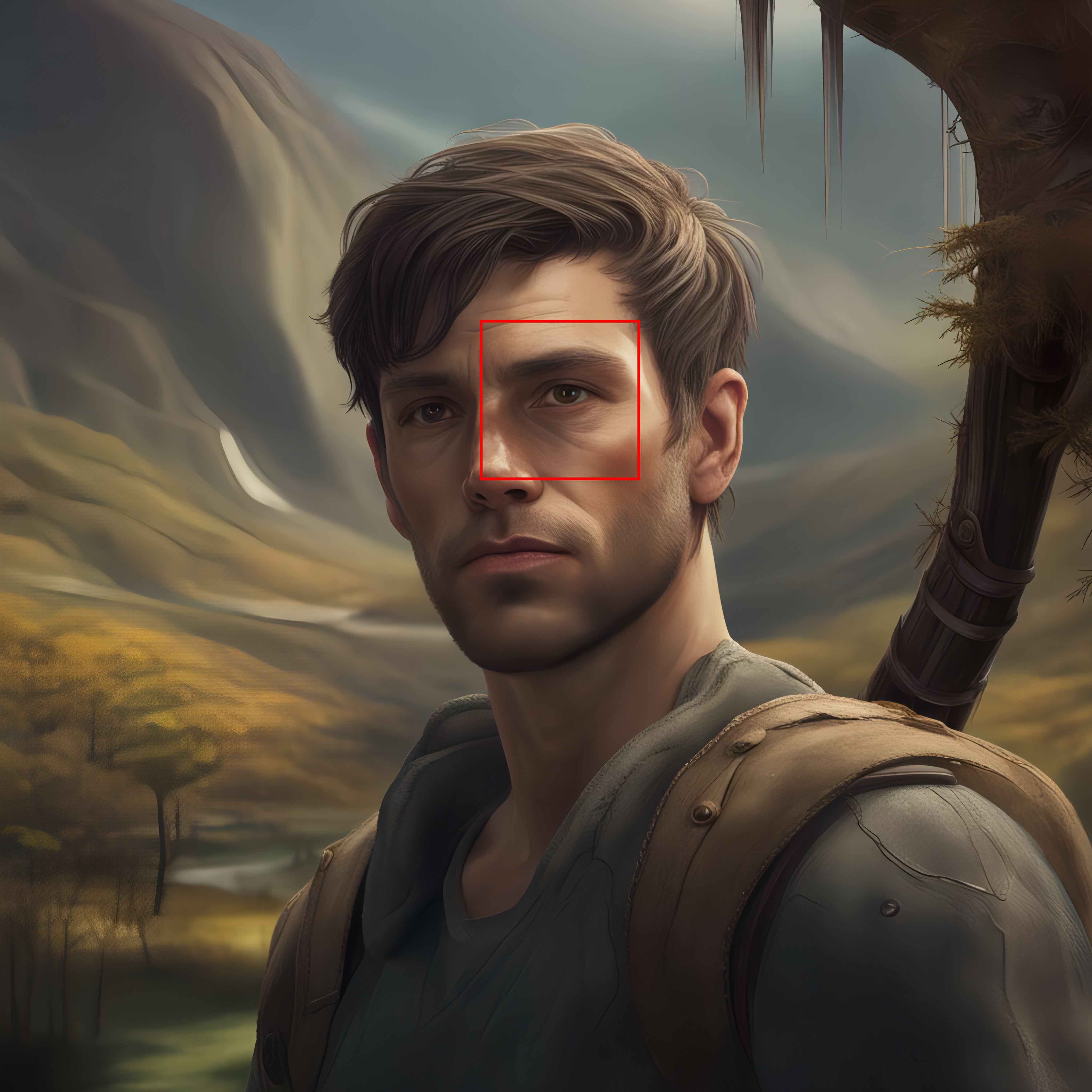}
\end{subfigure}%
\begin{subfigure}{.16\textwidth}
  \centering
  \includegraphics[width=\linewidth]{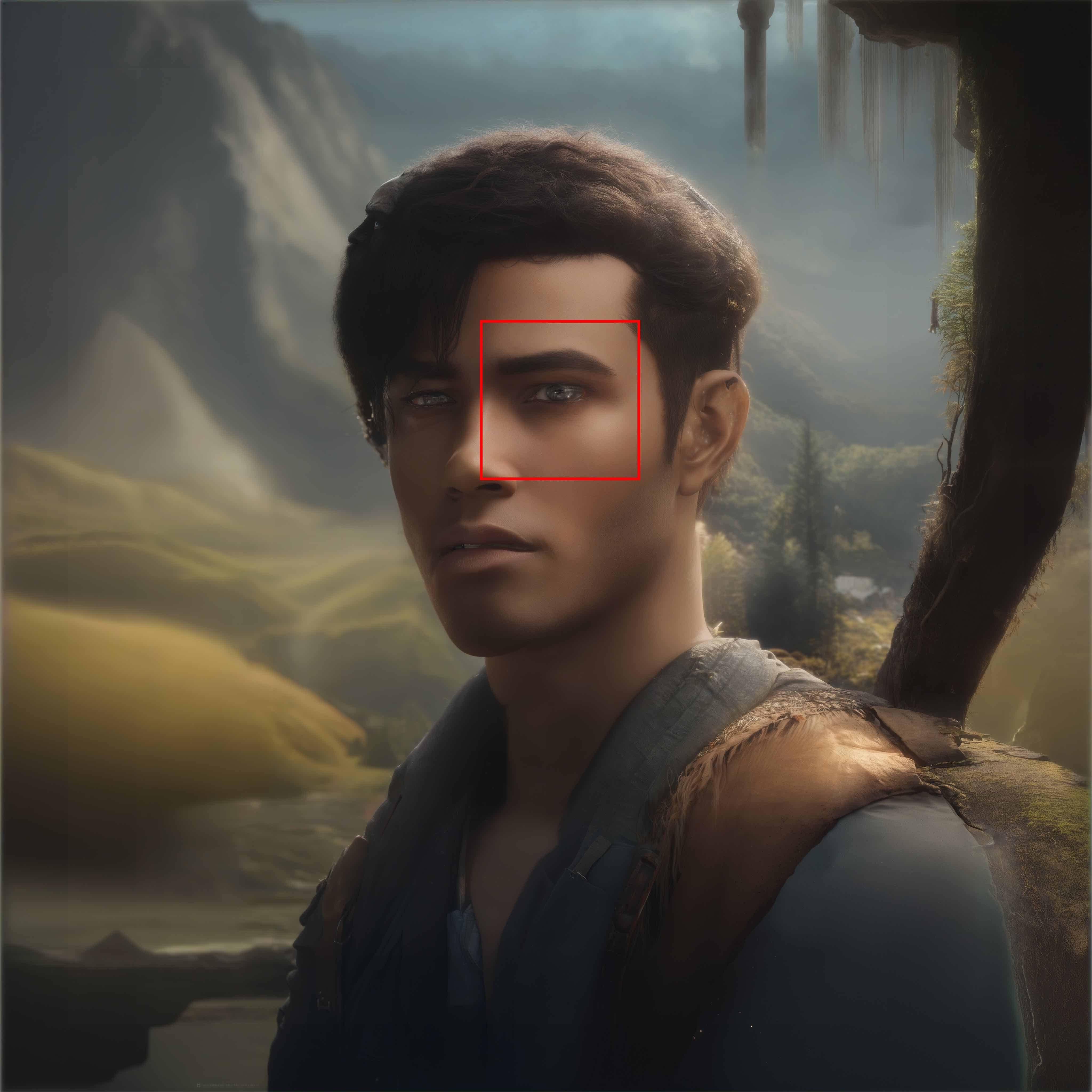}
\end{subfigure}%
\begin{subfigure}{.16\textwidth}
  \centering
  \includegraphics[width=\linewidth]{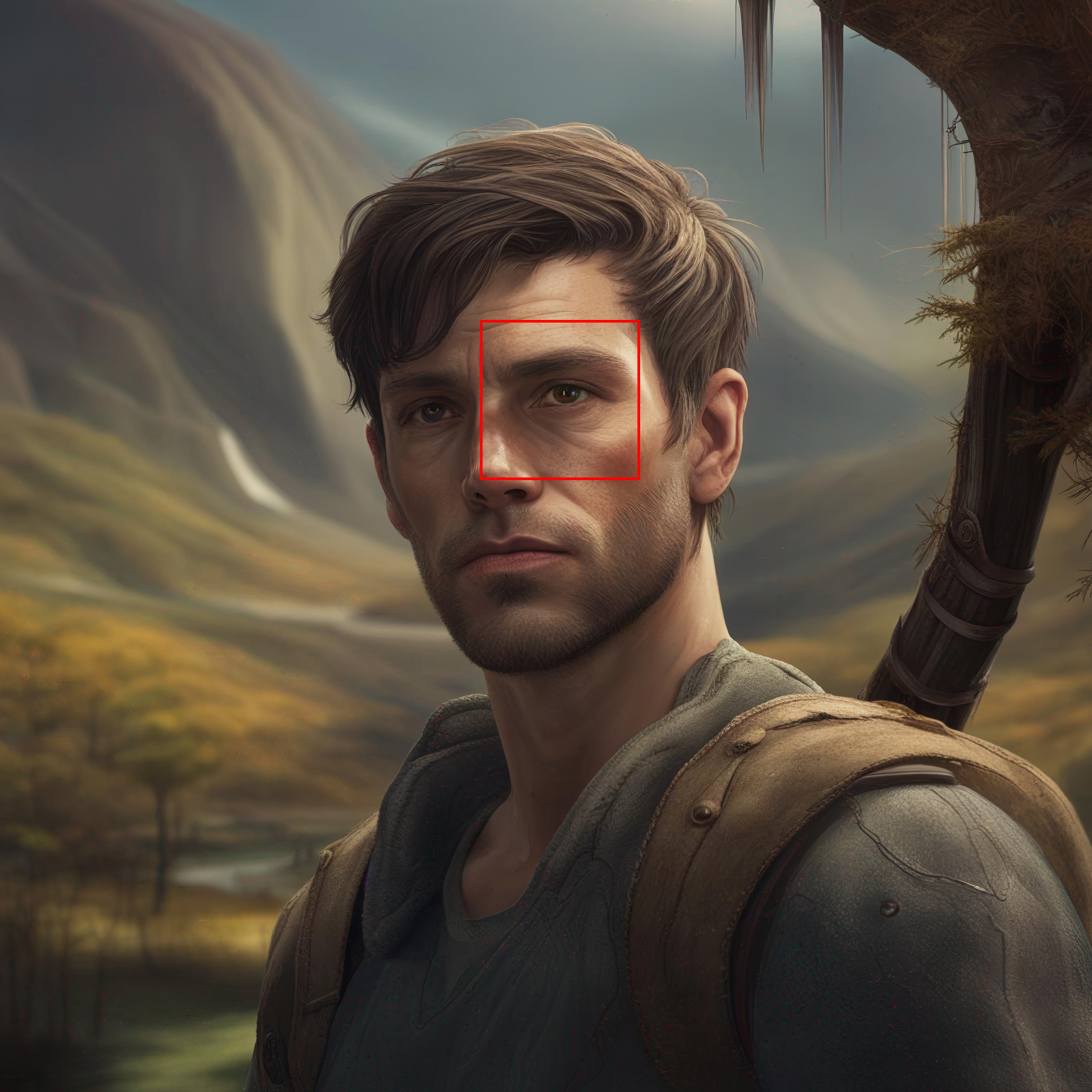}
\end{subfigure}
\begin{subfigure}{.16\textwidth}
  \centering
  \includegraphics[width=\linewidth]{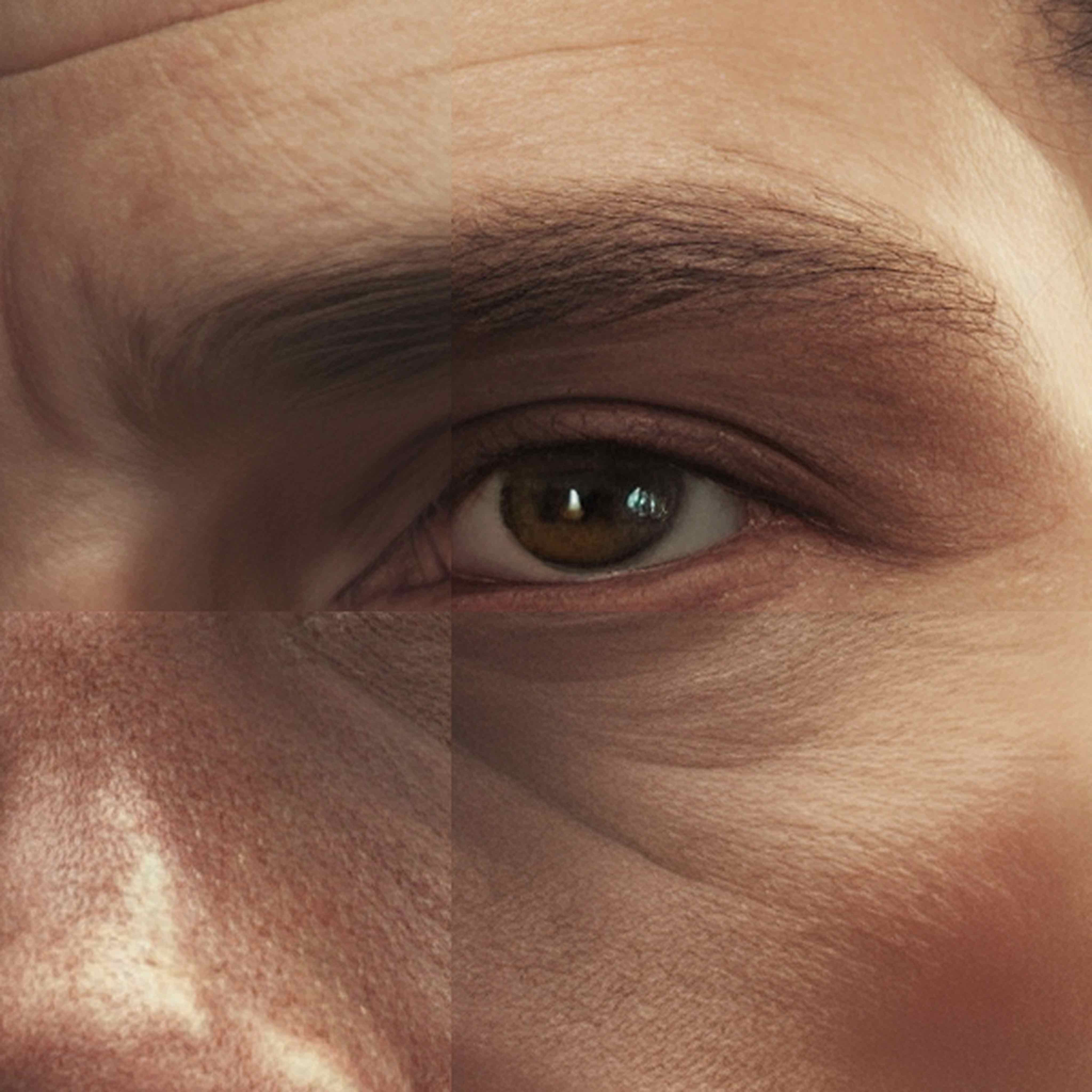}
  {\scriptsize\textbf{Patch}}
\end{subfigure}%
\begin{subfigure}{.16\textwidth}
  \centering
  \includegraphics[width=\linewidth]{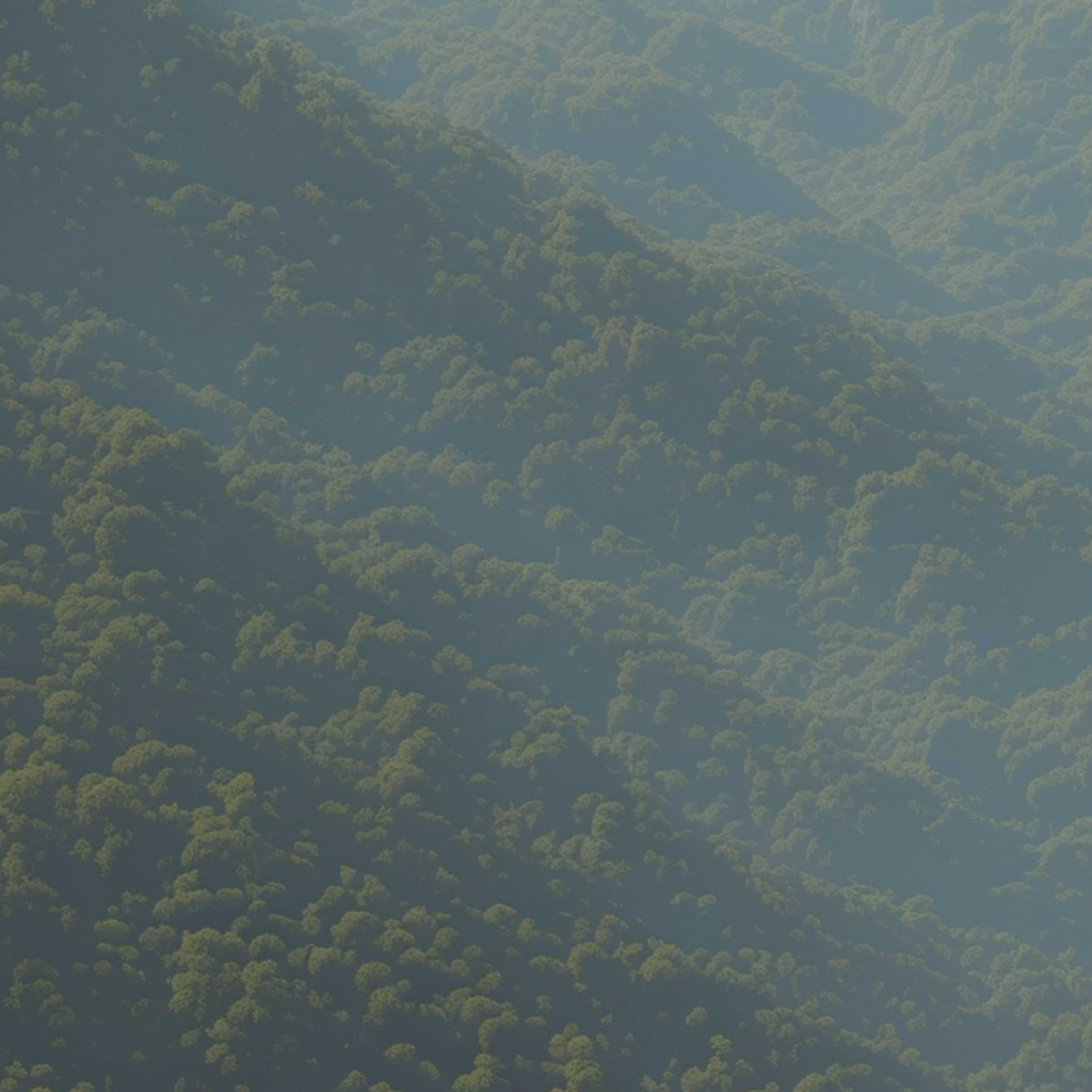}
  {\scriptsize\textbf{Scalecrafter}}
\end{subfigure}%
\begin{subfigure}{.16\textwidth}
  \centering
  \includegraphics[width=\linewidth]{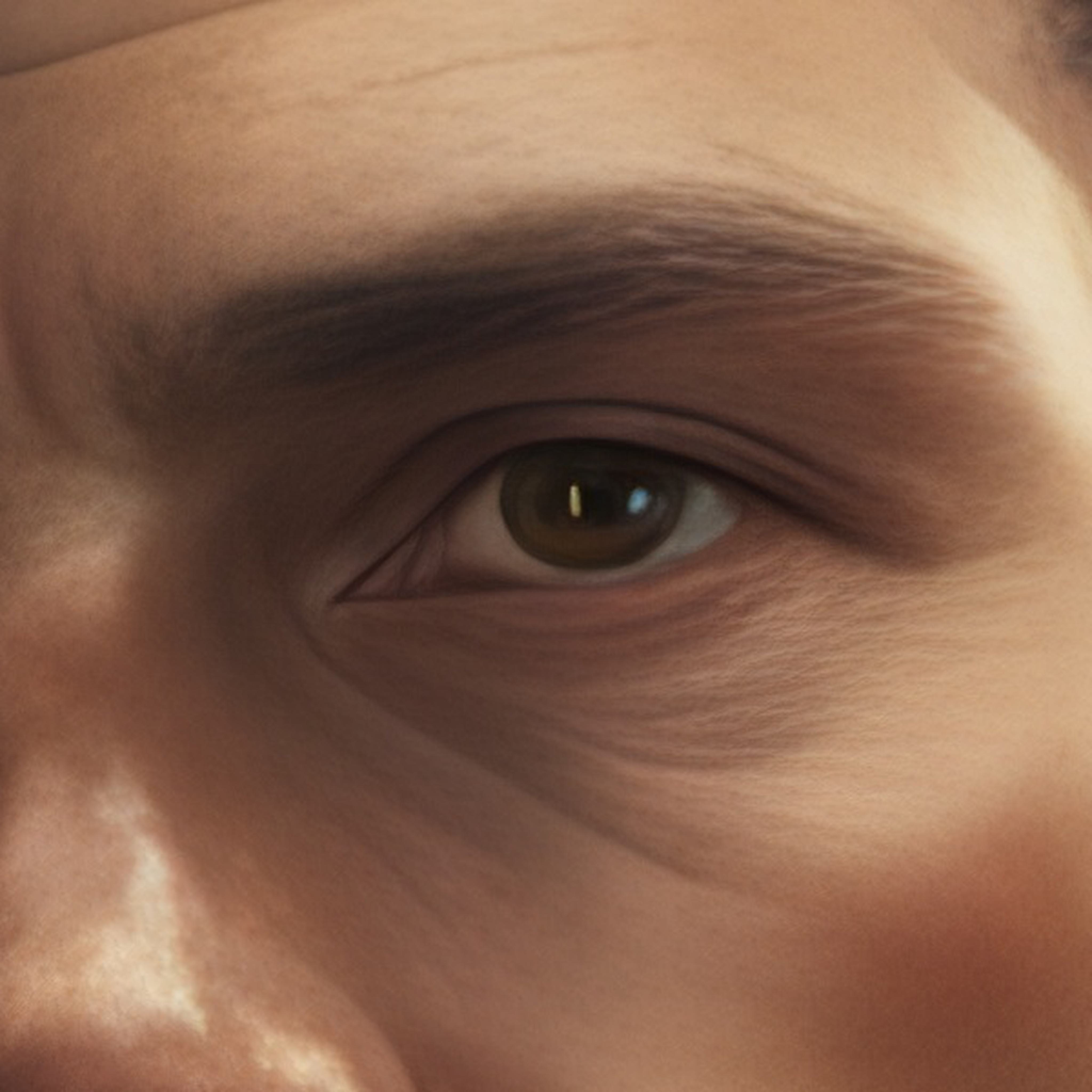}
  {\scriptsize\textbf{BSRGAN}}
\end{subfigure}%
\begin{subfigure}{.16\textwidth}
  \centering
  \includegraphics[width=\linewidth]{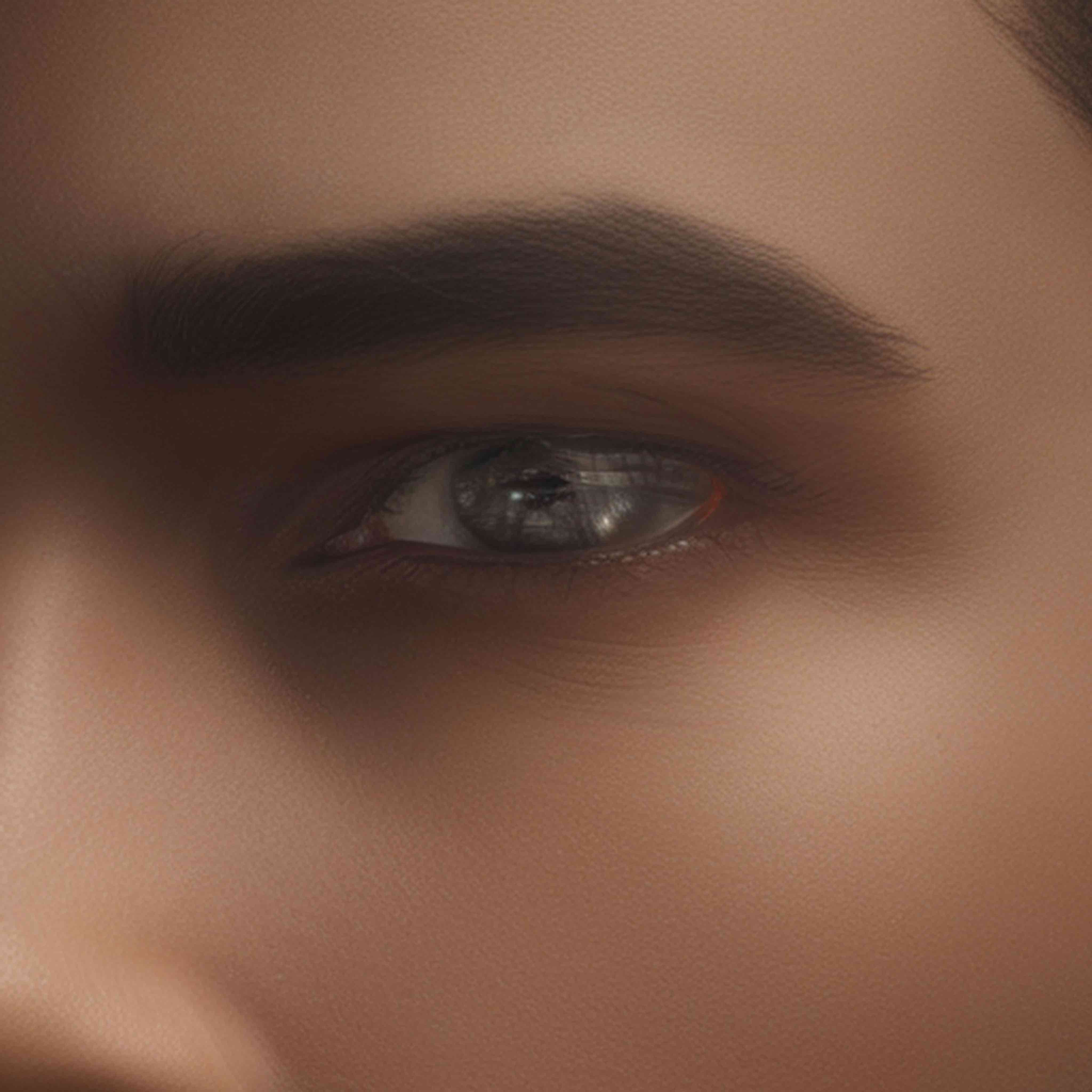}
  {\scriptsize\textbf{Demofusion}}
\end{subfigure}%
\begin{subfigure}{.16\textwidth}
  \centering
  \includegraphics[width=\linewidth]{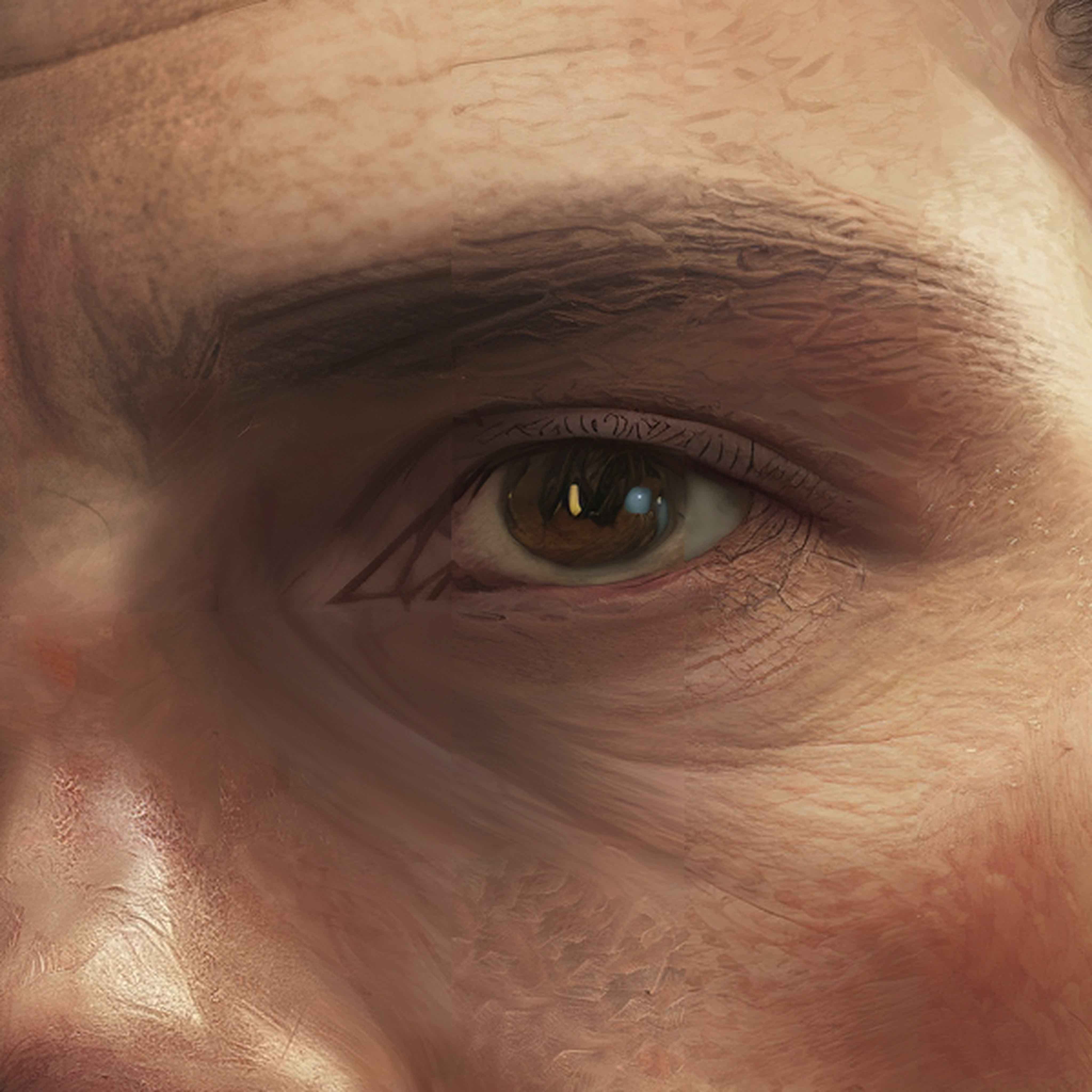}
  {\scriptsize\textbf{Inf-DiT}}
\end{subfigure}%
\captionsetup{skip=2pt}
\caption{{Qualitative comparison of different methods in detail at $4096\times4096$ resolution}}
\label{fig:images}
 \vspace{-7mm}
\end{figure}

\subsection{Machine Evaluation}
In this part, we conduct quantitative comparisons of Inf-DiT against state-of-the-art methodologies on ultra-high-resolution image generation tasks. The baselines encompass two main categories of high-resolution generation: 1. Direct high-resolution image generation, including Direct Inference of SDXL\cite{podell2023sdxl}, MultiDiffusion~\cite{bar2023multidiffusion}, ScaleCrafter~\cite{he2023scalecrafter}, and 2. High-resolution image generation based on super-resolution techniques, including BSRGAN~\cite{zhang2021designing}, 
DemoFusion~\cite{du2023demofusion}. We employ the FID (Fr\'echet Inception Distance)~\cite{heusel2017gans} to evaluate the quality of ultra-high-resolution generation, which is widely used to evaluate the perceptual quality of images in image generation tasks \cite{rombach2022high,peebles2023scalable,saharia2022photorealistic}. To further validate the super-resolution ability of our model, we additionally benchmark it against renowned super-resolution models over classic super-resolution tasks. 

\subsubsection{Ultra-High-Resolution Generation} \label{subsubsec:exp-high-generation}

We use the test set of HPDv2\cite{wu2023human} for evaluation. It contains 3200 prompts and is divided into four categories: ``Animation'', ``Concept-art'', ``Painting'', and ``Photo''. This allows for a comprehensive evaluation of the model's generative capabilities across various domains and styles. We test on two resolutions: $2048^2 $ and $4096^2$. For super-resolution-based models, we first use SDXL to generate a $1024^2$ resolution image and upsample it without text. We use $2 \times $ and $4 \times$ versions of BSRGAN for $2048^2$ and $4096^2$ generation, respectively. Although Inf-DiT was trained in a setting with $4 \times$ upsample, we find that it can generalize well at lower upsampling multiples. 
Therefore, for $2048^2$ generation, we directly resize the LR image from $1024^2$ to $2048^2$ and concatenate it with the noise input. We random samples 3200 images of $2048^2$ and $4096^2$ from LAION-5B as the distribution of real images.

As for the metrics, FID is designed to assess how closely generated images mimic real image distributions, and has become one of the default metrics in evaluating image generation. However, the original implementation of FID needs to downsample the input image to the resolution of $299\times299$ before the feature extraction, which ignores high-resolution details and can be inaccurate in assessing ultra-high-resolution image generation. Therefore, inspired by \cite{chai2022any,du2023demofusion}, we adopted the idea of $\text{FID}_{\text{crop}}$ and randomly crops $299\times299$ patches from high-resolution images for FID evaluation. Still, we retain the FID metric as it serves as an indicator of overall consistency, which the $\text{FID}_{\text{crop}}$ might overlook.

 The results in \cref{tbl:comparison} demonstrate that our model achieves state-of-the-art (SOTA) in three out of four metrics and surpasses all competitors in the mean score. This showcases our model's exceptional ability to generate high-resolution details and harmonious global information. The sole exception is the FID at $4096^2$, where it marginally trail behind BSRGAN by 1.0, but as discussed above, $\text{FID}_{\text{crop}}$ is a more representative metric for high-resolution features and Inf-DiT surpasses BSRGAN on $\text{FID}_{\text{crop}}$ on both $2048^2$ and $4096^2$ resolution. Our model is capable of being applied to all generative models, not just SDXL. Examples of upsampled images from other models are listed in the \cref{sec:appendix:morecase}. 
 
\subsubsection{Super-Resolution}
In addition to its ability to generate high-resolution images, Inf-DiT can also be used as a classic super-resolution model. We conduct the evaluation on DIV2k valid dataset\cite{agustsson2017ntire} which contains multiple real-world high-resolution images in different scenarios. Following \cite{rombach2022high, saharia2022image}, we fix image degradation to bicubic interpolation with $4 \times$ downsampling. Before comparing with fix-resolution models LDM\cite{rombach2022high} and StableSR\cite{wang2023exploiting}, we center-crop specific-sized patches from the high-resolution image as ground truth.
Throughout this process, we utilize both perceptual (FID, $\text{FID}_{\text{crop}}$) and fidelity (PSNR, SSIM) metrics to ensure a detailed and comprehensive assessment.


As demonstrated in \cref{tbl:resolution_comparison}, our model achieves state-of-the-art (SOTA) across all metrics. This signifies that, as a super-resolution model, ours not only excels in performing super-resolution at arbitrary scales but also in optimally preserving global and detailed information while restoring results that closely resemble the original images.
\subsection{Human Evaluation}
To further evaluate Inf-DiT and more accurately reflect its generative quality from a human perspective, we conduct a human evaluation. The comparison settings are the same as in \cref{subsubsec:exp-high-generation}, except that we excluded MultiDiffusion and Direct Inference to their non-competitive results. For each of the four categories, we randomly choose ten comparison sets, each encompassing outputs from four models, culminating in a total of 40 sets that form the human evaluation dataset. To ensure fairness, we randomize the sequence of model outputs in each comparison set. Human evaluators are asked to evaluate models based on three criteria: detail authenticity, global coherence, and consistency with the original low-resolution input. Each evaluator receives 20 sets of images on average. Within each set, evaluators are required to rank the images generated by four models from highest to lowest based on three criteria. 

\begin{figure}[h]
   \centering
   \includegraphics[width=\linewidth]{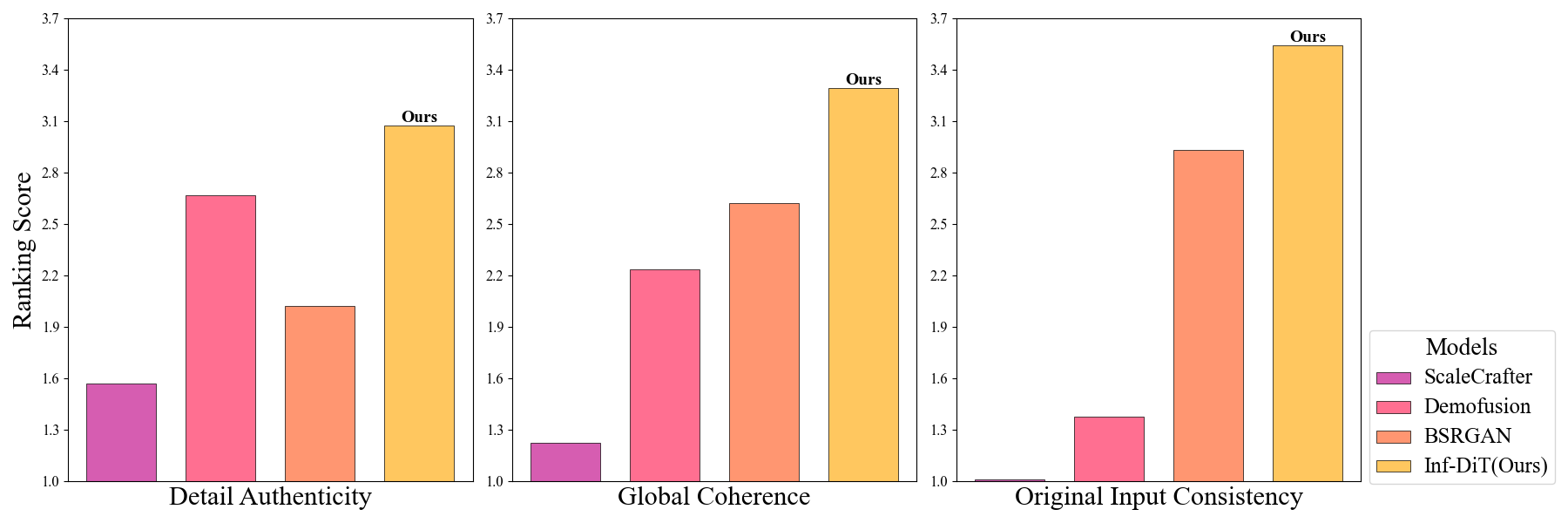}
   \captionsetup{skip=2pt}
   \caption{Human evaluation results. After participants rank the images from different models, We assign scores from 4 to 1 to models in order and finally calculate the average of all the results. Inf-DiT gets the highest scores in all three categories.}
   \label{fig:human_evalution_rank}
     \vspace{-9mm}
\end{figure}

We ultimately collected 3,600 comparisons. As depicted in \cref{fig:human_evalution_rank}, our model outperforms the other 3 methods in all the 3 criteria. It is particularly noteworthy that each of the other three models ranks relatively lower on at least one of the three evaluation criteria, while Inf-DiT achieves the highest score on all three criteria: detail authenticity, global coherence, and consistency with low-resolution input. This indicates that our model is the only one capable of excelling in both high-resolution generation and super-resolution tasks simultaneously.


 \vspace{-16pt} 

\subsection{Iterative Upsampling}
\vspace{-2mm}

\begin{figure}[t]
   \centering
   \includegraphics[width=\linewidth]{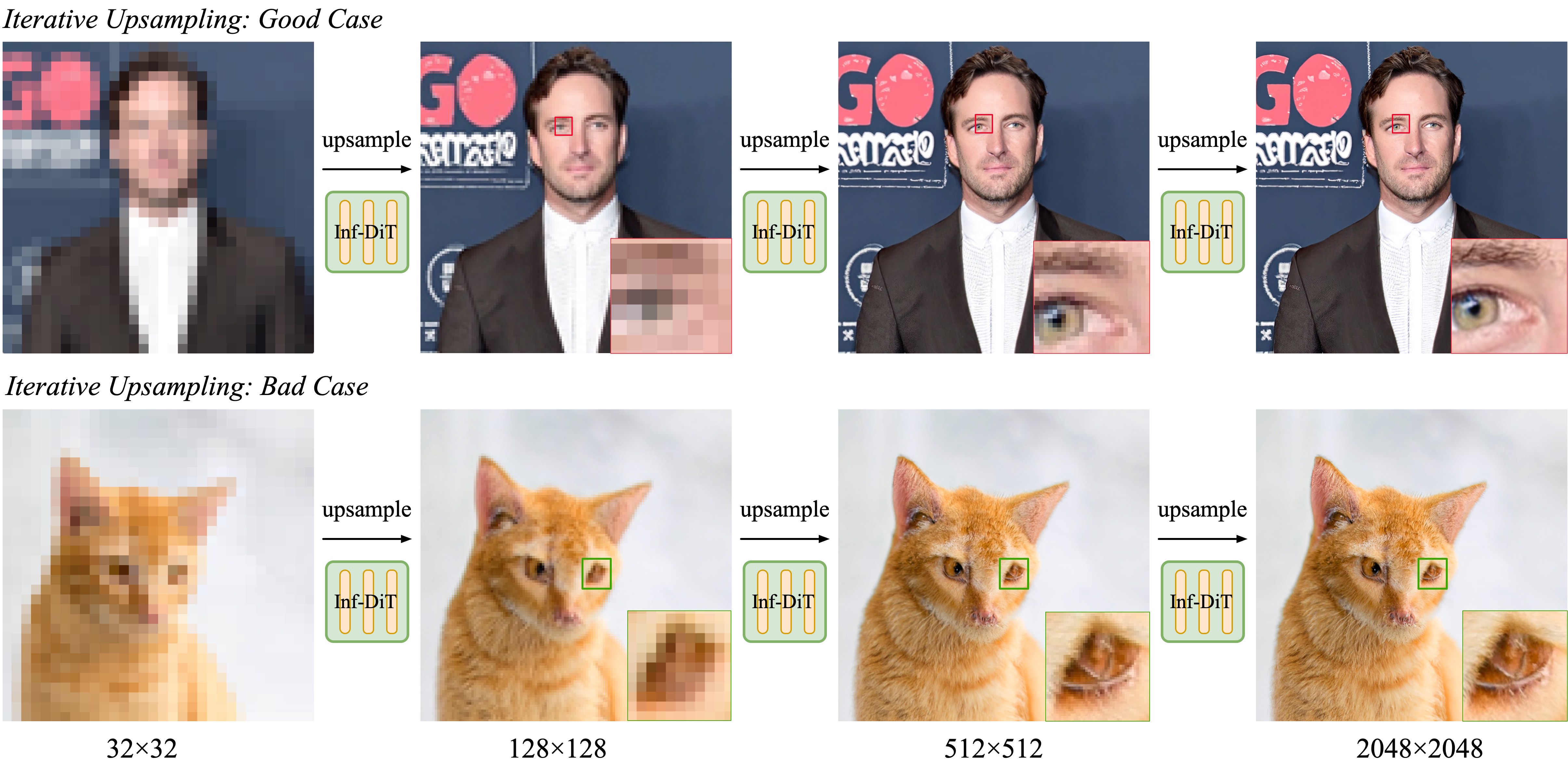}
   \caption{Generated samples of iterative upsampling. \textbf{Top}: Inf-DiT can upsample an image generated by itself several times, and generate details of different frequencies at corresponding resolution. \textbf{Bottom}: After failing to generate a pupul at the $128^2$ resolution, it is difficult for subsequent upsampling stages to correct this error. }
   \label{fig:interative_upsampling}
    \vspace{-5mm}
\end{figure}

Since our model can upsample images of arbitrary resolution, it's a natural idea to test if the model can iteratively upsample images generated by itself. In this study, we experiment on generating a $2048^2$ resolution image from a $32^2$ resolution image , $64 \times $ upscale, through three times of iterative upsampling. \Cref{fig:interative_upsampling} illustrates two cases of this process. In the first case, the model successfully produces a high-resolution image after three stages of upsampling. It generates details of different frequencies in different resolution upsampling: the contour of the face, the shape of the eyeball, and individual eyelashes. However, it is hard for the model to correct inaccuracies generated in earlier stages, leading to the accumulation of errors. An example of this issue is demonstrated in the second sample. We leave this problem for our future work.   

\begin{wraptable}[8]{r}{0.6\linewidth}
    \vspace{-14mm}
    \centering
    \caption{Ablation study on $2048 \times 2048$ resolution.}
    \begin{tabular}{ccc}
    \toprule
    Method & FID & $\text{FID}_\text{crop}$ \\
    \midrule
    Ours & \textbf{66.0} & \textbf{71.9} \\
    w/o Nearby LR Attention & 66.7 & 73.6 \\
    w/o Semantic Input & 66.2 & 72.8 \\
    Inner Block Attention & 66.3 & 83.7 \\
    \bottomrule
    \vspace{-15mm}
    \end{tabular}
    \label{tab:ablation_study}
\end{wraptable}

\subsection{Ablation Study}
We conduct ablation studies of our method on $2048^2$ resolution. We test the model performance on the HPDV2 dataset after separately removing the Global Input and Nearby LR attention. As a result, both the FID and $\text{FID}_{\text{crop}}$ of the model get worse to varying degrees. We also test Inner Block Attention setting, i.e. removing the attention between blocks,  and the $\text{FID}_{\text{crop}}$ experiences a significant decline. We list more case studies and analysis in the \cref{sec:appendix:abcases}. 


%% file: sections/2_related_works.tex
\section{Related Work}

\subsection{Diffusion Image Generation}
Diffusion models have emerged as a spotlight in the realm of image generation, boasting an array of groundbreaking advancements in recent years. Initially introduced in 2015\cite{sohl2015deep}, and further developed through works such as DDPM\cite{ho2020denoising} and DDIM\cite{song2020denoising}, these models leverage a stochastic diffusion process, conceptualized as a Markov chain, to convert a simple prior distribution, like Gaussian noise, into a complex data distribution. This methodology yield impressive outcomes in terms of the quality and diversity of generated images.

Recent enhancements have markedly elevated the generation capabilities of diffusion models. CDM\cite{ho2022cascaded} creates a cascade generation pipeline with multistage super-resolution models, which can be applied to large pretrained models\cite{saharia2022photorealistic, ramesh2022hierarchical}.

The introduction of Latent Diffusion Models (LDMs)\cite{rombach2022high, podell2023sdxl} represents a pivotal extension, incorporating a latent space to boost both efficiency and scalability. 
Alongside these advancements, there has been significant progress in the optimization of network architectures. The advent of Diffusion Transformers (DiT)\cite{peebles2023scalable} Replaces U-Net with ViT\cite{dosovitskiy2020image} for noise prediction.


\subsection{Image Super-Resolution}
Given a low-resolution (LR) image $I_{LR}$ degraded from high-resolution (HR) observation $I_{HR}$, image super-resolution (SR) aims to reconstruct a HR one $\hat{I}_{HR}$:
$ I_{LR} = D(I_{HR}; \delta), \quad \hat{I}_{HR} = F(I_{LR}; \theta) $. 
Here \(D\) and \(F\) refer to degradation process and the super-resolution model.  \(\delta\) and  \(\theta\) represent the parameters.  

In recent years, Blind SR has been a major focus: in which the degradation process is unknown and learnable. This perspective has led to the development of effective modeling techniques, such as BSRGAN\cite{zhang2021designing} and Real-ESRGAN\cite{wang2021real}.

Recently, diffusion-based SR methods have yielded exciting results. These works focus on finetuning pre-trained text-to-image diffusion models to take advantage of their excellent generative ability. Specifically, DiffBir\cite{lin2023diffbir} employs ControlNet\cite{zhang2023adding} on pre-trained stable-diffusion models, whereas PASD\cite{yang2023pixel} enhances it by executing pixel-aware cross-attention. Both approaches have garnered considerable success in fixed resolution super-resolution but cannot be directly used for higher resolution.

\subsection{Ultra-High-Resolution Image Upsampler}
Currently, image generation methods fall short in generating ultra-high-resolution images owing to the constraints of memory and issues with training efficiency. Under these circumstances, MultiDiffusion\cite{bar2023multidiffusion} and Mixture of Diffusers\cite{jimenez2023mixture} bind together multiple diffusion generation processes by dividing images into overlapping blocks, processing each separately, and then stitching them together, aiming to maintain continuity between blocks. However, because they only used the local weighted averaging for aggregation, it leads to a low interaction efficiency and makes it difficult to ensure the global consistency of the images.

Given this concern, DemoFusion\cite{du2023demofusion} and ScaleCrafter\cite{he2023scalecrafter} adapt dilated policies including dilated sampling and dilated convolution kernels, aiming at acquiring more global information. These methods indeed achieve improvement on a global semantic level without the need for additional training. However, the huge difference between training and generation leads these methods to easily produce illogical images.


%% file: sections/5_conclusion.tex
\section{Conclusion}
In this work, we observe that the major obstable to generating ultra-high-resolution images is the substantial memory occupied by model hidden states. Based on this, We propose Unidirectional Block Attention mechanism (UniBA) which can lower the space complexity by performing batch generation among blocks. With UniBA, we train Inf-DiT, a $4 \times $ memory-efficient image upsampler which achieves state-of-the-art performance in both generation and super-resolution tasks.

%% file: sections/6_appendix.tex
\appendix
\section{Implementation Detail} \label{sec:appendix:imp}
\subsection{Full List of Training Hyperparameters}
\begin{table}[h]
  \centering
  \begin{tabular}{lcclcc}
    \toprule
    \multicolumn{4}{c}{Training Hyperparameters of Inf-DiT} \\
    \midrule
    Number of Layers & 28 & & Attention heads & 16 \\
    Hidden size & 1280 & & Attention head size & 80 \\
    FFN inner hidden size & 5120 & & Dropout & 0 \\
    Warmup Steps & 10k& & Peak Learning Rate & 1e-4 \\
    Batch Size & 320 & & Weight Decay & 1e-4 \\
    Max Steps & 1M & & Learning Rate Decay & Linear \\
    Gradient Clipping  & 0.1 & & Adam $\epsilon$ & 1e-8 \\
    Adam $\beta_1$ & 0.9 & & Adam $\beta_2$ & 0.099 \\
    \bottomrule
  \end{tabular}
  \caption{Hyperparameters for Inf-DiT Model}
  \label{Hyperparam}
\end{table}

\subsection{Efficient Implementation of UniBA}
\begin{figure}[htp]
    \centering
    \includegraphics[width=0.9\linewidth]{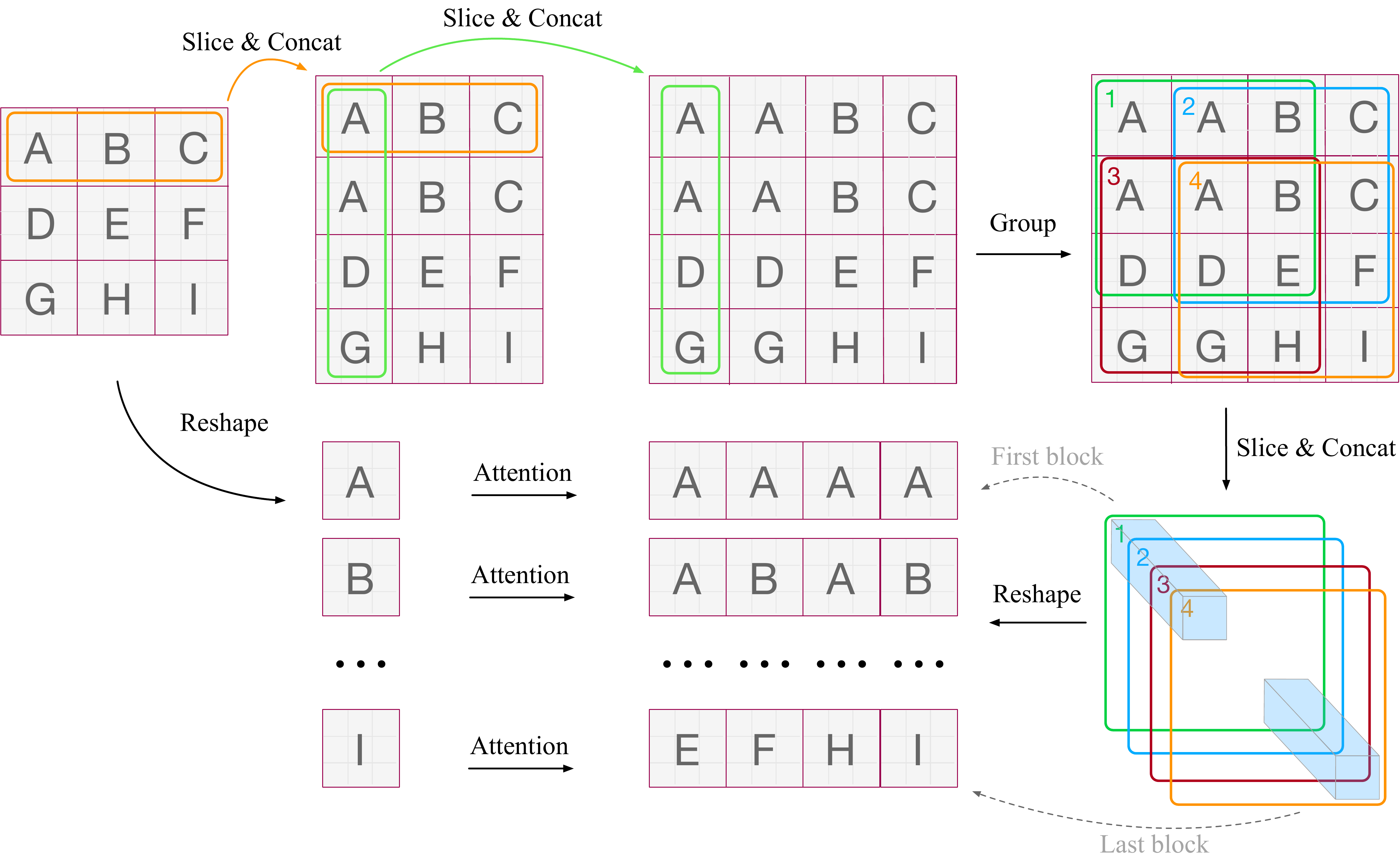}
    \caption{Unidirectional Block Attention during whole image input}
    \label{fig:uniba}
\end{figure}

\begin{figure}
\centering
\begin{Verbatim}[fontsize=\scriptsize] 

def get_cat_layers(x):
    # Duplicate and prepend first row
    cat_x = concat([x[:, 0:1], x], dim=1)

    # Duplicate and prepend first column
    cat_x = concat([cat_x[:, :, 0:1], cat_x], dim=2)
    
    # Concat four group: self, left-top, left, top
    x = concat([x, cat_x[:, :x.shape[1], :x.shape[2]],
                   cat_x[:, 1:, :x.shape[2]], 
                   cat_x[:, :x.shape[1], 1:]], dim=3)
    return x

# The shape of query, key, value : 
# [batch_size, h/block_size, w/block_size, block_size*block_size, hidden_size]
key = get_cat_layers(key)
value = get_cat_layers(value)
output = attention(query, key, value)
\end{Verbatim}
\hfill%
\caption{Pytorch-style code implementation of UniBA.}
\label{fig:pytorch_code}
\end{figure}

In \cref{fig:uniba} and \cref{fig:pytorch_code}, we demonstrate the efficient implementation of UniBA, where input contains $n\times n$ blocks. It is particularly noteworthy that for the blocks located in the first row and the first column, the number of blocks they depend on is fewer than three. Due to the property 
$\text{Attention}(\bm{x}\bm{W}^Q, \bm{y}\bm{W}^K, \bm{y}\bm{W}^V) = \text{Attention}(\bm{x}\bm{W}^Q, [\bm{y}\bm{W}^K, \bm{y}\bm{W}^K,...], [\bm{y}\bm{W}^V, \bm{y}\bm{W}^V,...])$, our approach can correctly handle these situations.  

\subsection{Stabilization of Training}

Currently, many large diffusion models are trained under FP16 (16-bit precision) to reduce GPU memory and computation time. In practice, however, a sudden increase in loss often occurs after training for several thousand steps. We observe that when the number of patches in a single attention window is too many (e.g. there are 4096 patches with $block\ size=128$ and $patch\ size=4$), the attention score $Q^TK/\sqrt{d}$ can become very large, leading to unstable gradients or even overflow. Drawing on the experience of training large-scale ViT\cite{dehghani2023scaling}, we apply QK Normalization, which add an extra LayerNorm to Query and Key vector before multiplication: 
$$\text{Attention}(Q,K,V) = \text{Softmax}(\frac{LN(Q)LN(K)^T}{\sqrt{d}})V $$

Also, we employed BF16 format during training to further address the issue of overflow due to its broader numerical range. 

\subsection{Initial Noise during Inference}

\begin{figure}[htp]
    \centering
    \includegraphics[width=0.6\linewidth]{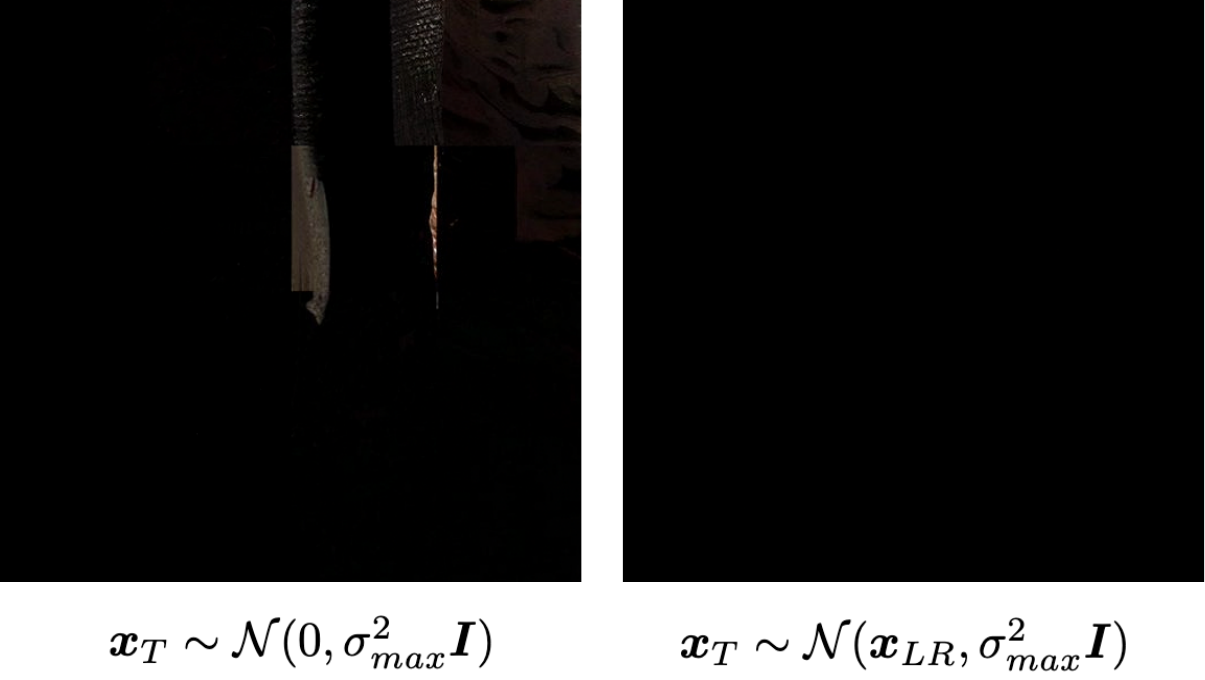}
    \caption{When upsampling an image of pure black, it is observed that the default initial noise may generate blocks of other colors. This issue can be resolved by incorporating LR information into the initial noise. }
    \label{fig:initnoise}
\end{figure}

Previous experience\cite{crosslabs2023diffusion} shows that the diffusion modal tends to generate images that are closely aligned with the initial noise, which may lead to color mismatching with the original image during upsampling, e.g. \cref{fig:initnoise}. Fortunately, for the setting of Inf-DiT, there is a natural prior that can control the mean of the initial noise, the LR image. The idea is, to replace the initial noise from $\bm{x}_T \sim \mathcal{N}({0, \sigma_{max}^2\bm{I})}$ to $\bm{x}_T \sim \mathcal{N}({\bm{x}_{LR}, \sigma_{max}^2\bm{I})}$during sampling, where $\bm{x}_{LR}$ is resized low-resolution input. However, this method can sometimes lead to the generated images appearing somewhat blurred just like LR images.

\section{Ablation Study Analysis}  \label{sec:appendix:abcases}
\begin{figure}[H]
    \centering
    \includegraphics[width=1\linewidth]{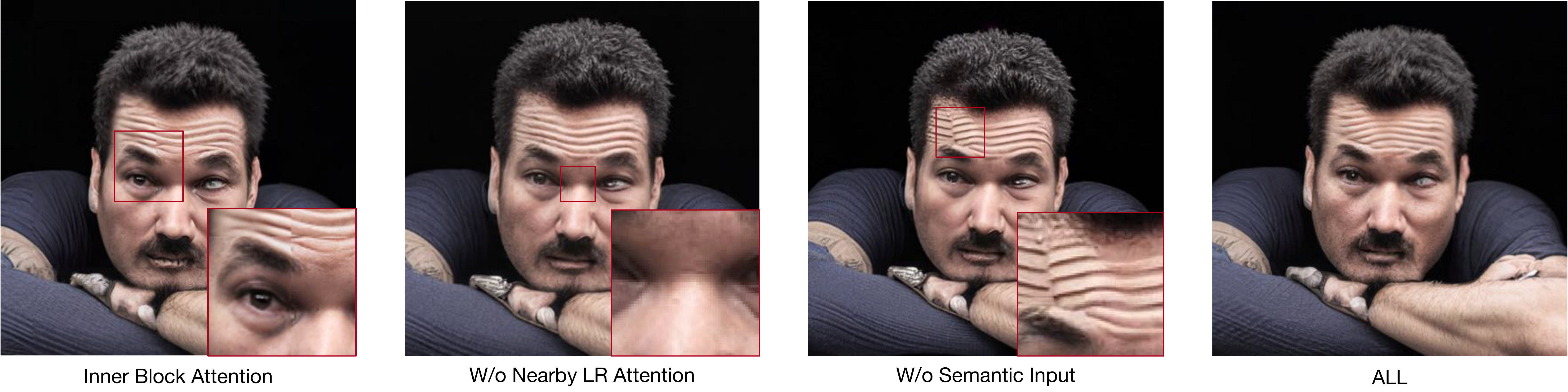}
    \caption{Ablation study cases. The first three images separately remove cross-block attention, Semantic Input (CLIP image embedding), and Nearby LR attention.}
    \label{fig:ab}
\end{figure}

In \cref{fig:ab}, we demonstrate the results of image generation after the removal of various components. When attention is restricted to individual blocks, there is a noticeable discontinuity between different blocks. The UniBA allows for generating continuous content across blocks, though occasional bad cases may arise. Incorporating cross-attention to LR images can further reduce the probability of discontinuous blocks. Additionally, the removal of semantic input, which corresponds to CLIP image embedding, results in the generation of textures that are mismatched with the global semantics. 

\section{More show cases}  \label{sec:appendix:morecase}
\subsection{Results based on different Generation Model}
Inf-DiT is capable of performing upsampling on images generated by any generation model, we show more cases here.
\vspace{4mm}

\noindent \textbf{Stable Diffusion V2}
 
\FloatBarrier
\begin{figure}[H]
\vspace{-4mm}
\centering
\begin{subfigure}{.18\textwidth}
  \centering
  \includegraphics[width=\linewidth]{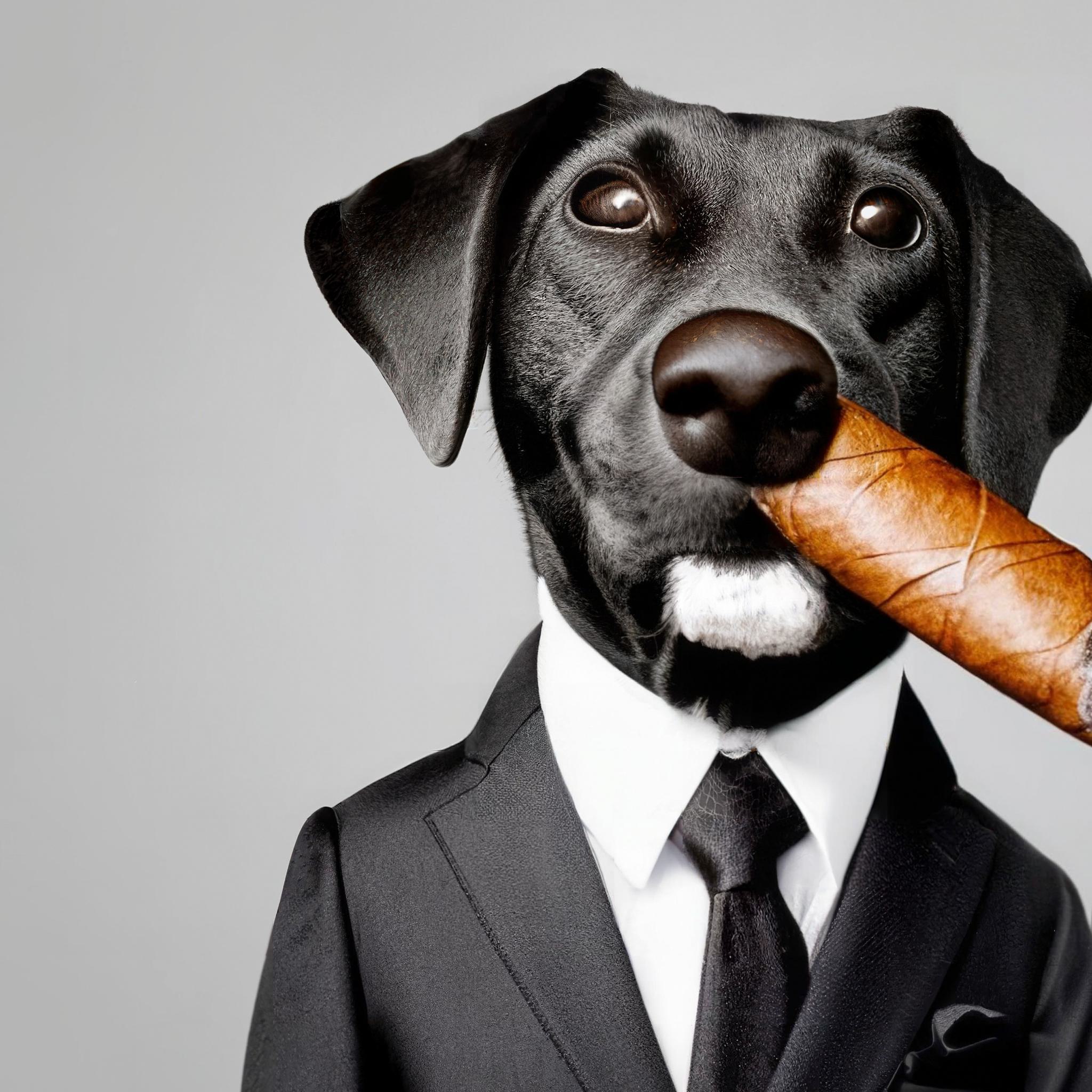}
\end{subfigure}%
\begin{subfigure}{.18\textwidth}
  \centering
  \includegraphics[width=\linewidth]{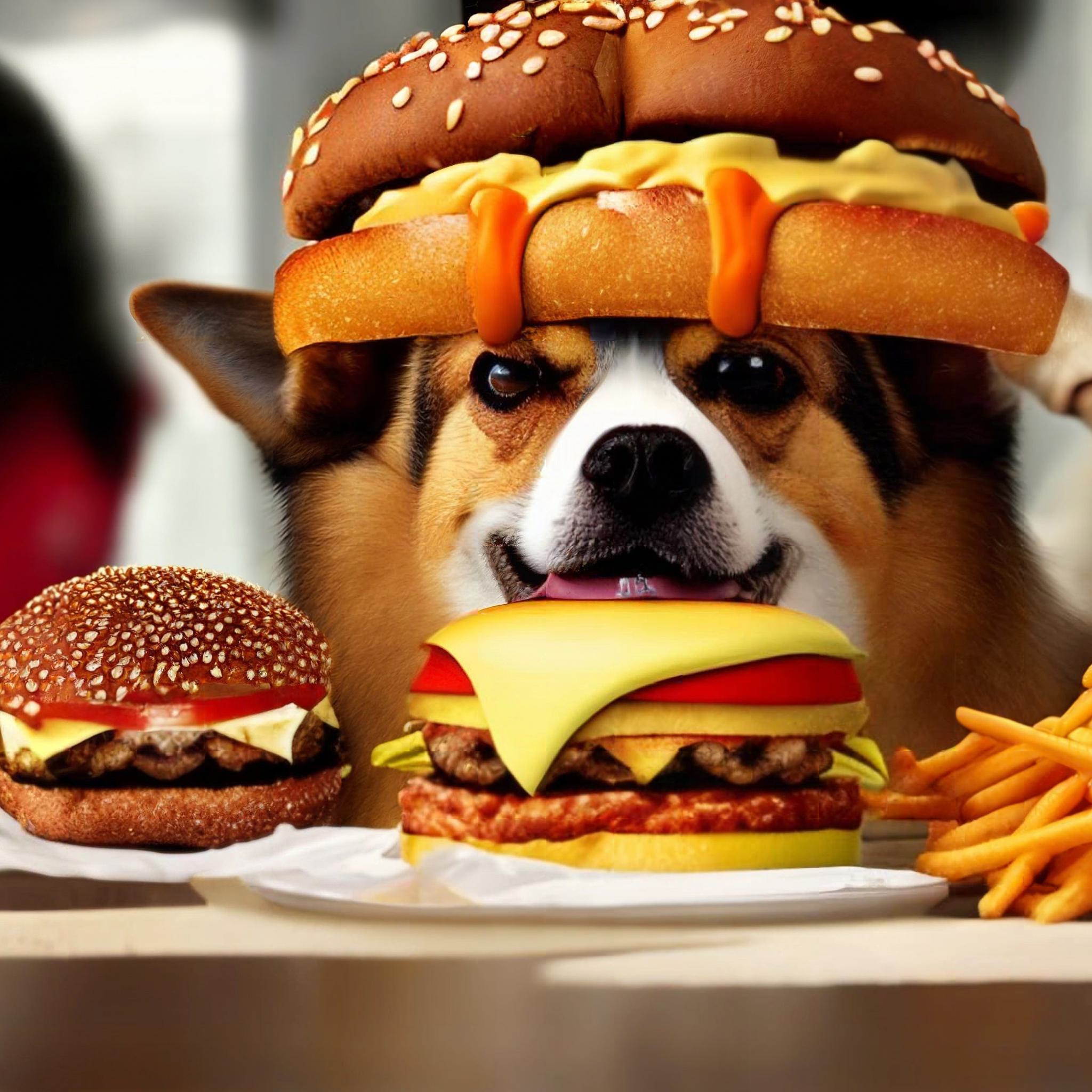}
\end{subfigure}%
\begin{subfigure}{.18\textwidth}
  \centering
  \includegraphics[width=\linewidth]{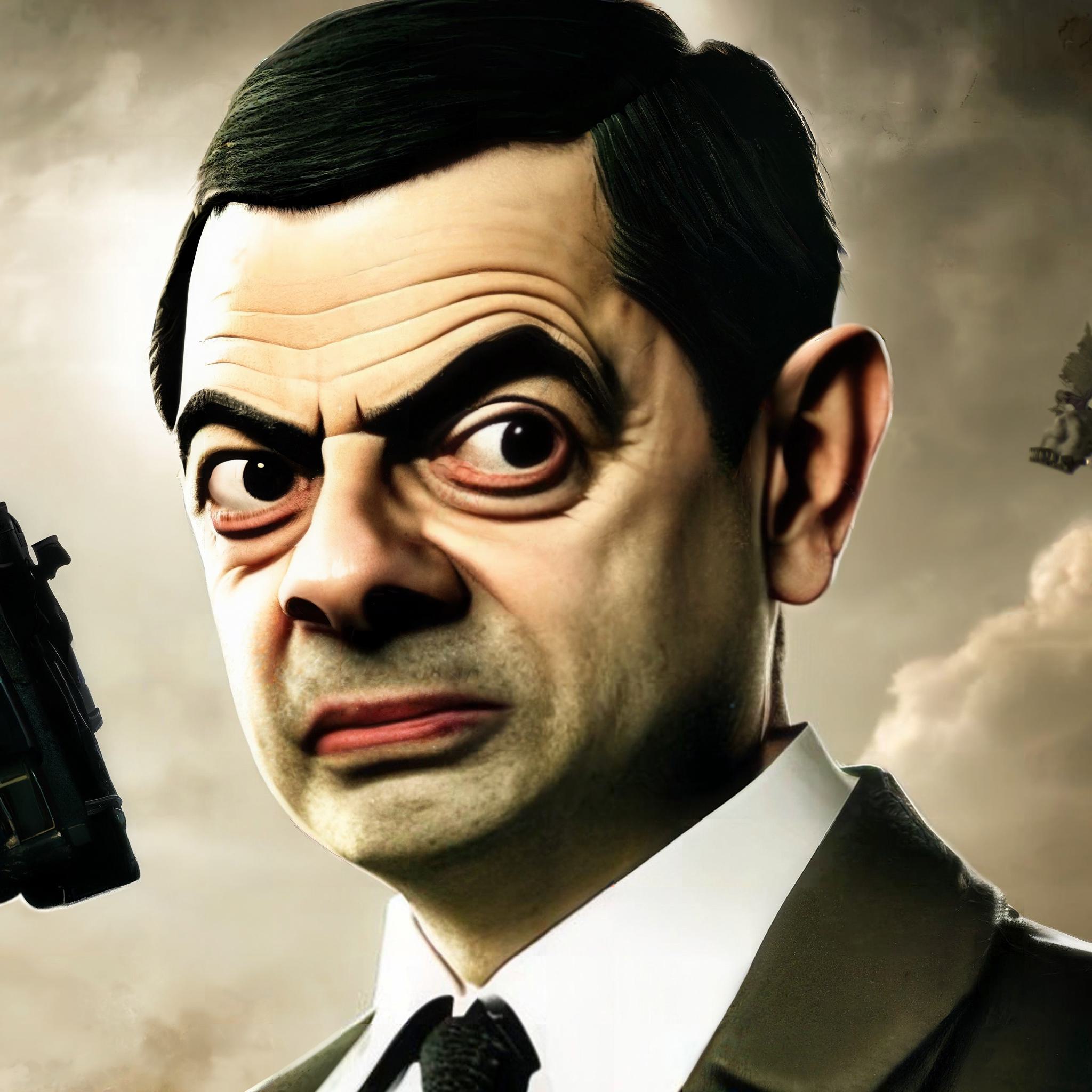}
\end{subfigure}%
\begin{subfigure}{.18\textwidth}
  \centering
  \includegraphics[width=\linewidth]{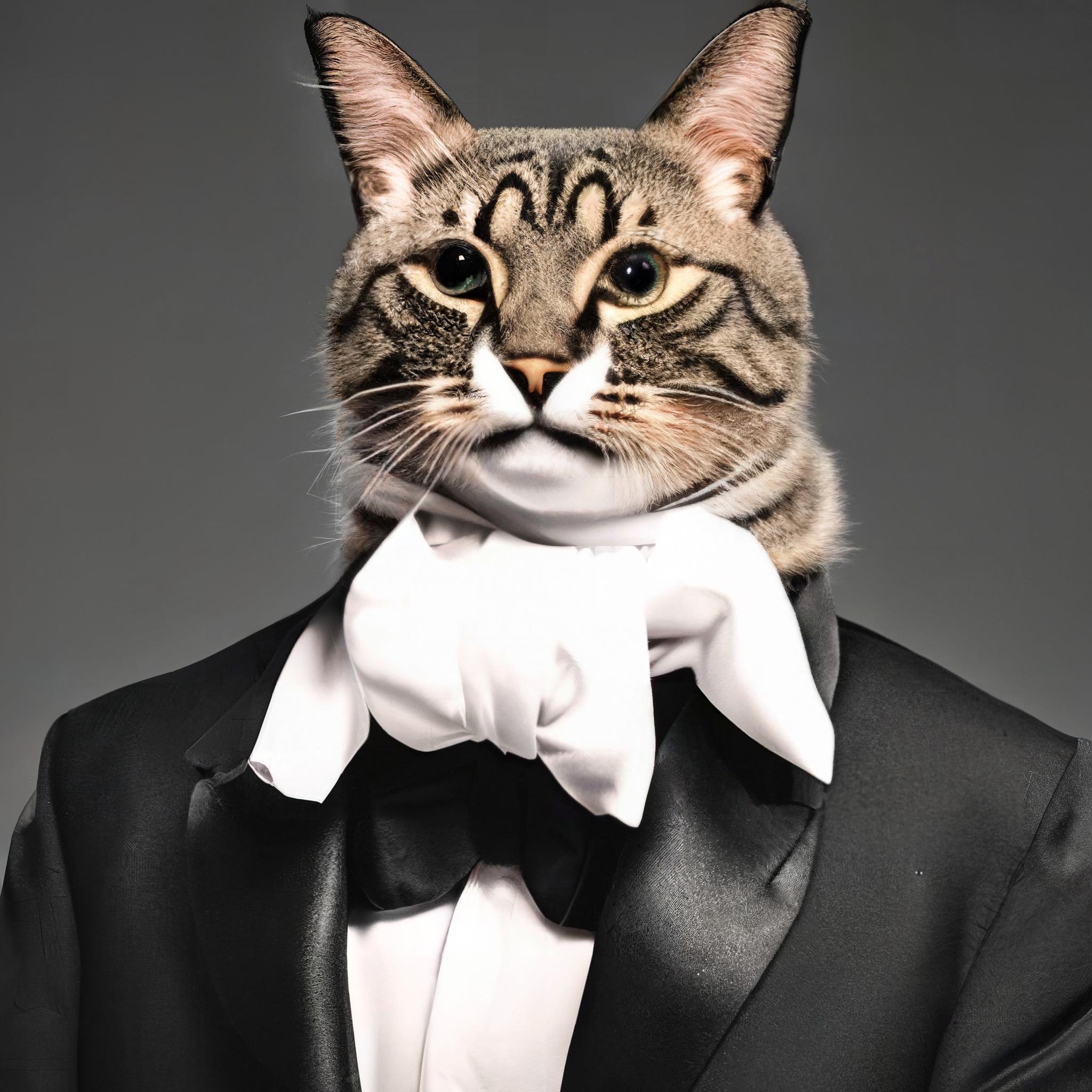}
\end{subfigure}%
\begin{subfigure}{.18\textwidth}
  \centering
  \includegraphics[width=\linewidth]{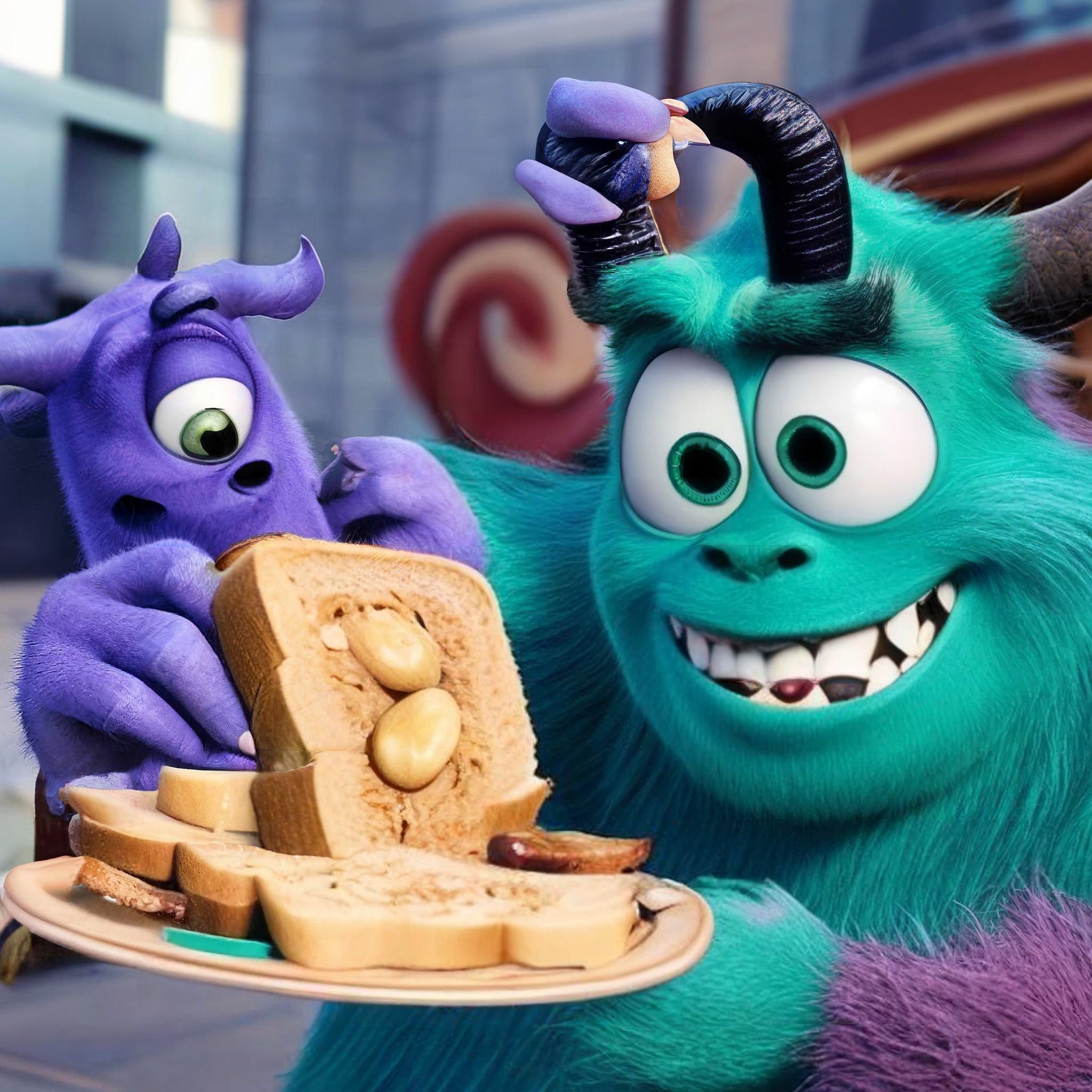}
\end{subfigure}%
\caption{Super-resolution applied with Stable Diffusion V2 model: upsampling image to $2048\times 2048$.}
\label{fig:images}
\vspace{-4mm}
\end{figure}
\noindent \textbf{DALL-E2}
\begin{figure}[H]
\vspace{-4mm}
\centering
\begin{subfigure}{.18\textwidth}
  \centering
  \includegraphics[width=\linewidth]{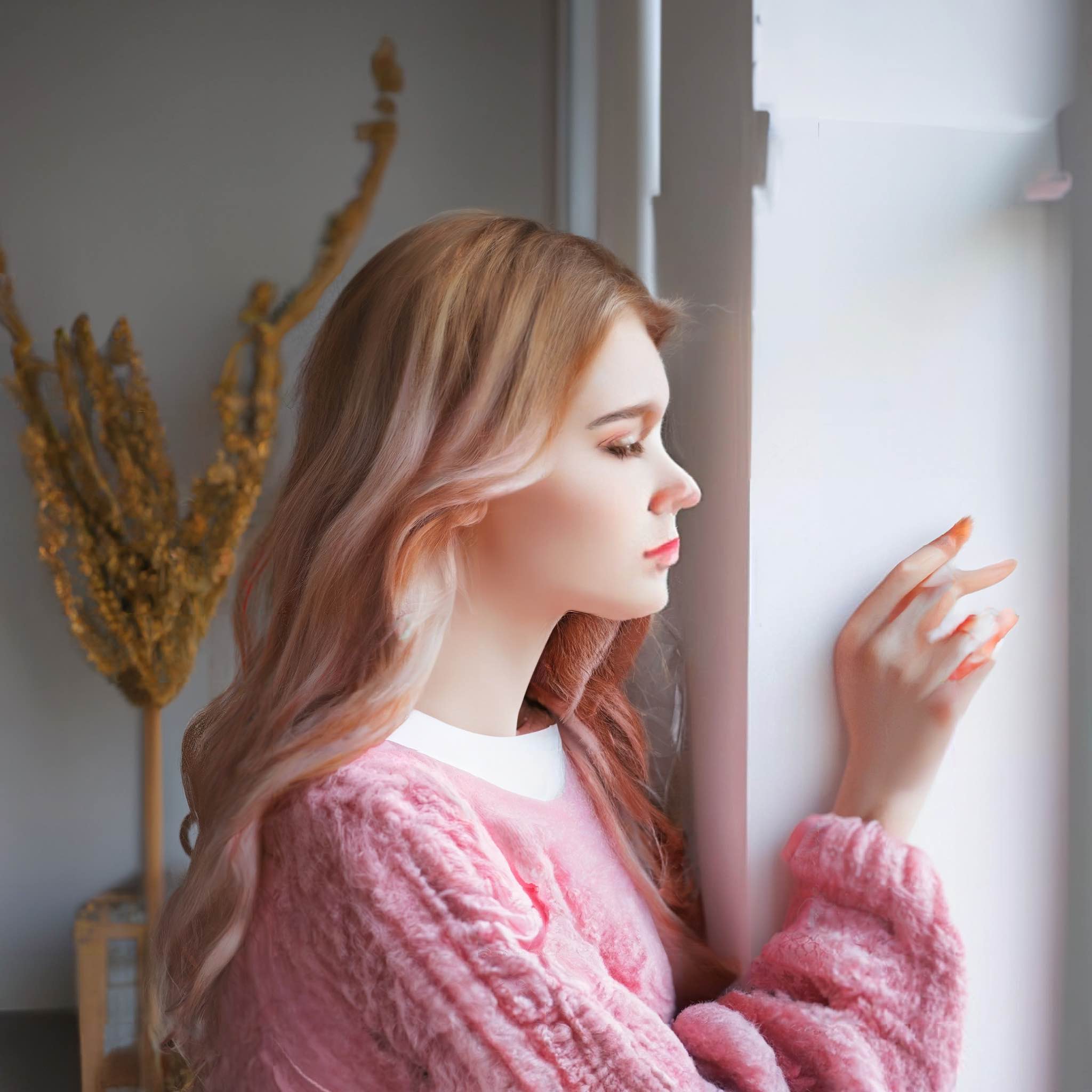}
\end{subfigure}%
\begin{subfigure}{.18\textwidth}
  \centering
  \includegraphics[width=\linewidth]{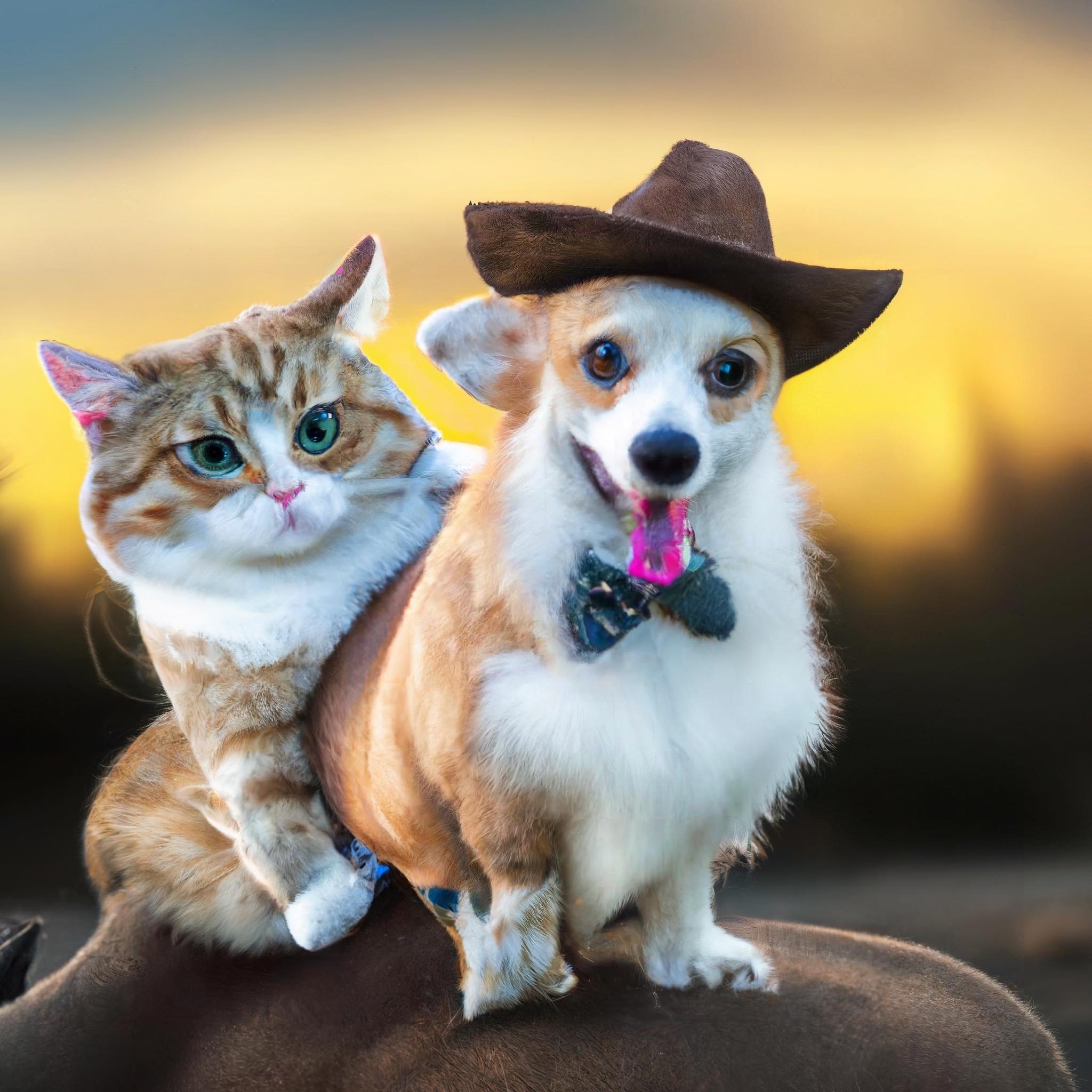}
\end{subfigure}%
\begin{subfigure}{.18\textwidth}
  \centering
  \includegraphics[width=\linewidth]{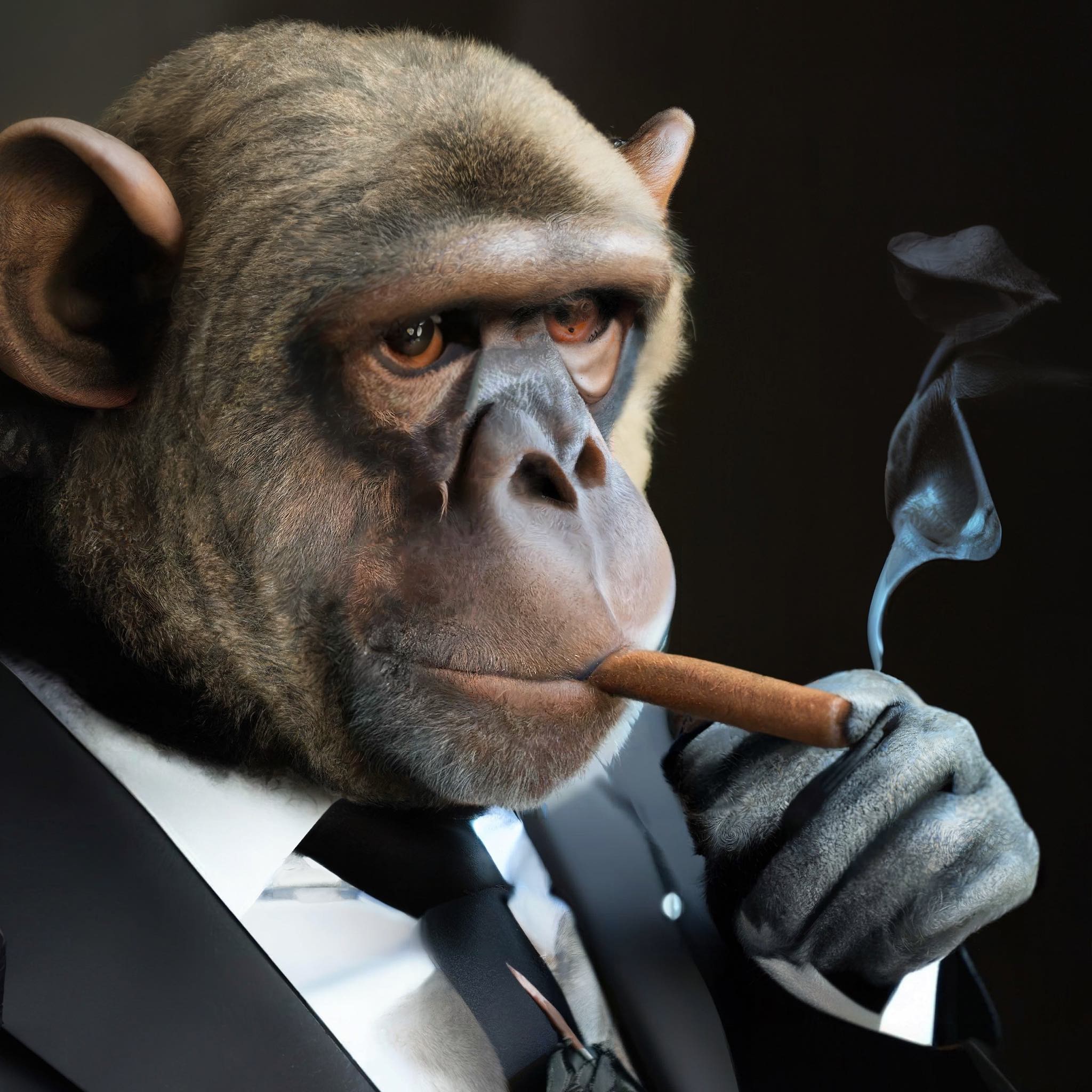}
\end{subfigure}%
\begin{subfigure}{.18\textwidth}
  \centering
  \includegraphics[width=\linewidth]{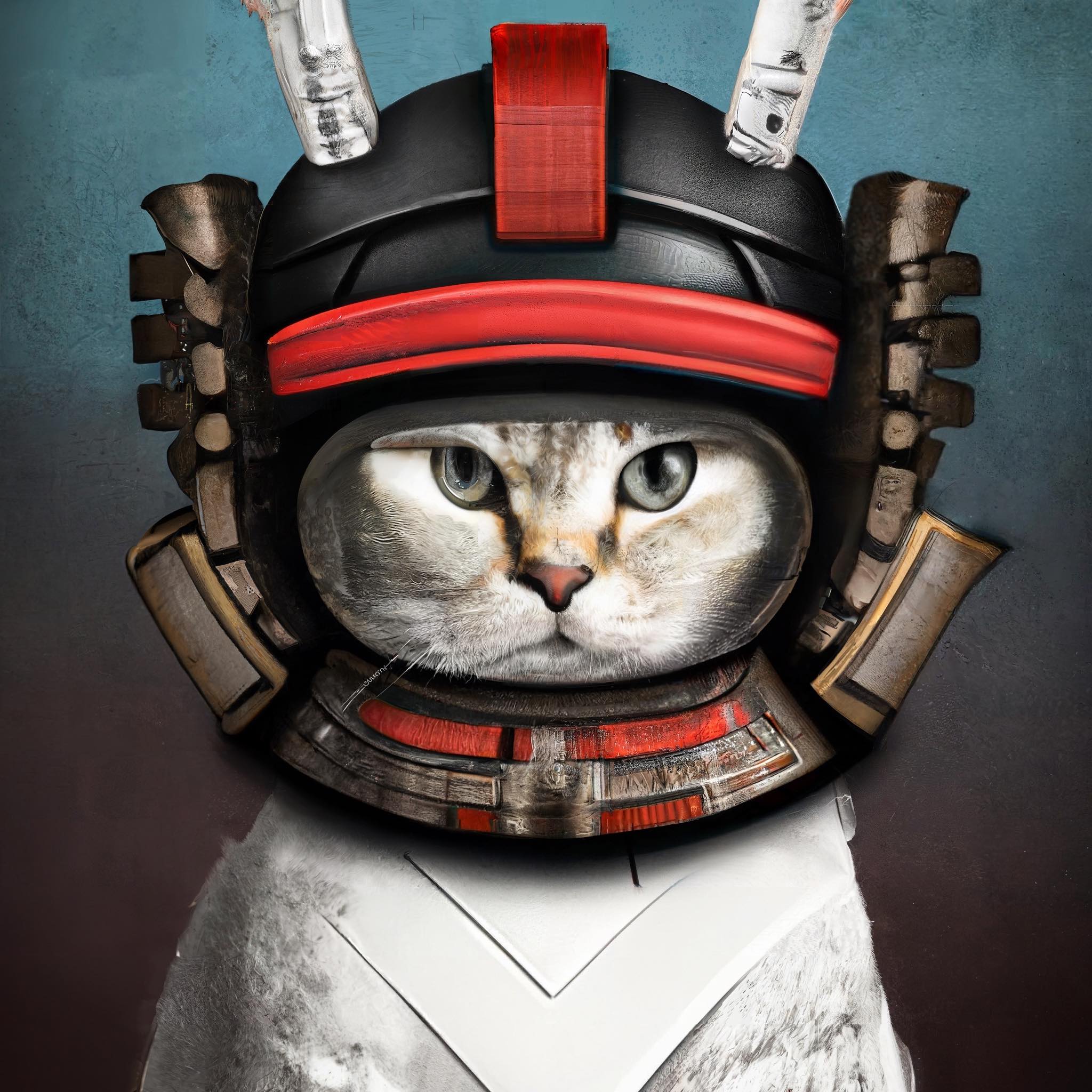}
\end{subfigure}%
\begin{subfigure}{.18\textwidth}
  \centering
  \includegraphics[width=\linewidth]{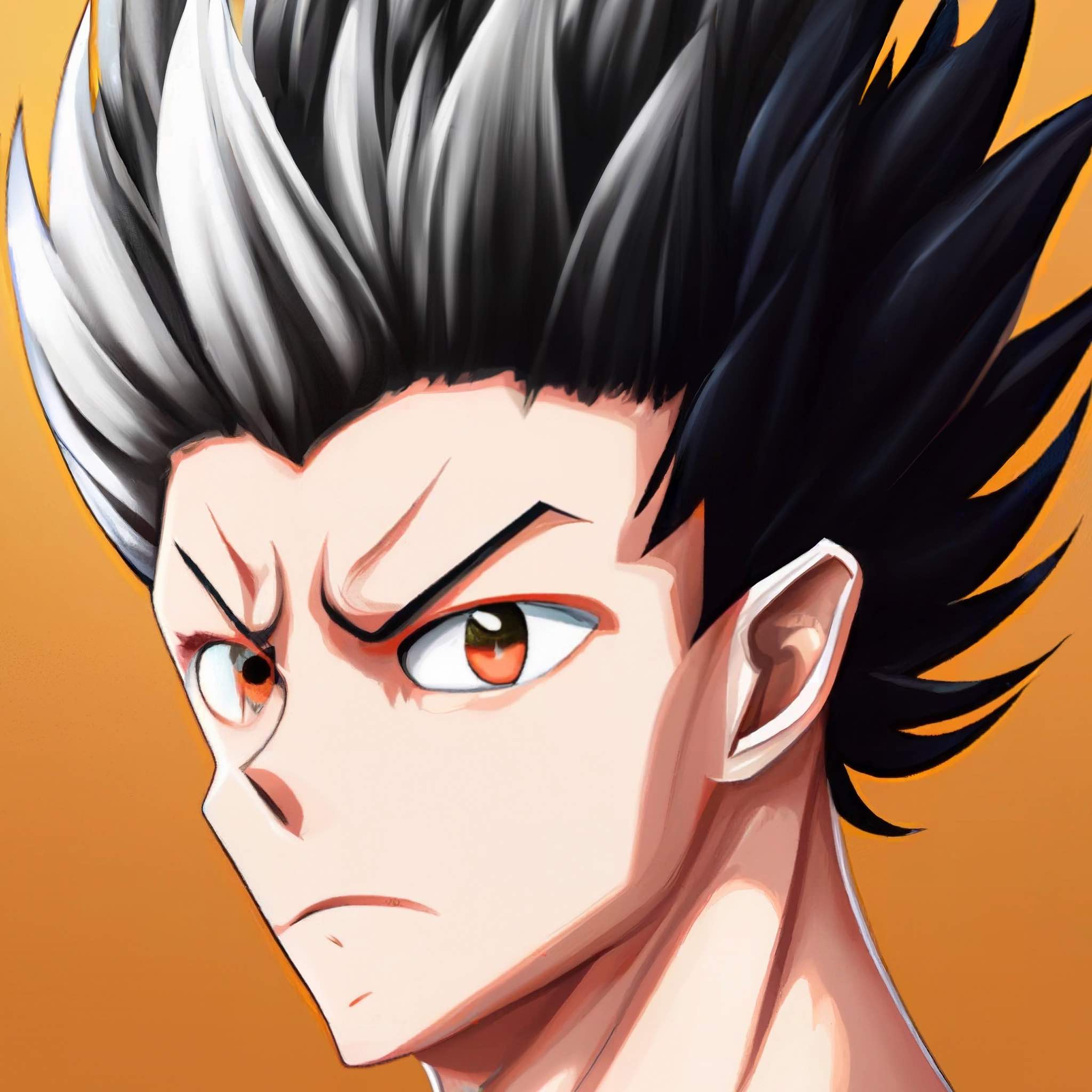}
\end{subfigure}%
\caption{Super-resolution applied with DALL-E2 model: upsampling image to $2048\times 2048$. In order to reduce the size of the PDF, this part has been compressed by JPEG. Access the project page to view the original image.}
\label{fig:images}
\vspace{-4mm}
\end{figure}
\FloatBarrier
\noindent \textbf{GLIDE}
 
\FloatBarrier
\begin{figure}[H]
\vspace{-4mm}
\centering
\begin{subfigure}{.18\textwidth}
  \centering
  \includegraphics[width=\linewidth]{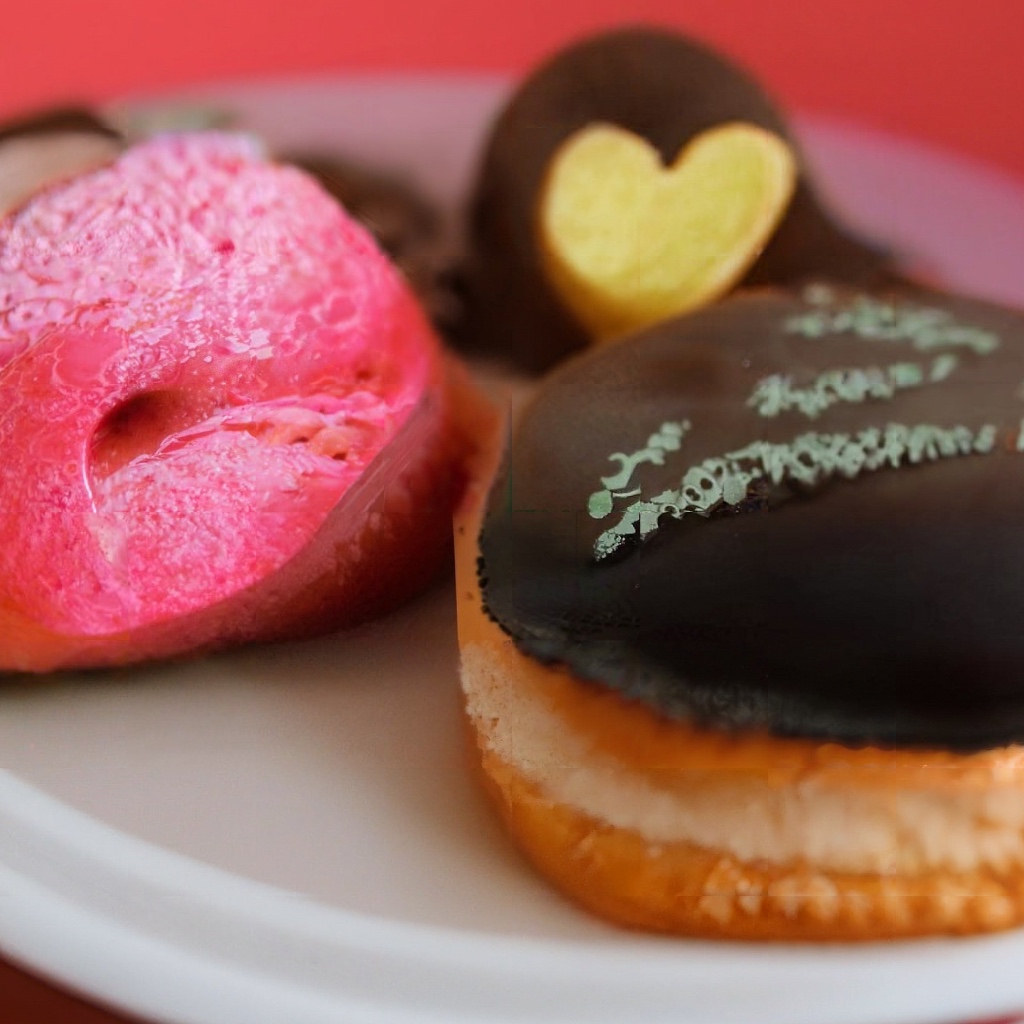}
\end{subfigure}%
\begin{subfigure}{.18\textwidth}
  \centering
  \includegraphics[width=\linewidth]{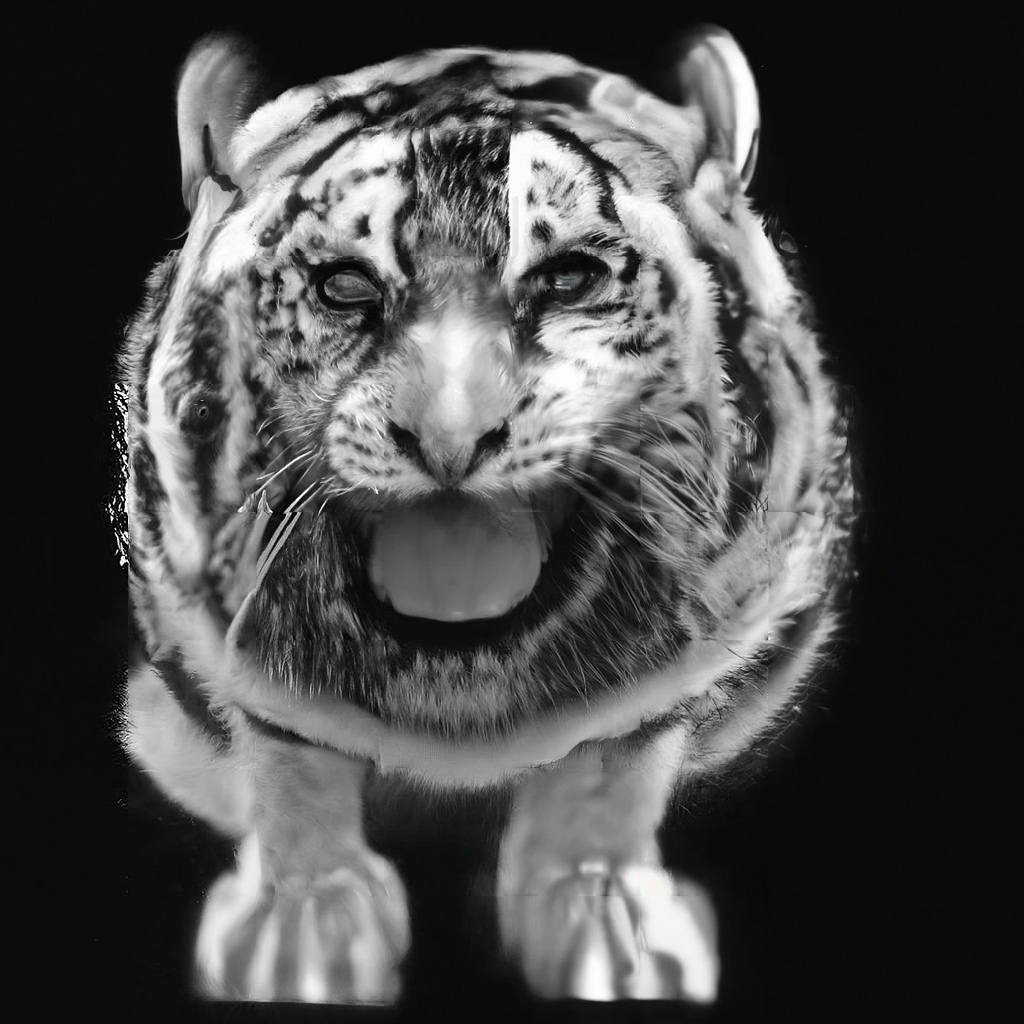}
\end{subfigure}%
\begin{subfigure}{.18\textwidth}
  \centering
  \includegraphics[width=\linewidth]{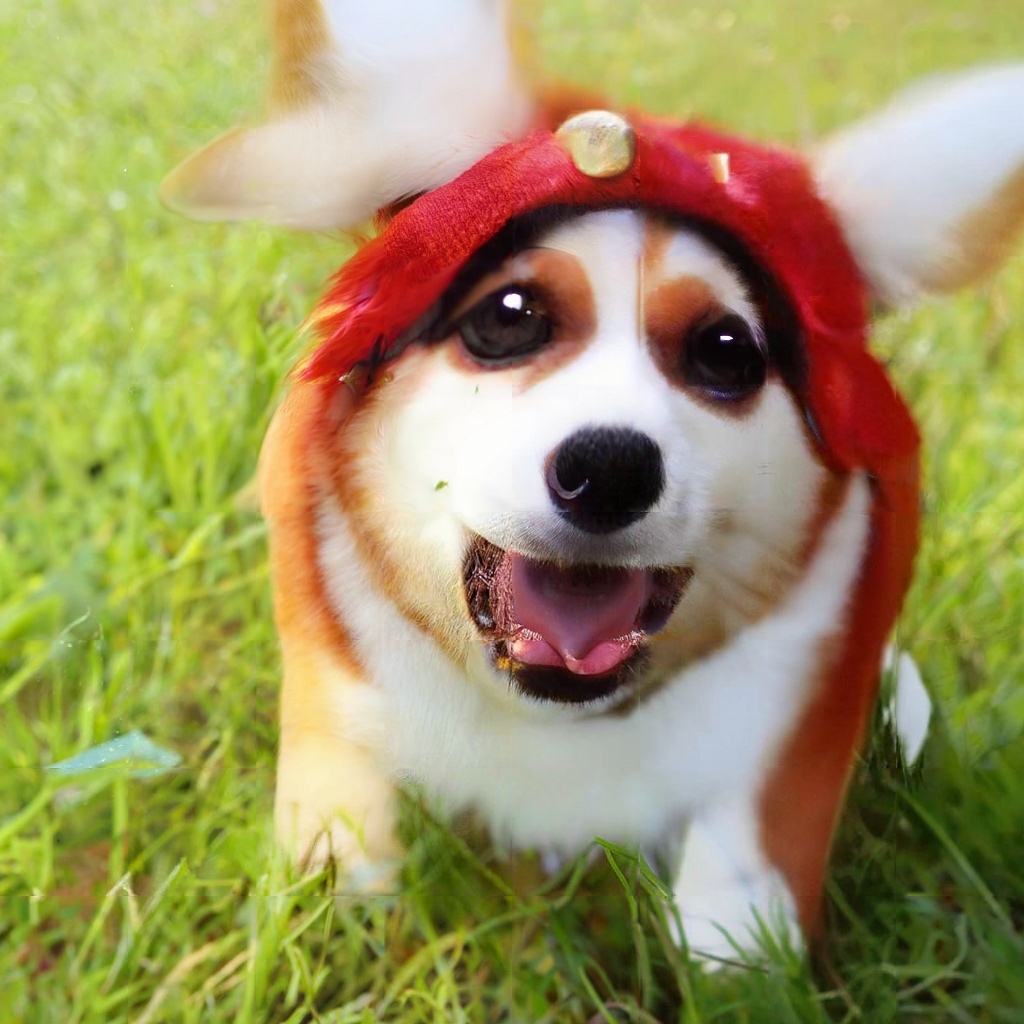}
\end{subfigure}%
\begin{subfigure}{.18\textwidth}
  \centering
  \includegraphics[width=\linewidth]{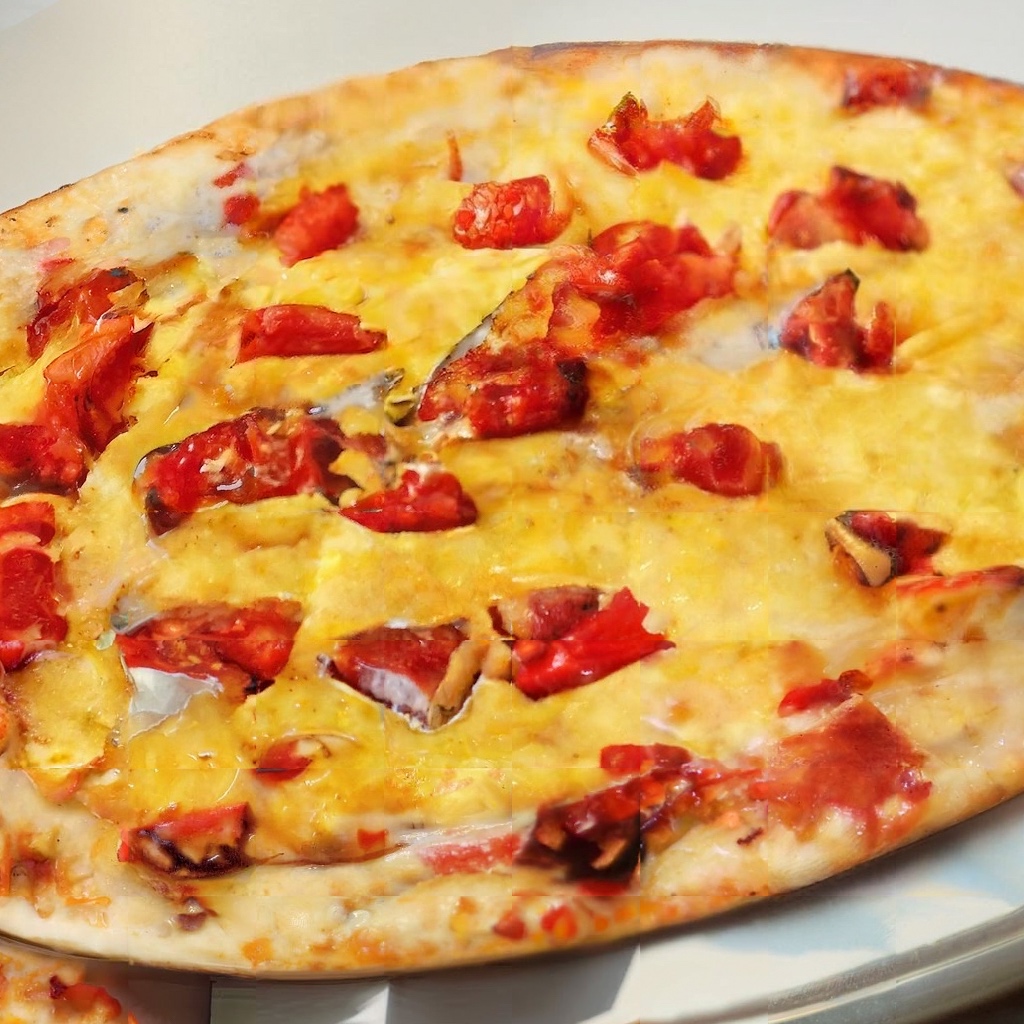}
\end{subfigure}%
\begin{subfigure}{.18\textwidth}
  \centering
  \includegraphics[width=\linewidth]{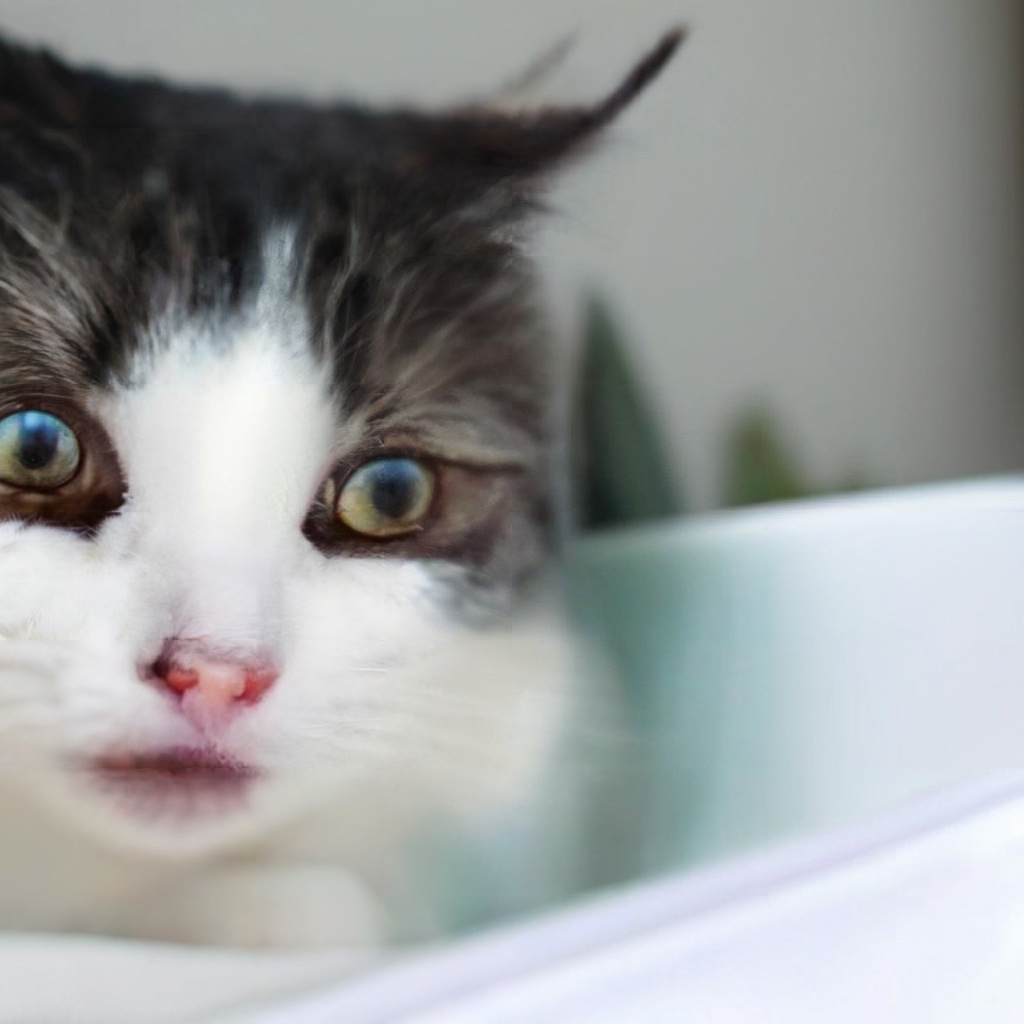}
\end{subfigure}%
\caption{Super-resolution applied with GLIDE model: upsampling image to $1024\times 1024$ resolution.}
\label{fig:images}
\vspace{-4mm}
\end{figure}

\subsection{Super-Resolution Cases}
We present several cases to demonstrate the model's super-resolution capabilities and to compare its performance with other super-resolution models.

\begin{figure}[H]
    \centering
    \includegraphics[width=0.9\linewidth]{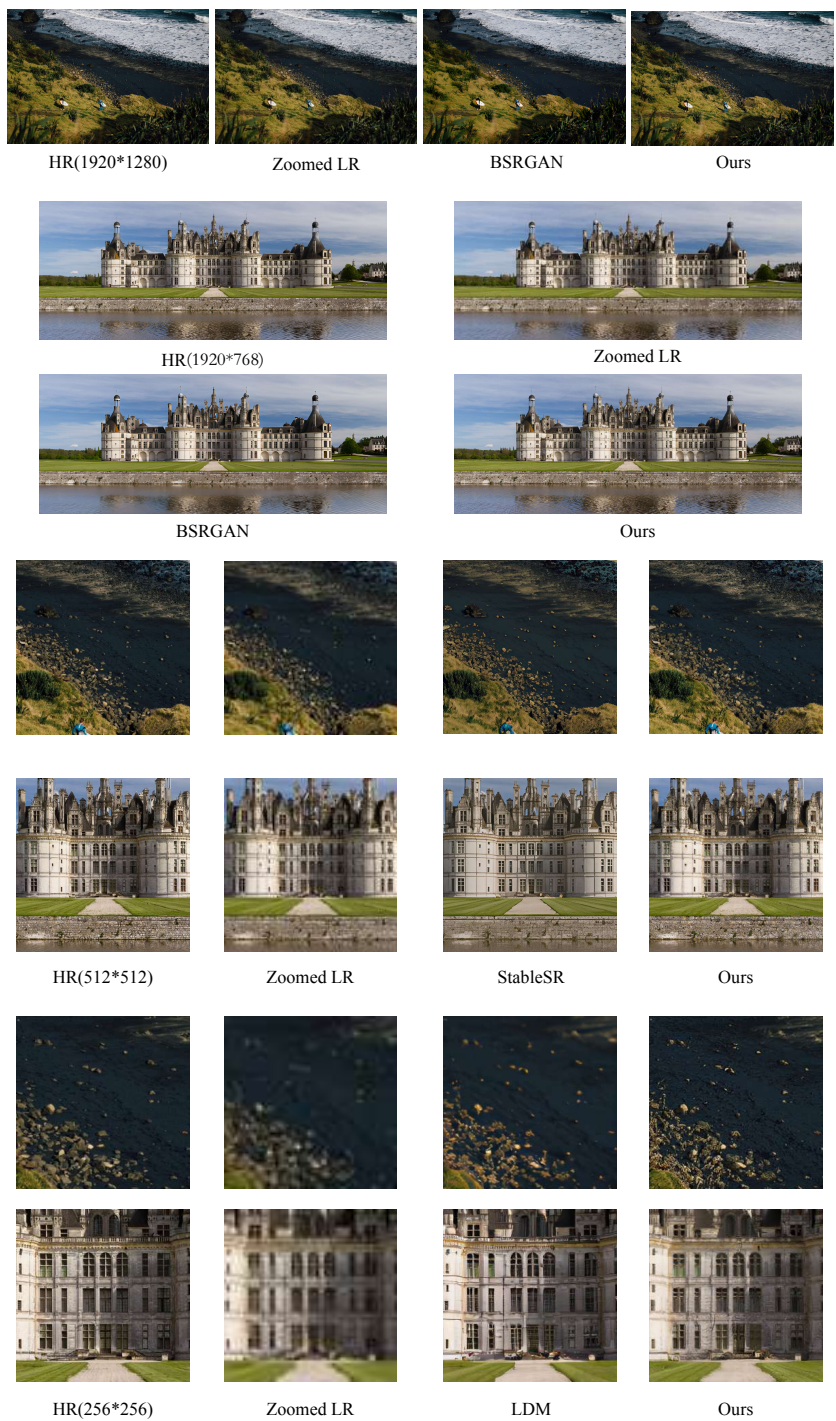}
    \caption{Comparison of super-resolution performance at different resolutions.(\textbf{Zoom in} for details). Images are from DIV2K-valid dataset.}
    \label{fig:sr}
\end{figure}

\section{Human Evalutaion Detail}
\begin{figure}[H]
    \centering
    \includegraphics[width=0.8\linewidth]{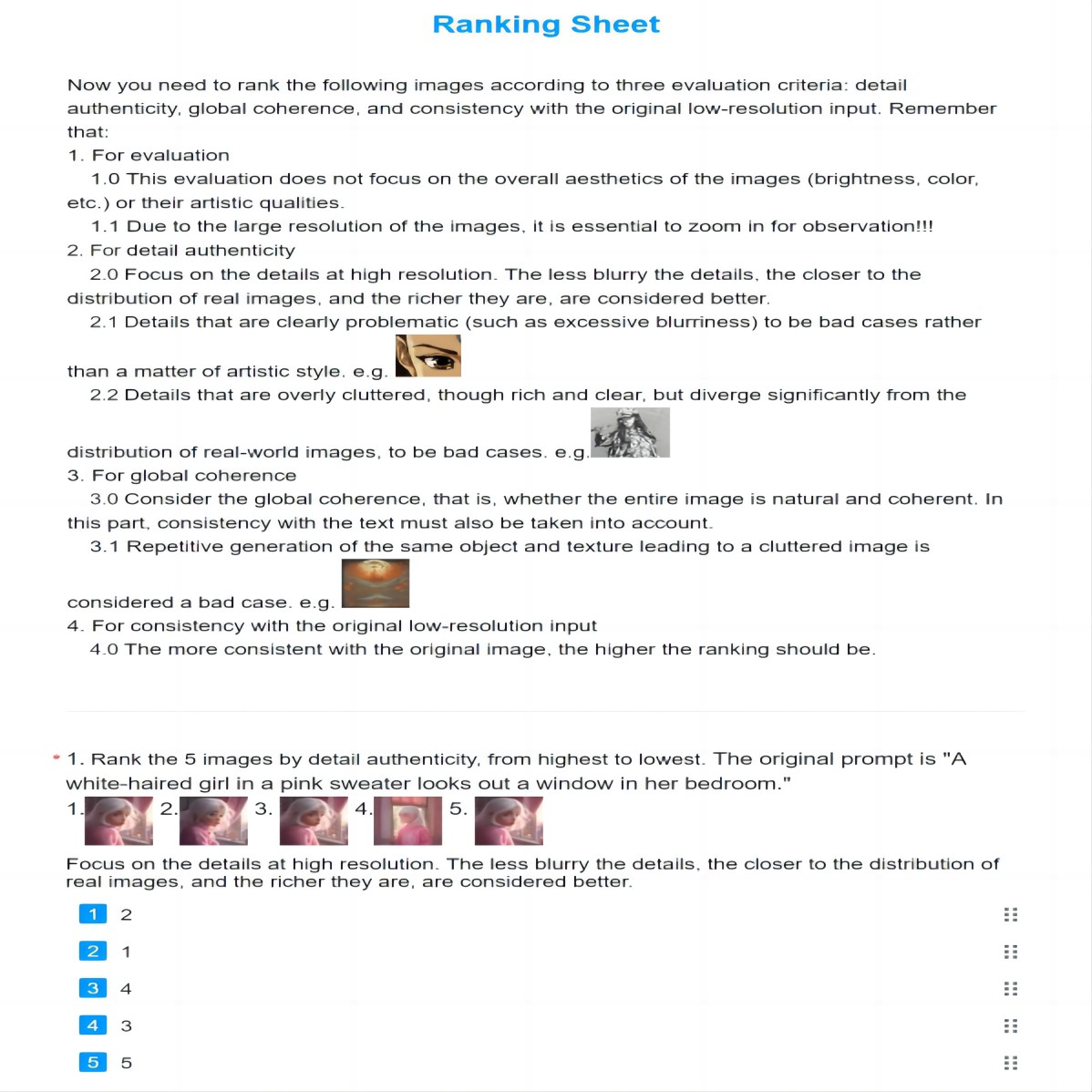}
    \caption{human evaluation questionnaire.}
    \label{fig:humanevl}
    \vspace{-10mm}
\end{figure}

Here, we provide more details about our user study settings. We adopt WenJuanXing as the evaluation platform. A simple example is shown in \cref{fig:humanevl}. A single evaluator would receive $3\times 20$ questions based on the given image and prompt. 
At the beginning of the survey, we provide clear explanations for the criteria used to evaluate each indicator. And we clarify some easily confusing cases, and also provide examples of bad cases. Evaluators are strongly urged to zoom in on the images before making their assessments.

\begin{figure}[H]
    \centering
    \includegraphics[width=\linewidth]{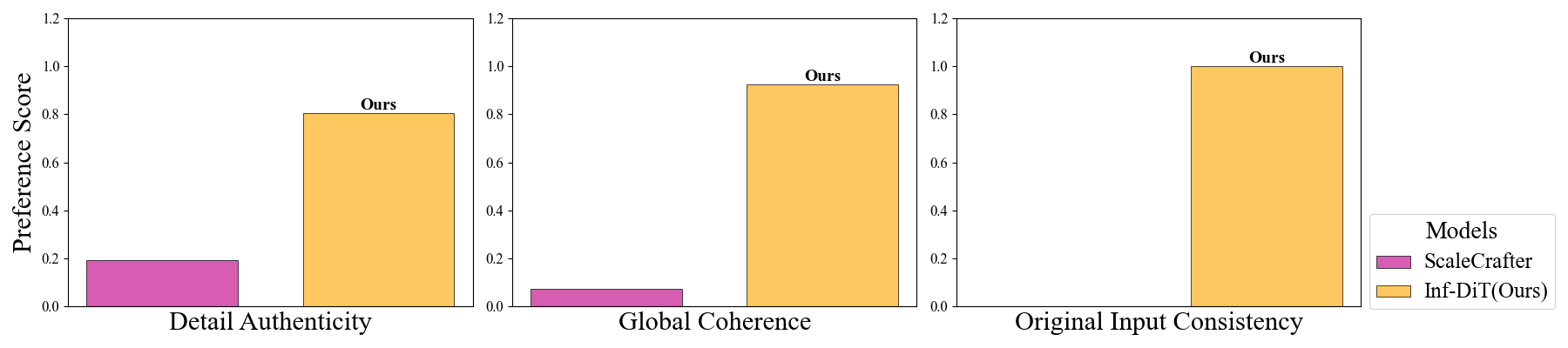}
    \caption{Preference comparison results with ScaleCrafter}
    \label{fig:prescale}
\end{figure}

\begin{figure}[H]
    \centering
    \includegraphics[width=\linewidth]{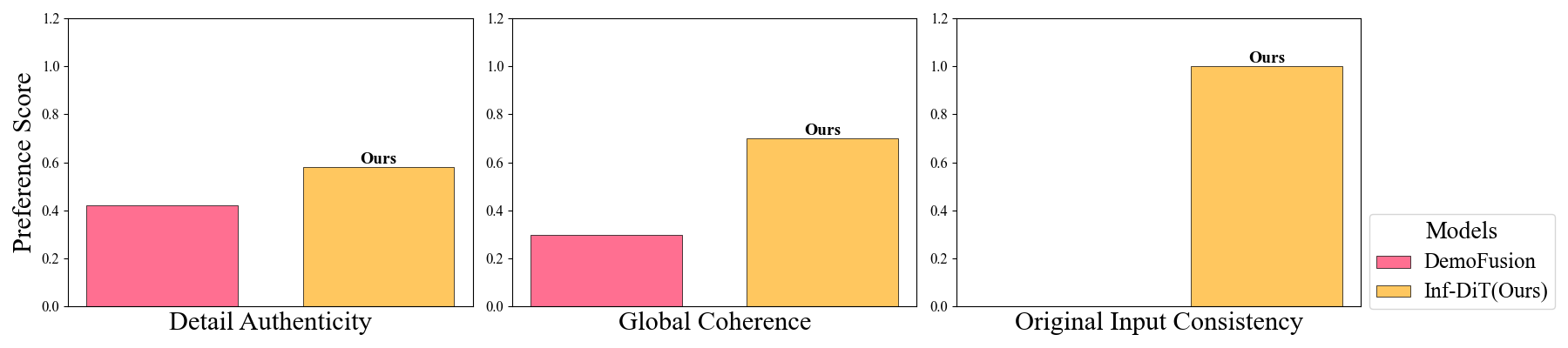}
    \caption{Preference comparison results with DemoFusion}
    \label{fig:predemo}
\end{figure}

\begin{figure}[H]
    \vspace{-10mm}
    \centering
    \includegraphics[width=\linewidth]{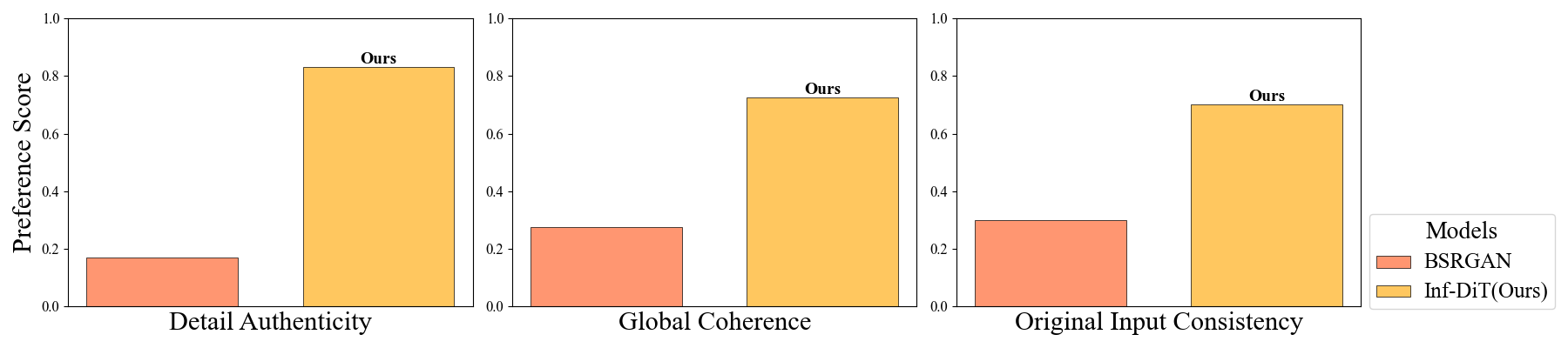}
    \caption{Preference comparison results with BSRGAN}
    \label{fig:prebsr}
\end{figure}

To better showcase the results of manual evaluations, we conduct a preference analysis on the rankings of our model versus other models across three criteria. The results are visible in \cref{fig:prescale}, \cref{fig:predemo} and \cref{fig:prebsr}. The results show that, on any criterion, evaluators exhibit a strong preference for our model.
